%% file: Flow_Blur_Motion_TIP_Final.tex
\documentclass[journal]{IEEEtran}

\usepackage{epsfig}
\usepackage{graphicx}
\usepackage{amsmath}
\usepackage{amssymb}
\usepackage{multirow}
\usepackage{epstopdf}
\usepackage[linesnumbered,ruled]{algorithm2e}
\usepackage{color}

\usepackage{mathtools}
\usepackage[x11names]{xcolor}
\newtagform{red}{\color{black}(}{)}
 
\newcommand\ie{\emph{i.e.}} 
 
\newcommand\etc{\emph{etc.}}
\newcommand\wrt{w.r.t.} 
\newcommand\etal{\emph{et al.}}
\newcommand{\vB}{\mathbf{B}}
\newcommand{\vA}{\mathbf{A}}

\newcommand{\vL}{\mathbf{L}}

\newcommand{\vM}{\mathbf{M}}
\newcommand{\vp}{\mathbf{p}}
\newcommand{\vq}{\mathbf{q}}
\newcommand{\vx}{\mathbf{x}}

\newcommand{\vo}{\mathbf{o}}
\newcommand{\vn}{\mathbf{n}}

\newcommand{\calS}{\mathcal{S}}

\usepackage{changes}
	\definechangesauthor[name={Per cusse}, color=orange]{per}
	\setremarkmarkup{(#2)}

\def\rc#1{{\color{black}{}{{#1}}{}}}
\def\rcs#1{{\color{black}{}{{#1}}{}}}
\begin{document}
%%%%%%%%% TITLE
\title{Joint Stereo Video Deblurring, Scene Flow Estimation and Moving Object Segmentation}

%\title{\LARGE Closing the Loop: Joint Stereo Video Deblurring, Scene Flow Estimation and Moving object segmentation}

\author{Liyuan~Pan,
        Yuchao~Dai, %~\IEEEmembership{Member,~IEEE,}
        Miaomiao~Liu, %~\IEEEmembership{Member,~IEEE,}
        Fatih~Porikli, %~\IEEEmembership{Fellow,~IEEE,}
        and~Quan~Pan%~\IEEEmembership{Member,~IEEE}% <-this % stops a space
\thanks{Liyuan Pan, Miaomiao Liu and Fatih Porikli are with Research School of Engineering, the Australian National University, Canberra, Australia.}% <-this % stops a space
\thanks{Yuchao Dai is with School of Electronics and Information, Northwestern Polytechnical University, Xi'an, China. Yuchao Dai (daiyuchao@gmail.com) is the corresponding author.}% <-this % stops a space
\thanks{Quan Pan is with School of Automation, Northwestern Polytechnical University, Xi'an, China.}% <-this % stops a space
}

\maketitle
%%%%%%%%%%%%%%%%%%% ABSTRACT%%%%%%%%%%%%%%%%%%%%
\begin{abstract}
Stereo videos for the dynamic scenes often show unpleasant blurred effects due to the camera motion and the multiple moving objects with large depth variations. 
Given consecutive blurred stereo video frames, we aim to recover the latent clean images, estimate the 3D scene flow and segment the multiple moving objects. These three tasks have been previously addressed separately, which fail to exploit the internal connections among these tasks and cannot achieve optimality.
In this paper, we propose to jointly solve these three tasks in a unified framework by exploiting their intrinsic connections. To this end, we represent the dynamic scenes with the piece-wise planar model, which exploits the local structure of the scene and expresses various dynamic scenes. Under our model, these three tasks are naturally connected and expressed as the parameter estimation of 3D scene structure and camera motion (structure and motion for the dynamic scenes).
By exploiting the blur model constraint, the moving objects and the 3D scene structure, we reach an energy minimization formulation for joint deblurring, scene flow and segmentation.
We evaluate our approach extensively on both synthetic datasets and publicly available real datasets with fast-moving objects, camera motion, uncontrolled lighting conditions and shadows.
Experimental results demonstrate that our method can achieve significant improvement in stereo video deblurring, scene flow estimation and moving object segmentation, over state-of-the-art methods.
\end{abstract}


%%%%%%%%%%%%%%%%%%%%%%%%%%% Introduction%%%%%%%%%%%%%%%%%%%%%%
\section{Introduction}
\IEEEPARstart{I}{mage} deblurring aims at recovering the latent clean image from a single or multiple blurred images, which is a classic and fundamental task in image processing and computer vision. Image blur could be caused by various reasons, for example, optical aberration, medium perturbation, defocus, and motion~\cite{schuler2012blind,shi2015just, gupta2010single, jia2014mathematical, sun2015learning}.
%The blur not only reduces the quality of the images causing loss of important details, but also hampers further analysis.
\rc{In this work, we only focus on motion blur, which is widely encountered in real-world applications such as autonomous driving~\cite{franke2000real, geiger2012we}}. The effects become more apparent when the exposure time increased due to low-light conditions. %

%=======figure deblurresult=====
\begin{figure}[!ht]
\begin{center}
\begin{tabular}{cc}
\hspace{-0.2cm}
\includegraphics[width=0.219\textwidth]{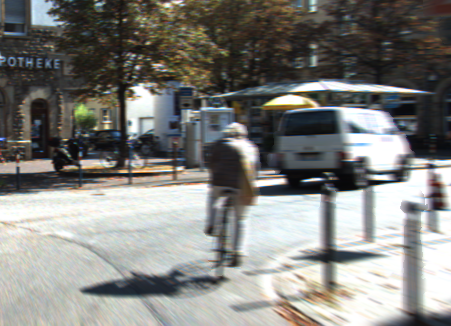}
&\includegraphics[width=0.219\textwidth]{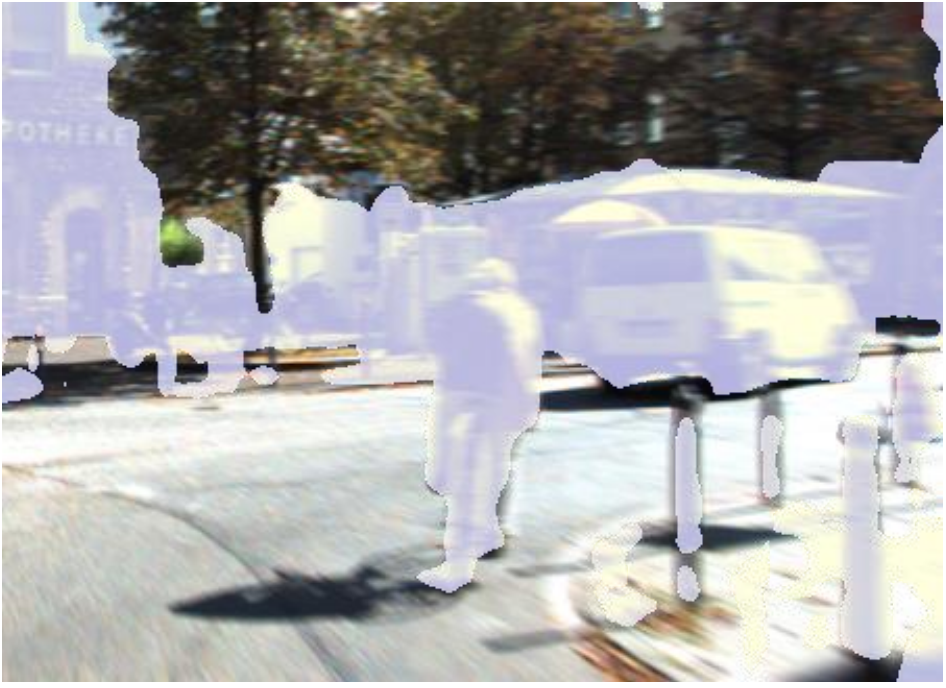}\\
\hspace{-0.2cm}
(a) \rcs{Blurred image} & (b) \rc{Initial segmentation} \\
\hspace{-0.2cm}
\includegraphics[width=0.219\textwidth]{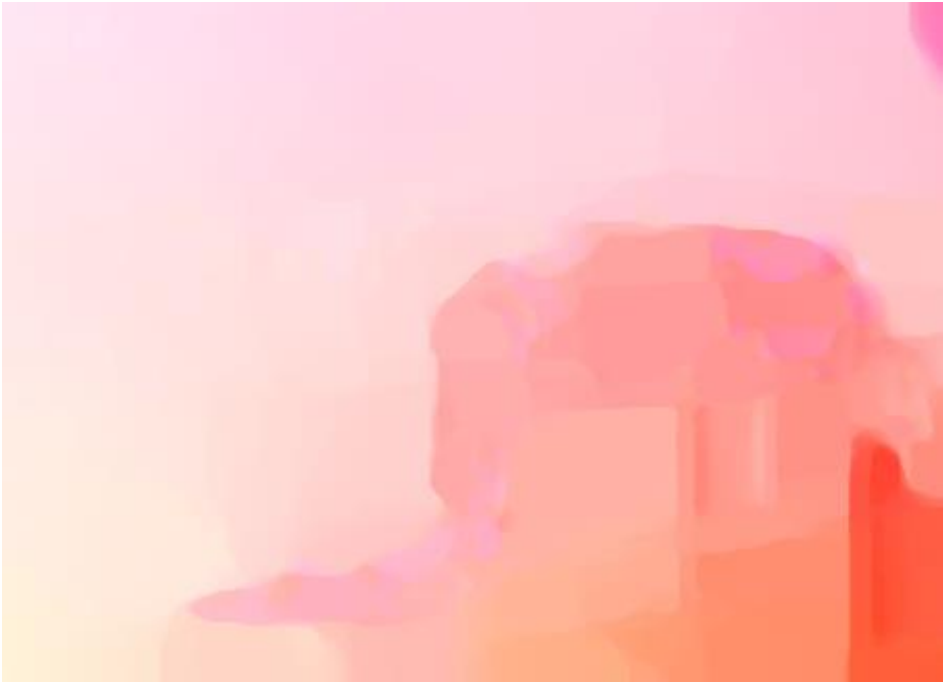}
&\includegraphics[width=0.219\textwidth]{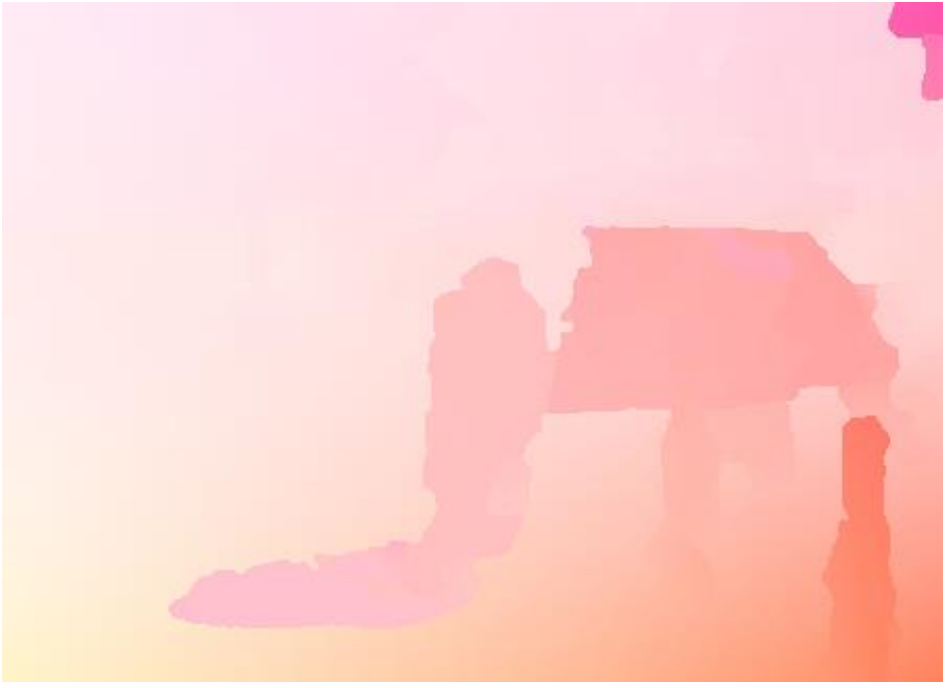}\\
\hspace{-0.2cm}
(c) Kim and Lee~\cite{hyun2015generalized} &(d) Our flow  \\
\hspace{-0.2cm}
\includegraphics[width=0.219\textwidth]{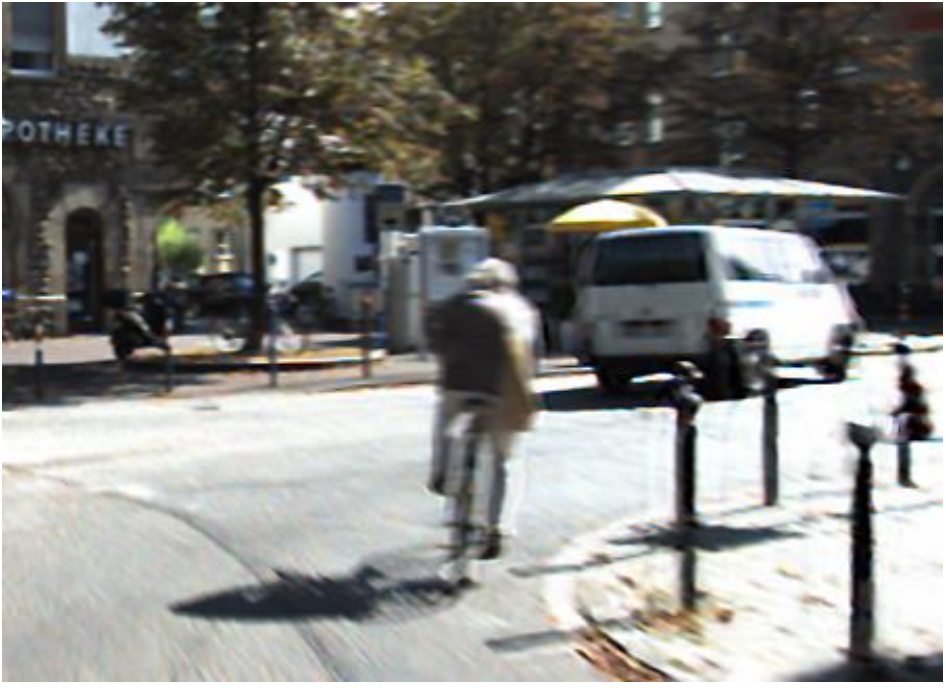}
&\includegraphics[width=0.219\textwidth]{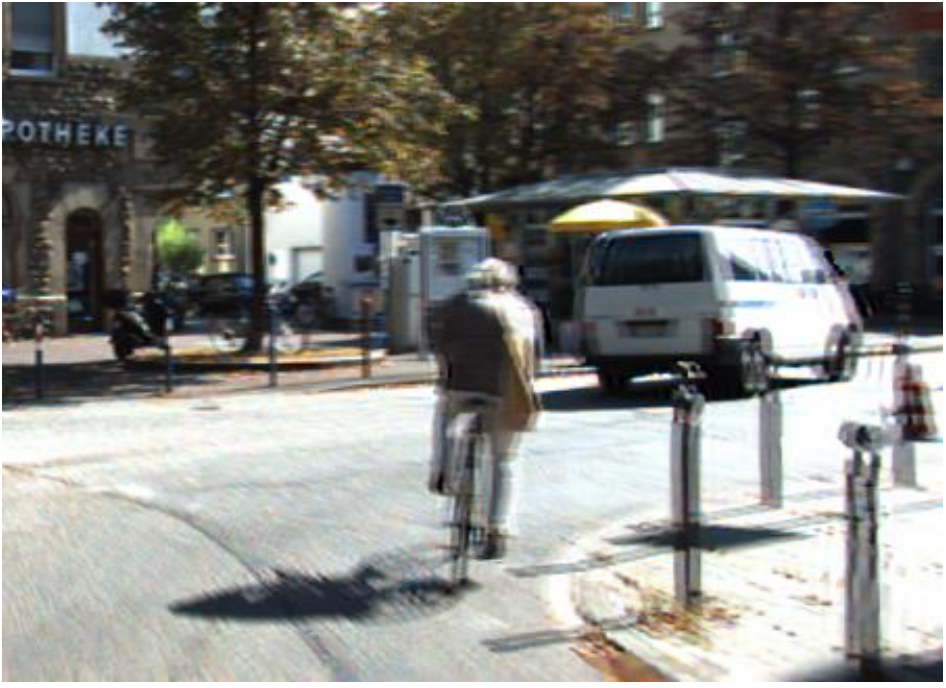}\\
\hspace{-0.2cm}
(e) Kim and Lee~\cite{hyun2015generalized} &(f) Sellent \etal~\cite{sellent2016stereo}\\
\hspace{-0.2cm}
\includegraphics[width=0.219\textwidth]{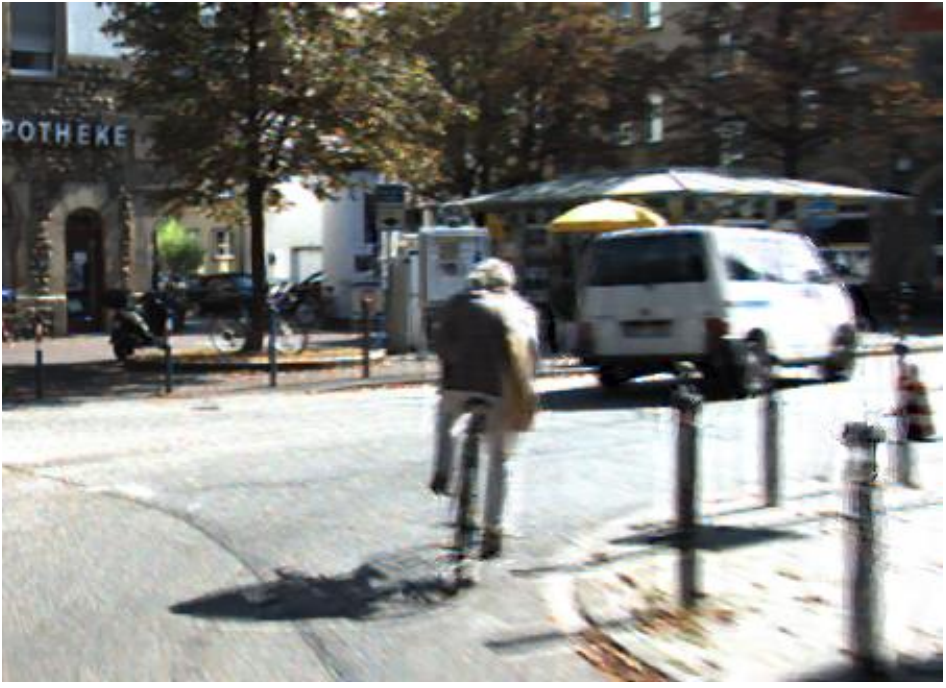}
&\includegraphics[width=0.219\textwidth]{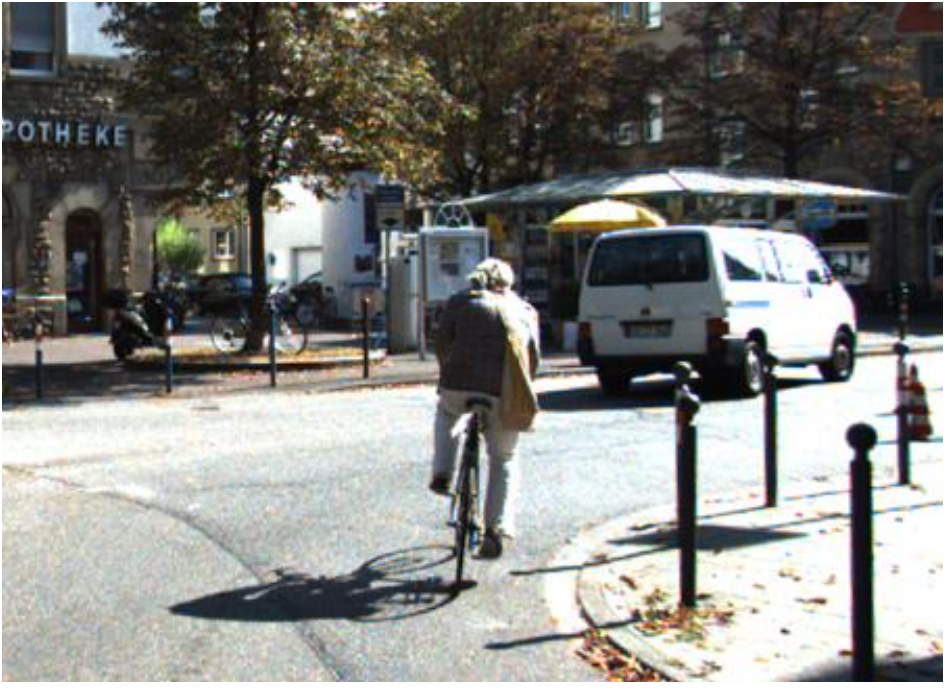}\\
\hspace{-0.2cm}
(g) \rcs{Pan \etal~\cite{Pan_2017_CVPR}}   &(h) Ground-truth  \\
\hspace{-0.2cm}
\includegraphics[width=0.219\textwidth]{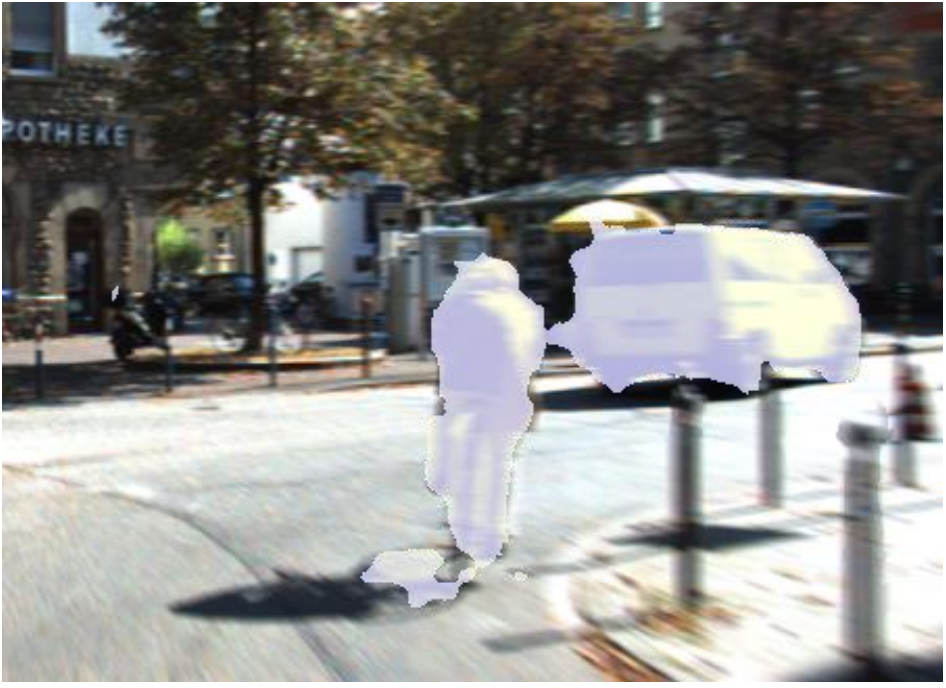}
&\includegraphics[width=0.219\textwidth]{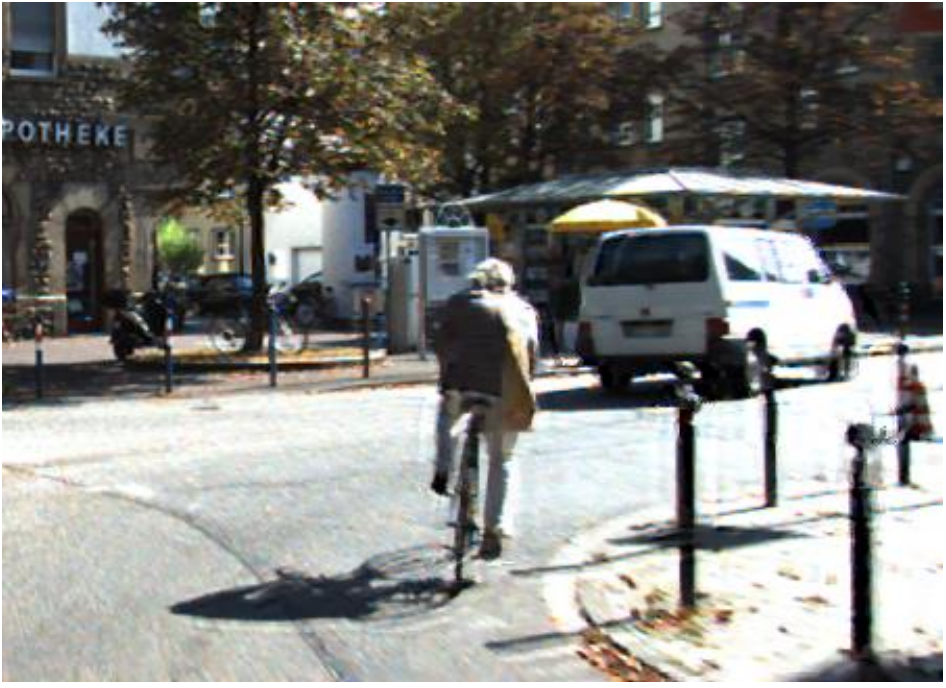}\\
\hspace{-0.2cm}
(i) \rcs{Our segmentation}   &(j) Our deblurred result \\
\end{tabular}
\end{center}
\vspace{-2.9 mm}
\caption{\rc{Stereo deblurring, scene flow estimation and moving object segmentation results with (a) and (b) as input. 
(a) Blurred image. 
(b) Initial segmentation prior. 
(c) Flow estimation by~\cite{hyun2015generalized}. 
(d) Our flow estimation result. 
(e) Deblurring results by~\cite{hyun2015generalized}. 
(f) Stereo deblurring results by~\cite{sellent2016stereo} which uses~\cite{vogel20153d} to estimate scene flow. 
(g) Deblurring results by~\cite{Pan_2017_CVPR}. 
(h) Ground-truth latent image. 
(i) Our moving object segmentation result. 
(j) Our stereo deblurring result. Best viewed in colour on the screen.}
}
\label{fig:fig1}
\end{figure}

Motion deblurring has been extensively studied and various methods have been proposed in the literature. It is common to model the blur effect using kernels~\cite{jia2014mathematical, lee2013recent}. Early deblurring methods mainly focus on the blur caused by camera shake in constant depth or static scenes with moving objects~\cite{hu2014joint, xu2013unnatural}.
In this work, we focus on a more generalized motion blur caused by both camera motion and moving objects. Therefore, conventional blur removal methods, such as~\cite{gupta2010single, krishnan2011blind} cannot be directly applied since they are restricted to a single or a fixed number of blur kernels, making them inferior in tackling general motion blur problems.

For a scenario where both camera motion and multiple moving objects exist, the blur kernel is, in principle, defined for each pixel individually. Recently, several researchers have studied to handle the blurred images with \emph{spatially-variant blur}~\cite{hyun2015generalized,  sellent2016stereo, Pan_2017_CVPR} which uses accurate motion estimation to model the blur kernel. The phenomenon around motion and blur can be viewed as a chicken-egg problem: effective motion blur removal requires accurate motion estimation. Yet, the accuracy of motion estimation highly depends on the quality of the images.

It is a problem for any of the algorithms exploiting motion information as the condition is a major challenge to reliable flow computation.
 
In this paper, we aim to tackle a `generalized stereo deblurring' problem. The moving stereo cameras observe a dynamic scene with varying depth, and the moving objects' boundaries are mixed with the background pixels. Thus we propose to utilize the motion boundary information provided by semantic segmentation~\cite{wu2016wider}. In our approach, we jointly estimate scene flow, segment the moving objects and deblur the images under a unified framework. Using our formulation, we attain significant improvement in numerous real challenging scenes as illustrated in Fig.~\ref{fig:fig1}.

\begin{figure}
\begin{center}
\begin{tabular}{cc}
\hspace{-0.2cm}
\includegraphics[width=0.225\textwidth]{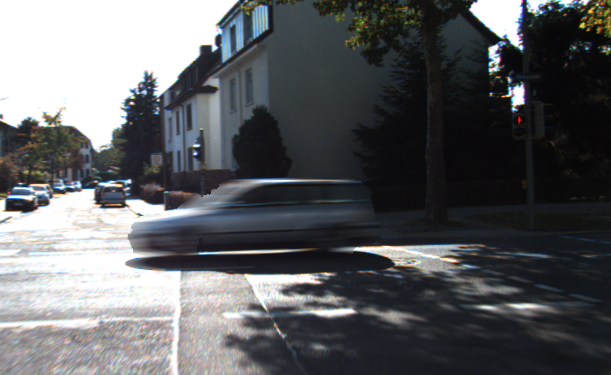}
&\includegraphics[width=0.225\textwidth]{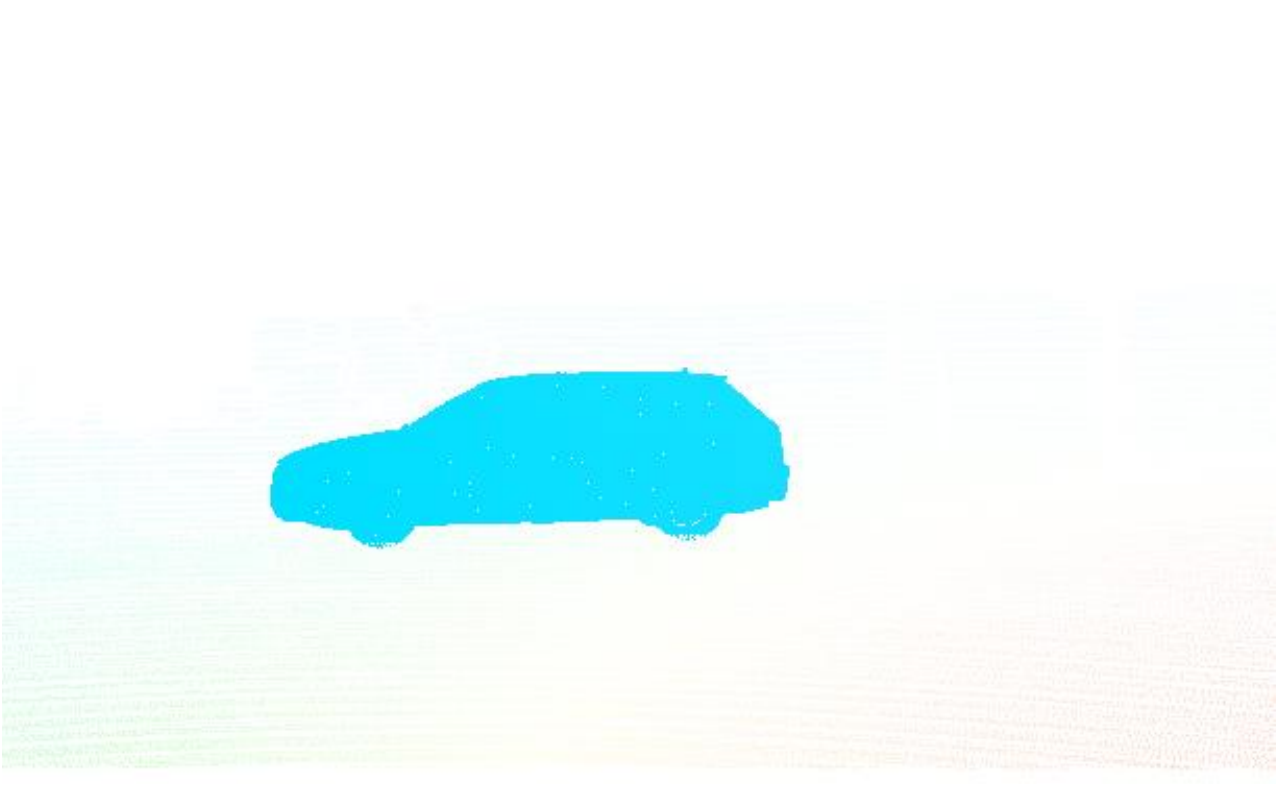}\\
\hspace{-0.2cm}
(a) \rc{Blurred image}  & (b) GT Flow  \\
\hspace{-0.2cm}
\includegraphics[width=0.225\textwidth]{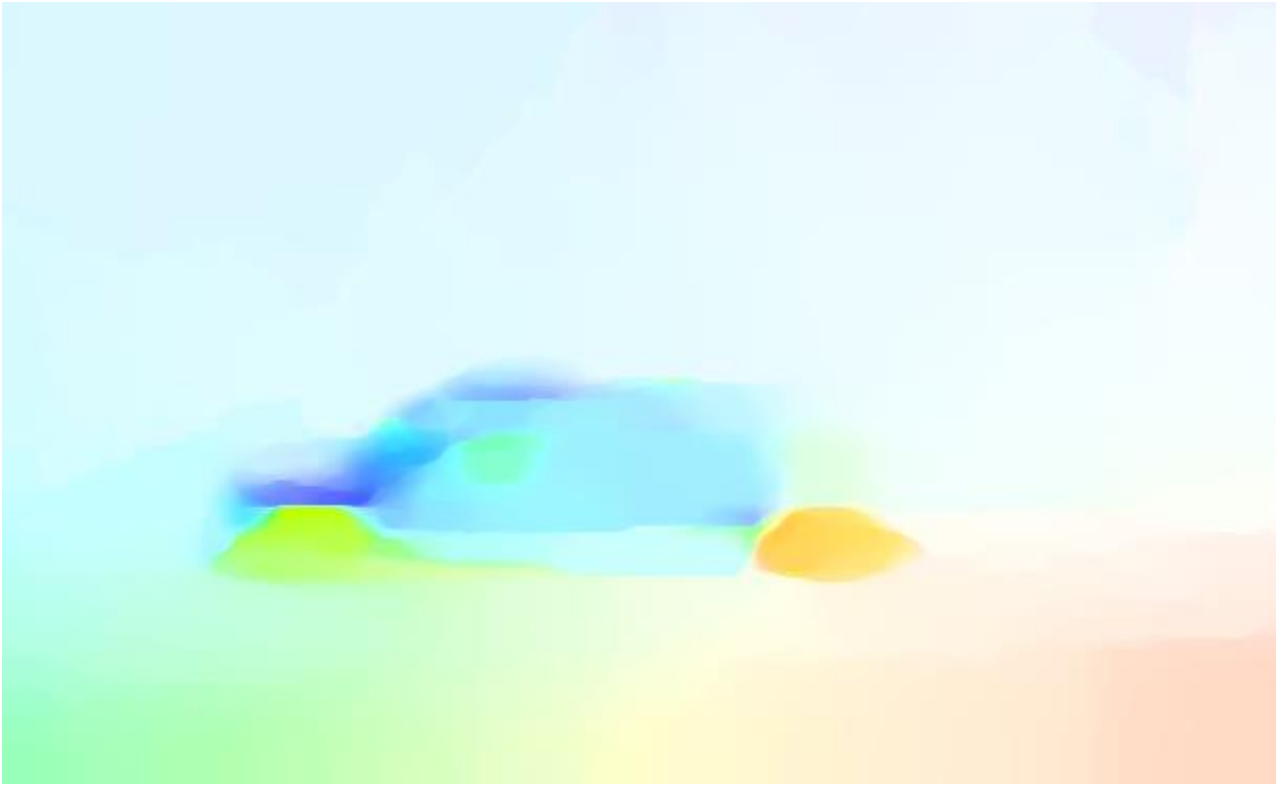}
&\includegraphics[width=0.225\textwidth]{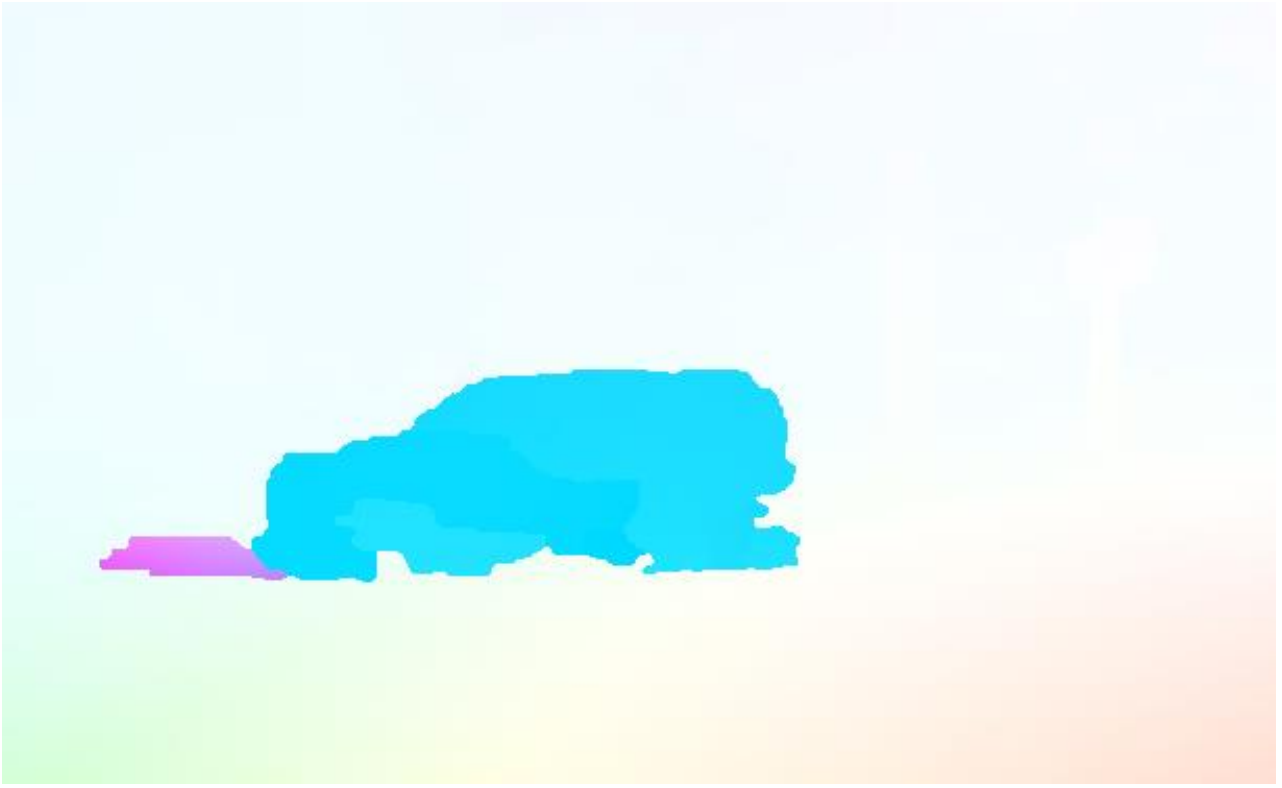}\\
\hspace{-0.2cm}
(c) Kim and Lee~\cite{hyun2015generalized} & (d) Sellent \etal~\cite{sellent2016stereo}\\
\hspace{-0.2cm}
\includegraphics[width=0.225\textwidth]{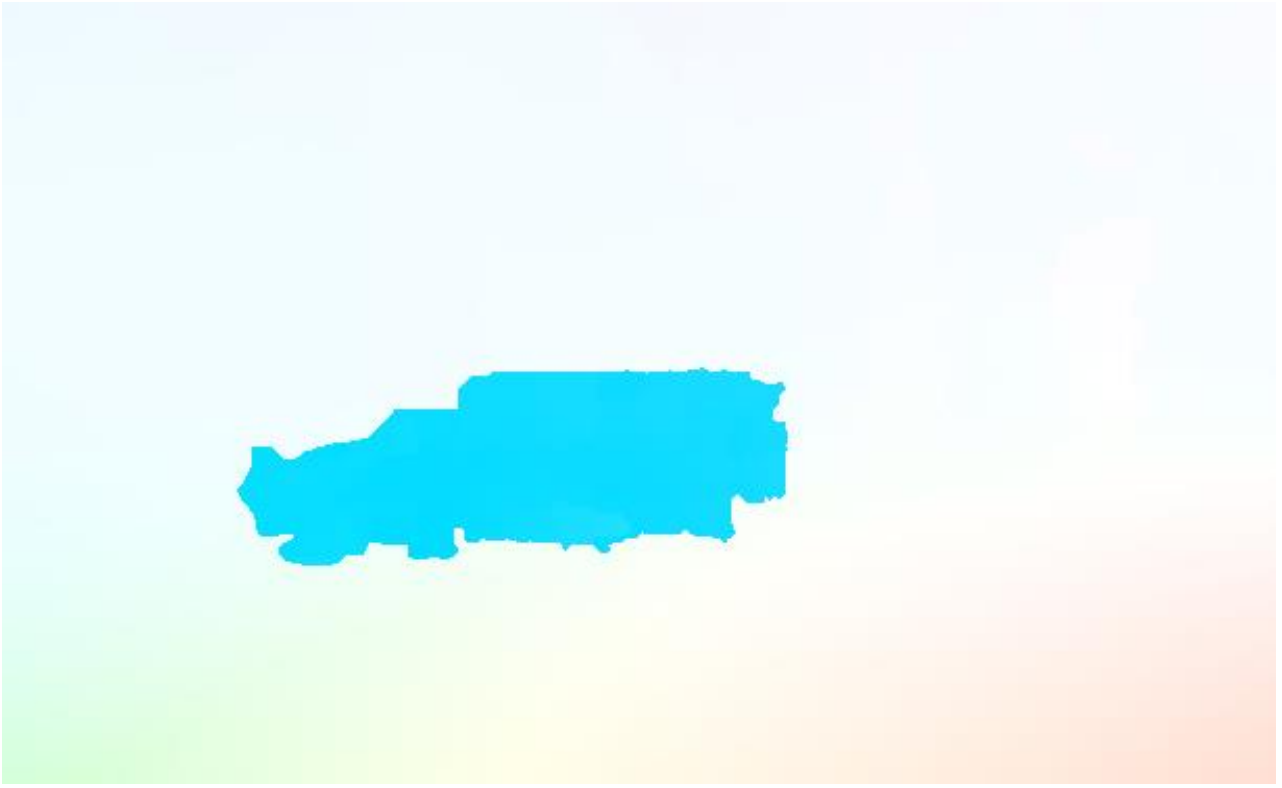}
&\includegraphics[width=0.225\textwidth]{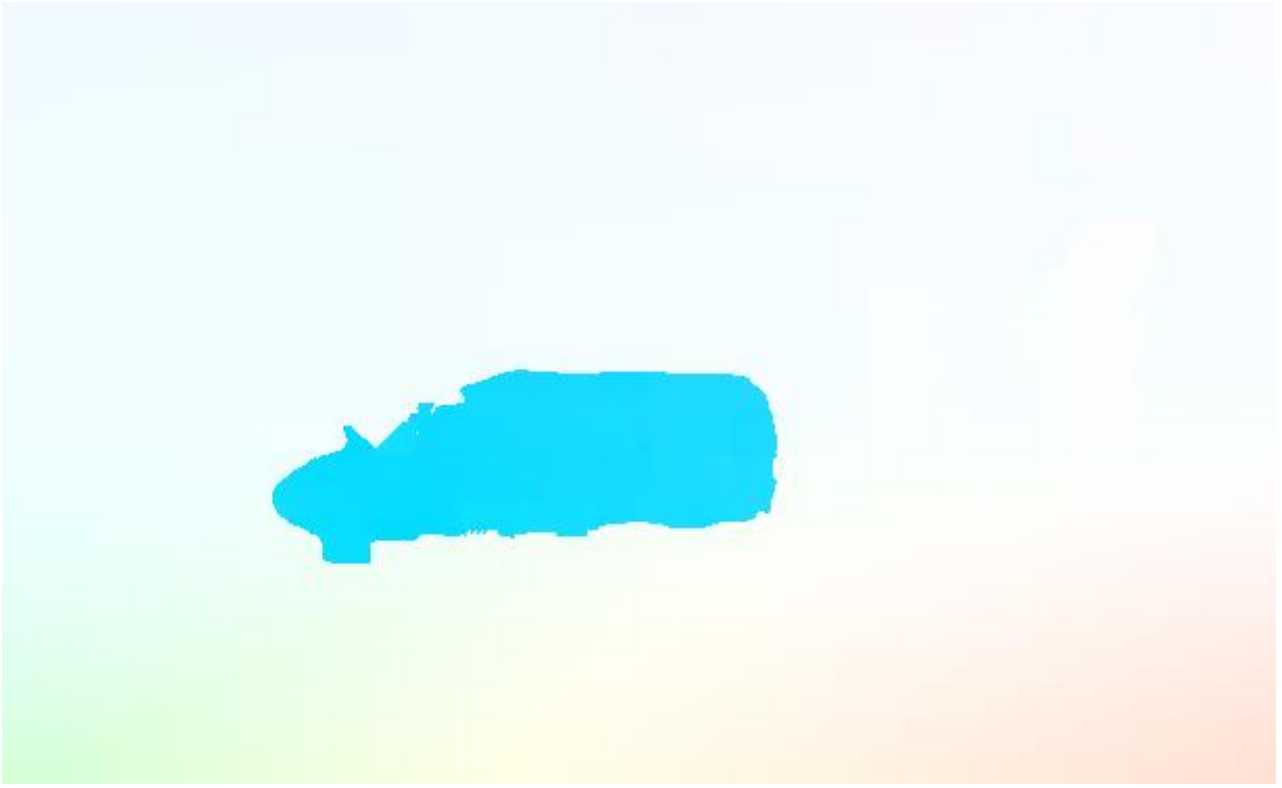}\\
\hspace{-0.2cm}
(e) Pan \etal~\cite{Pan_2017_CVPR} & (f) Our flow estimation\\
\end{tabular}
\end{center}
% \vspace{-2 mm}
\caption{\rc{
Scene flow estimation results for an outdoor scene. (a) Blurred reference image from {\bf BlurData-1}. (b) Ground truth optical flow for the scene. (c) Estimated flow by Kim and Lee~\cite{hyun2015generalized}. (d) Estimated flow by Sellent \etal~\cite{sellent2016stereo} which uses~\cite{vogel20153d} to estimate scene flow. This approach ranks as one of the top 3 approaches on KITTI scene flow benchmark~\cite{geiger2013vision}. (e) Estimated flow by Pan \etal~\cite{Pan_2017_CVPR}. (f) Our flow estimation result. Compared with these state-of-the-art methods, our method achieves the best performance. 
}}
\label{fig:flowcompare}
\end{figure}

We would like to argue that, the scene flow estimation approaches that make use of colour brightness constancy may be hindered by the blurred images. Existing optical flow methods make generic, spatially homogeneous, assumptions about the spatial structure of the flow. 
\rc{Due to the inherent correlation between semantic segmentation and Moving object segmentation (for example, the movement of pixels a vehicle tends to be the same and be different from the background), semantic segmentation has been used to provide motion segmentation prior.} Thus, we investigate the benefits of semantic grouping~\cite{wu2016wider} which are more beneficial for the scene flow estimation task. Here, we only need a coarse and simple semantic segmentation prior to distinguish foreground and background.
The more of the boundary information can be detected during the deconvolution process, the better quality of the estimated results~\cite{zhou2014map,pan2016soft}.
In Fig.~\ref{fig:flowcompare}, we compare the scene flow estimation results with the state-of-the-art solutions on different blurred images. It could be observed that the scene flow estimation performance deteriorates quickly \wrt ~the image blur because of the inaccuracy at boundaries.

On the other hand, motion segmentation or Moving object segmentation alone is also very challenging as the objects could be rigid, non-rigid, and deformable. How to unify these different scene models and achieve moving object segmentation is an active research direction. In this paper, we focus on outdoor traffic scenes with multiple moving objects, such as vehicles, cyclists, and pedestrians. Specifically, we exploit both the semantic cue and 3D geometry cue to better handle moving object segmentation together with scene flow estimation and stereo deblurring.

%There have been considerable efforts in extending the success of deep learning in high level-vision tasks to the low-level vision tasks. However, these supervised learning based methods strongly depend on the consistency between the training datasets and the testing datasets, which hinder their generalization ability for real-world applications.

%=======figure flowcompare=====

Furthermore, existing works fail to exploit the connections between stereo deblurring, scene flow estimation and Moving object segmentation, which actually are closely connected. Specifically, better scene flow estimation and Moving object segmentation will enable better stereo deblurring. Correspondingly, stereo deblurring and Moving object segmentation also help scene flow estimation. However, building their intrinsic connections is not easy as the dynamic scenes could be rather generic, from a static scene to a highly dynamic scene consisting of multiple moving objects (vehicles, pedestrians and etc). Having a unified formulation for the dynamic scenes is highly desired. We propose to exploit the piecewise plane model for the dynamic scene structure, and under this formulation, the joint task of scene flow estimation, stereo deblurring and moving object segmentation has been expressed as the parameter estimation for each planar, the camera motion and pixel labelling. Therefore, we put these three tasks in a loop under a unified energy minimization formulation in which the intra-relation has been effectively exploited. 
 
In our previous work~\cite{Pan_2017_CVPR}, we only consider the relationship between optical flow and deblurring without adding segmentation information. We extend the previous work significantly in the following ways:
\begin{itemize}
\item We propose a novel joint optimization framework to estimate the scene flow, segment moving objects and restore the latent images for generic dynamic scenes. Our deblurring objective benefits from improved boundaries information and the estimated scene structure.

\item We integrate high-level semantic cues for camera motion and scene structure estimation by exploiting the intrinsic connection between semantic segmentation and Moving object segmentation.

\item We propose a method to exploit motion segmentation information in aiding the challenging video deblurring task. 
%The detection results of our method are able to segment all moving objects and also achieve \rc{superpixel-level} accuracy. 
Similarly, the scene flow and objects boundary objective allow deriving more accurate pixel-wise spatially varying blur kernels (see Section.\ref{sec:3.2}).

\item Extensive experiments demonstrate that our method can successfully handle complex real-world scenes depicting fast-moving objects, camera motions, uncontrolled lighting conditions, and shadows. 

\end{itemize}

%%%%%%%%%%%%%%%%%% Related Work %%%%%%%%%%%%%%%%%%
\section{Related Work}

%=================figure flowkernel=========
\begin{figure}
\begin{center}
\begin{tabular}{cc}
\hspace{-0.2cm}
\includegraphics[width=0.225\textwidth]{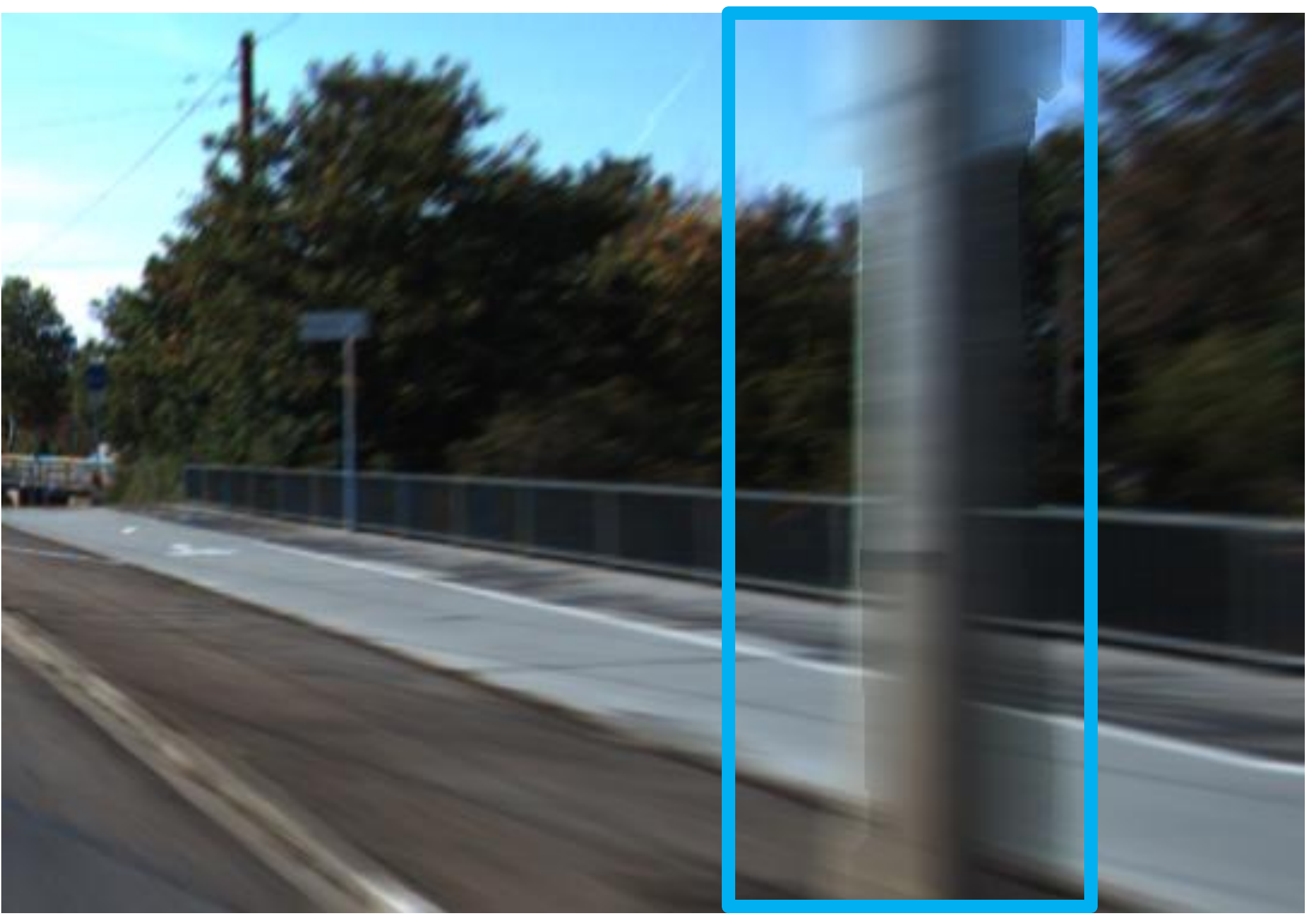}
&\includegraphics[width=0.225\textwidth]{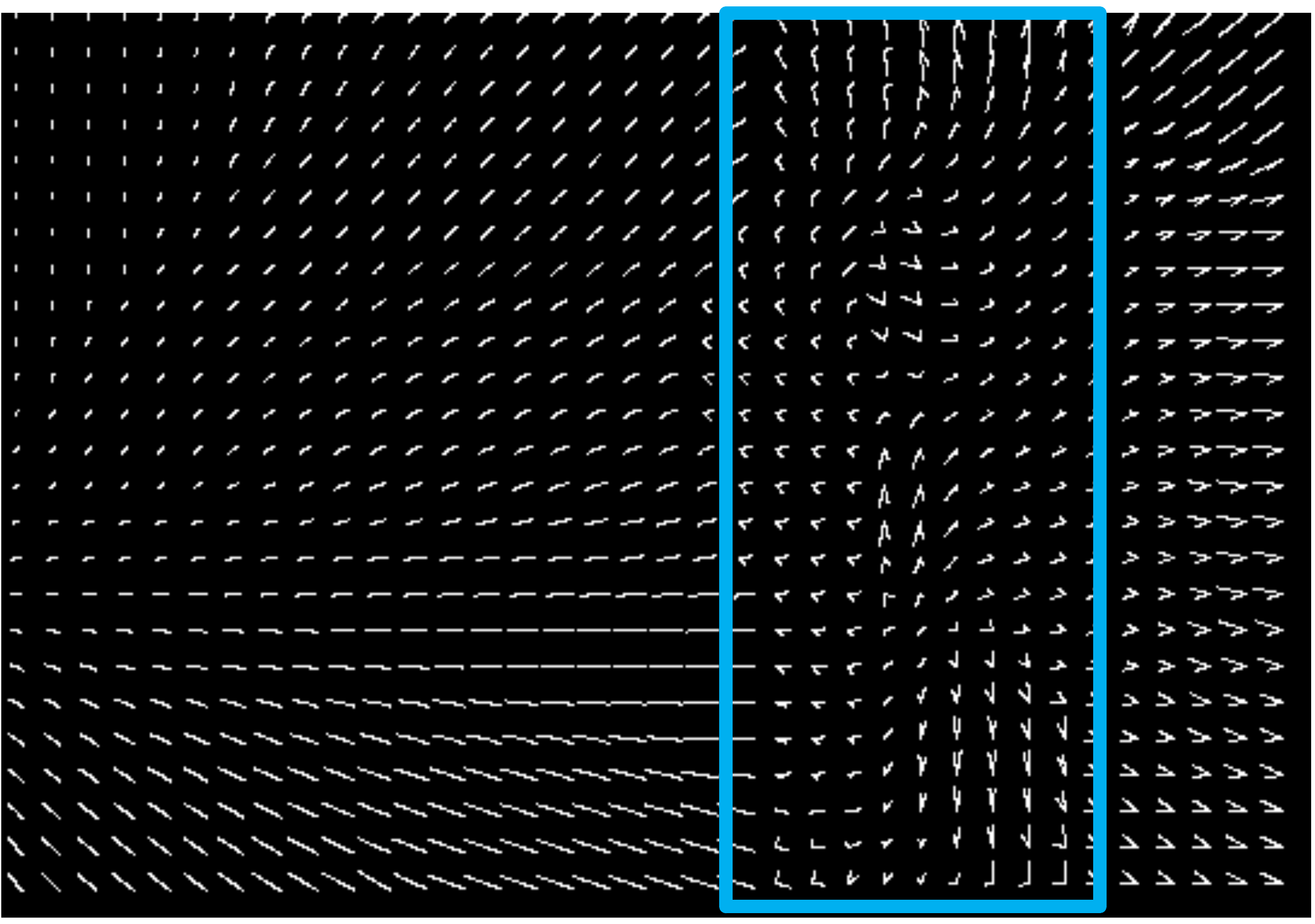}\\
\hspace{-0.2cm}
(a) Original Blurred image & (b) Kim and Lee~\cite{hyun2015generalized} \\
\hspace{-0.2cm}
\includegraphics[width=0.225\textwidth]{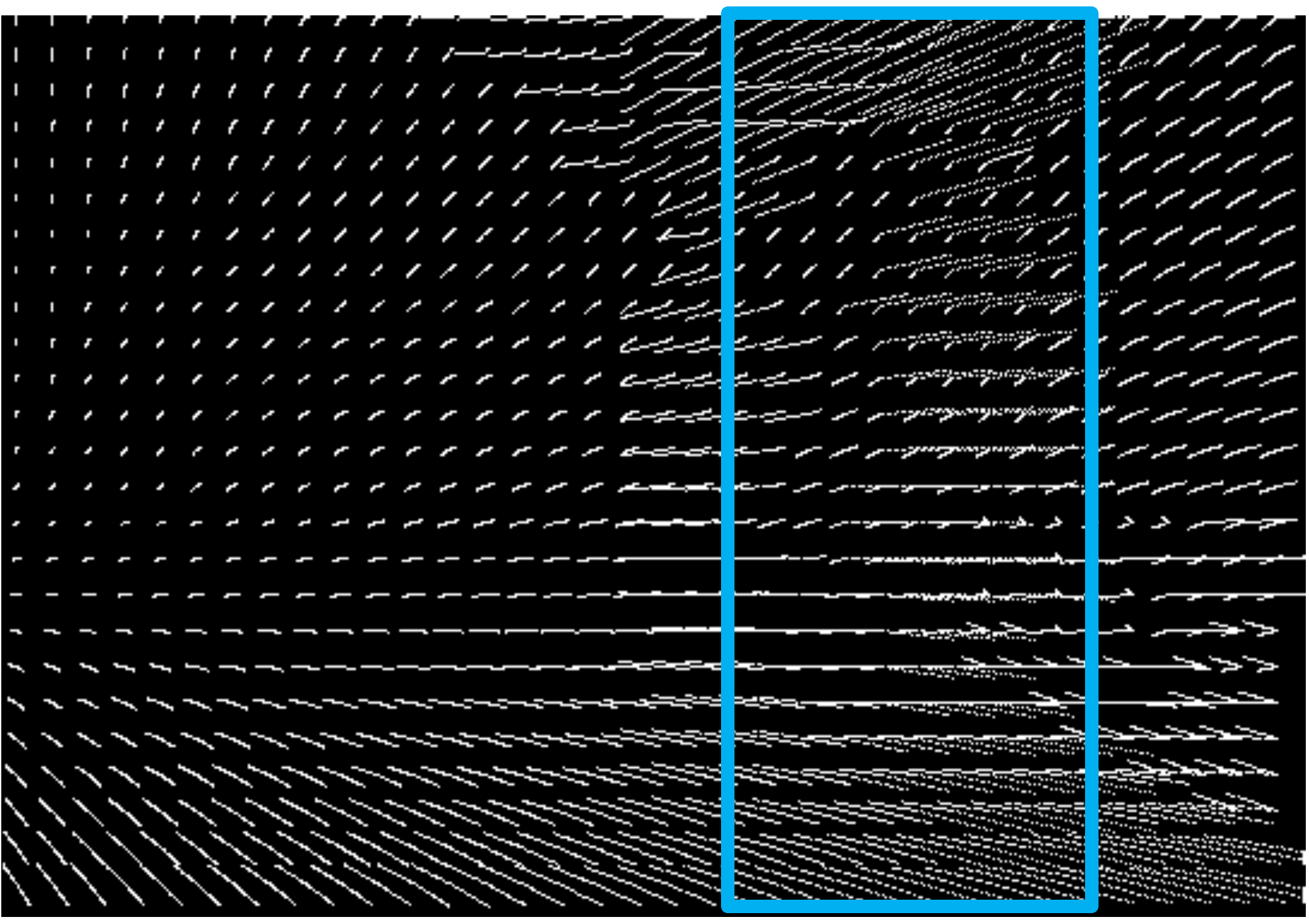}
&\includegraphics[width=0.225\textwidth]{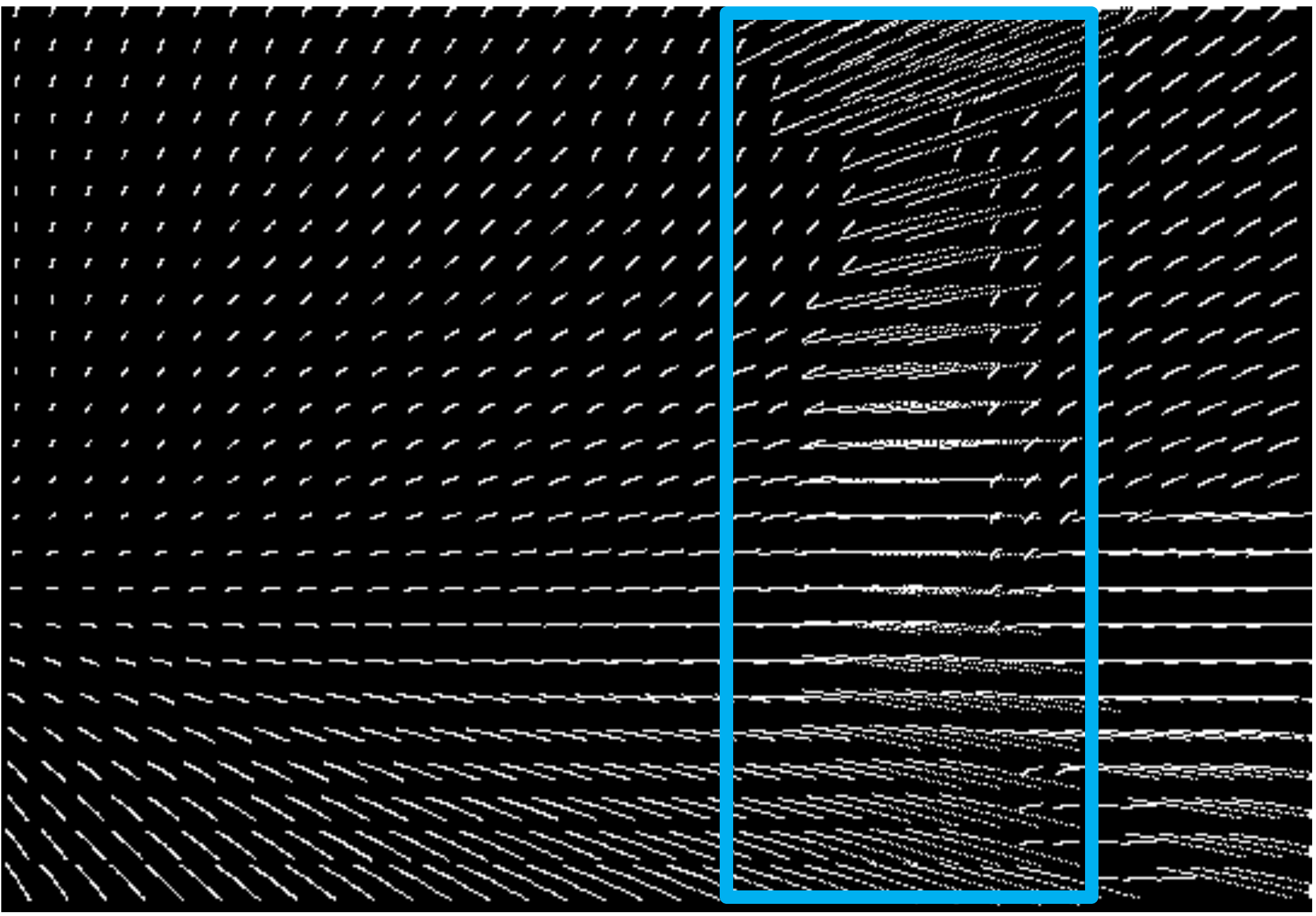}\\
\hspace{-0.2cm}
(c) Sellent \etal~\cite{sellent2016stereo} & (d) Ours\\
\end{tabular}
\end{center}
% \vspace{-2 mm}
\caption{Blur kernel estimation for an outdoor scene. \rc{(a) Blurred reference image from {\bf BlurData-1}.} (b) Blur kernel estimation by Kim and Lee~\cite{hyun2015generalized}. (c) Blur kernel estimation by Sellent \etal~\cite{sellent2016stereo}. (d) Our blur kernel estimation. Compared with these monocular and stereo deblurring methods, our method achieves more accurate blur kernel estimation.}
\label{fig:flowkernel}
\end{figure}

Image deblurring (even under stereo configuration) is generally an ill-posed problem, thus certain assumptions or additional constraints are required to regularize the solution space. Numerous methods have been proposed to address the problem~\cite{hyun2015generalized, sellent2016stereo, Pan_2017_CVPR, Li_2018_CVPR, hu2014joint, sun2015learning, pan2016soft, ren2017video, gong2017blur2mf}. As per the system configuration, the methods can be roughly categorized into two groups: monocular based approaches and binocular or multi-view based approaches. \rc{We also briefly discuss recent efforts in deep learning-based deblurring, Moving object segmentation, semantic segmentation, and scene flow estimation.} 

%%%%%%%%%%%%%%%%%%%% fig 3 %%%%%%%%%%%%%%%%%%%%%
\begin{figure*}
\begin{center}
\begin{tabular}{cccccc}
\hspace{-0.4cm}
\includegraphics[width=0.165\textwidth]{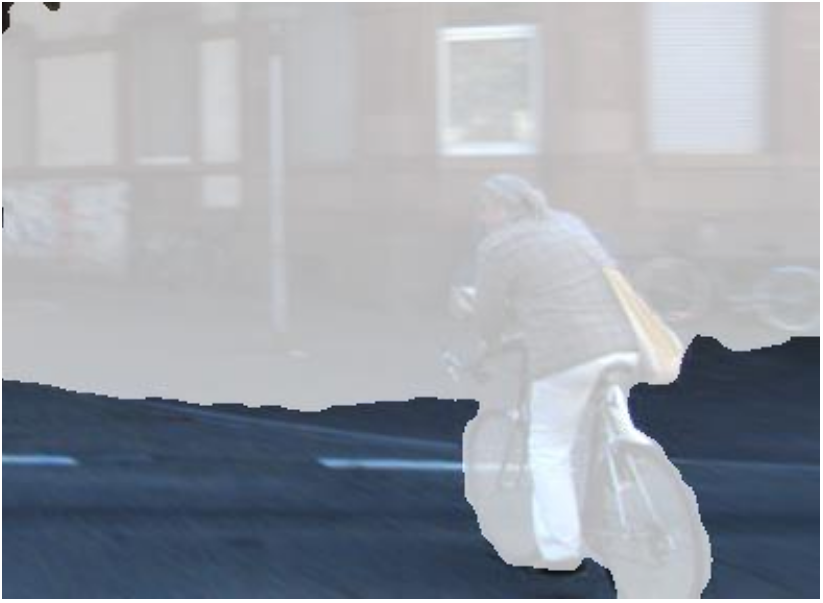}
&\hspace{-0.5cm}
\includegraphics[width=0.165\textwidth]{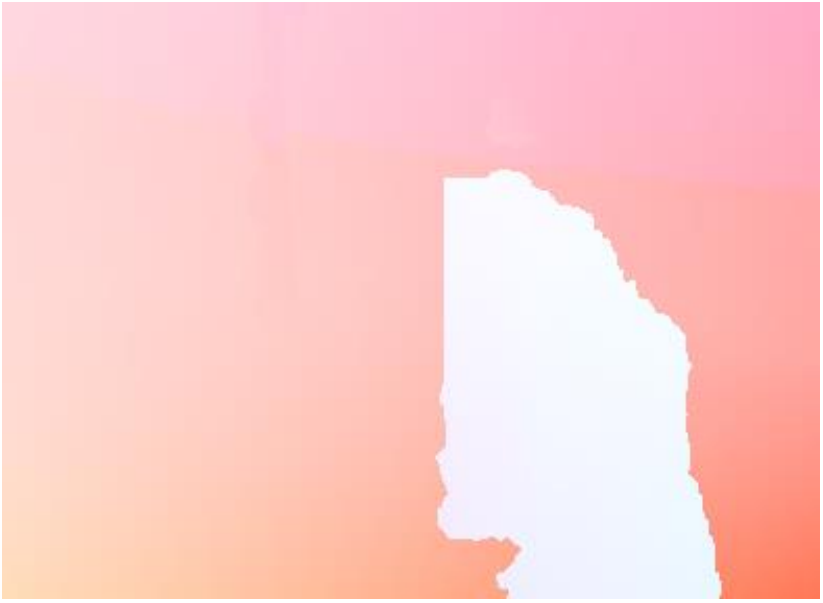}
&\hspace{-0.5cm}
\includegraphics[width=0.165\textwidth]{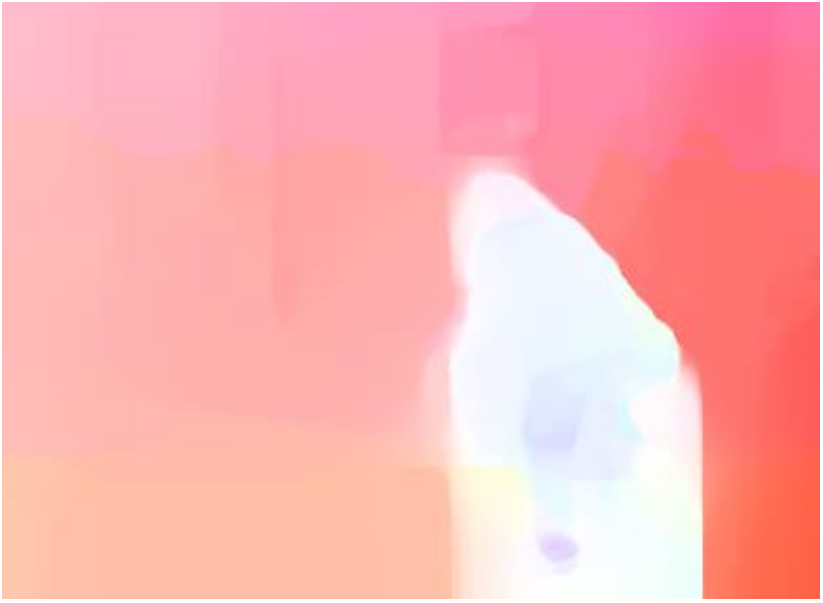}
&\hspace{-0.5cm}
\includegraphics[width=0.165\textwidth]{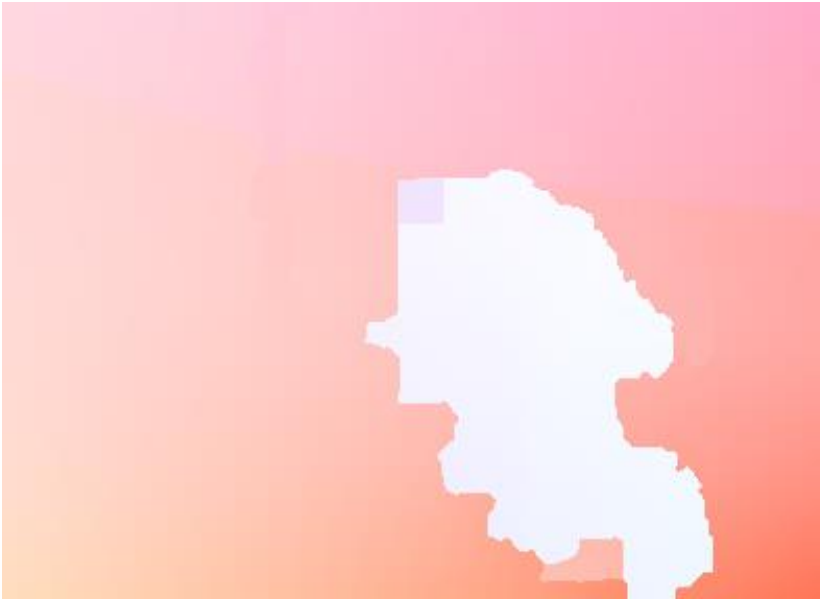}
&\hspace{-0.5cm}
\includegraphics[width=0.165\textwidth]{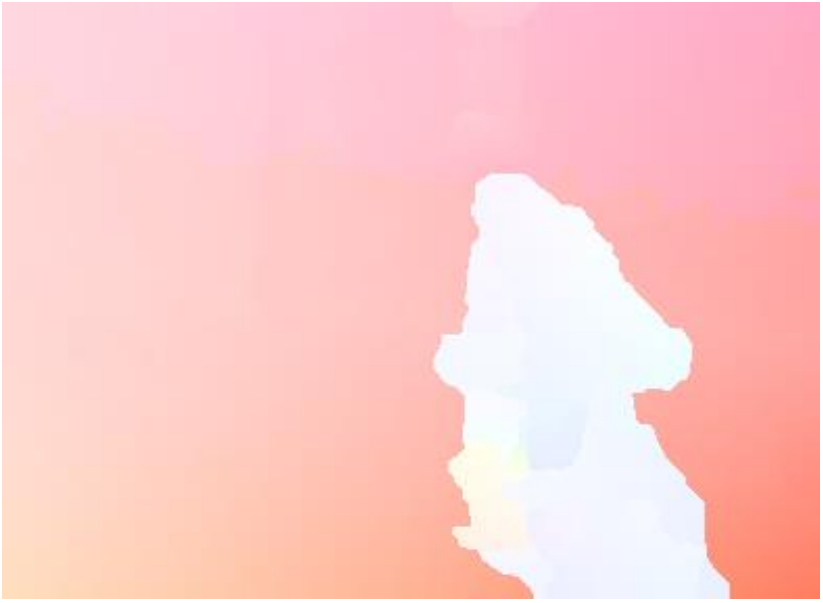}
&\hspace{-0.5cm}
\includegraphics[width=0.165\textwidth]{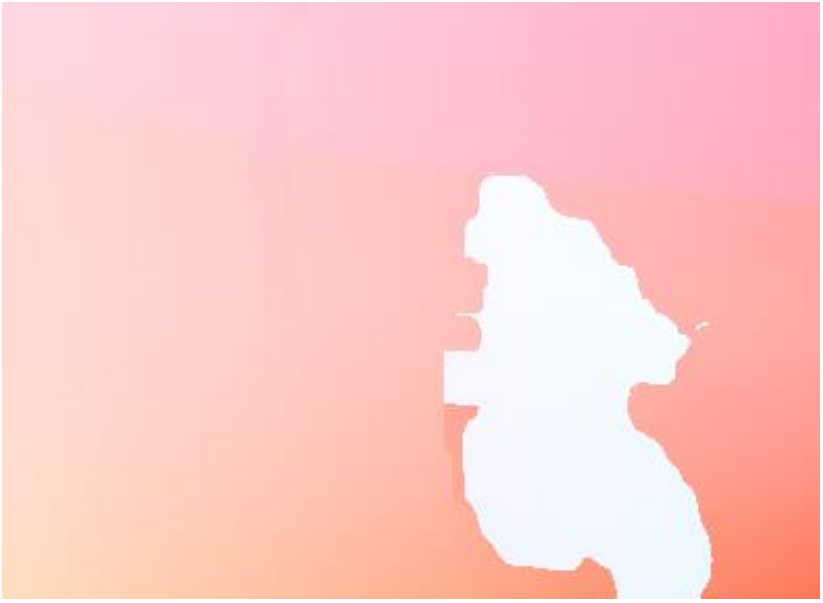}\\

\hspace{-0.4cm}  (a) Initial mask
&\hspace{-0.5cm} (b) Menze~\etal~\cite{menze2015object}
&\hspace{-0.5cm} (c) Kim and Lee~\cite{hyun2015generalized}
&\hspace{-0.5cm} (d) Ours (no seg)
&\hspace{-0.5cm} (e) Ours (one layer)
&\hspace{-0.5cm} (f) Ours (two layer) \\
\hspace{-0.4cm}
\includegraphics[width=0.165\textwidth]{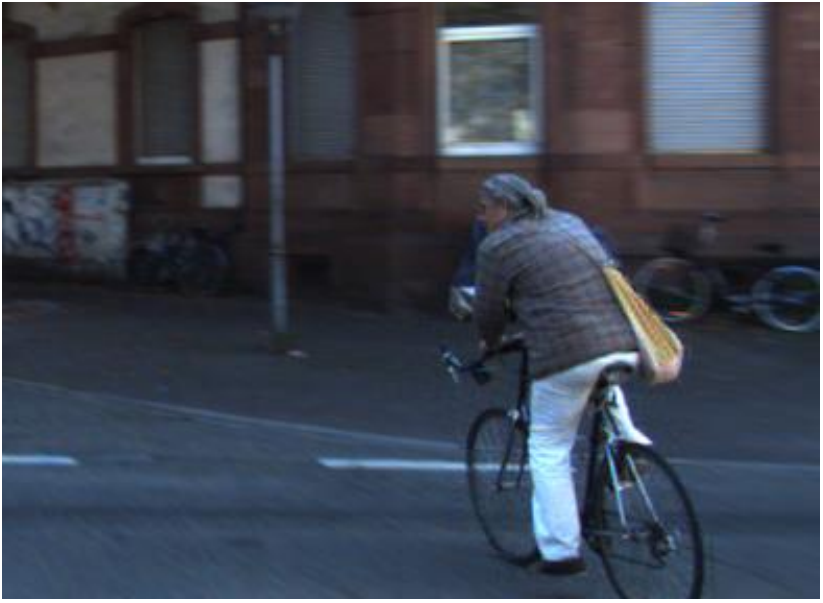}
&\hspace{-0.5cm}
\includegraphics[width=0.165\textwidth]{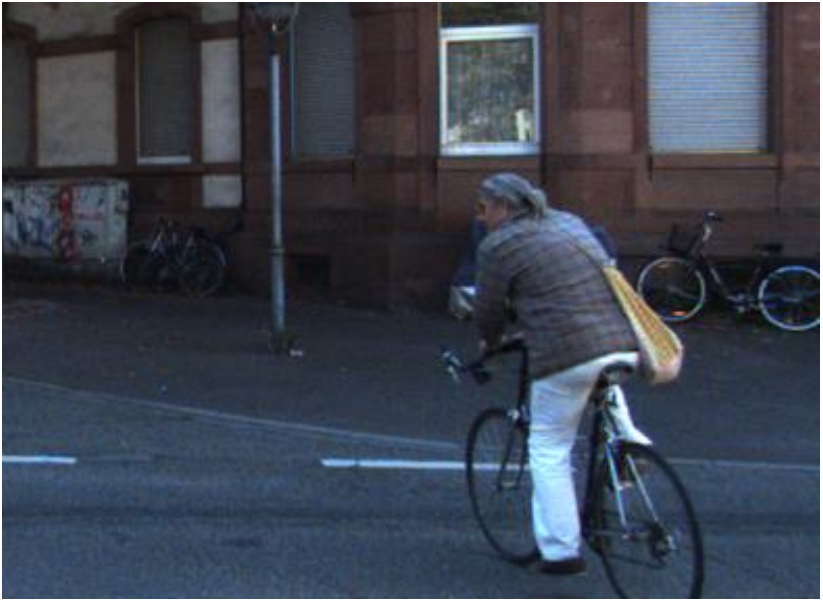}
&\hspace{-0.5cm}
\includegraphics[width=0.165\textwidth]{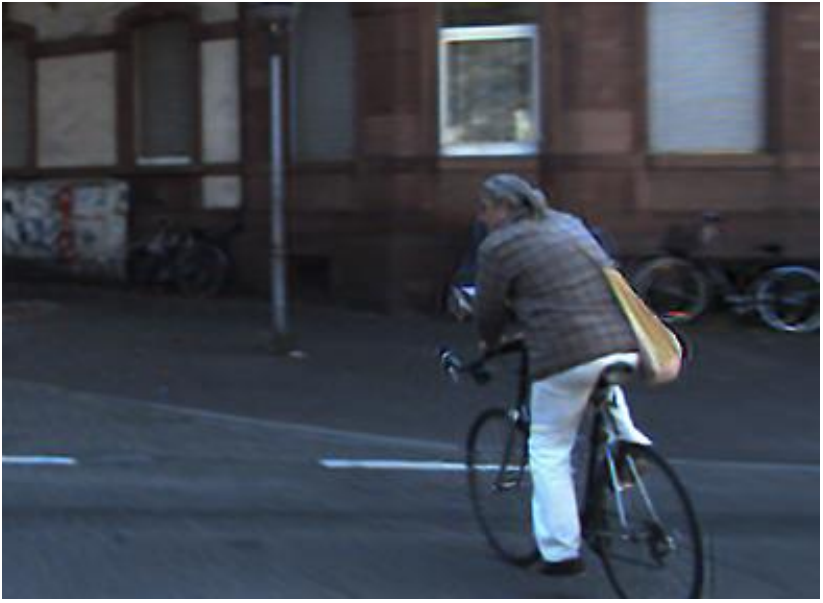}
&\hspace{-0.5cm}
\includegraphics[width=0.165\textwidth]{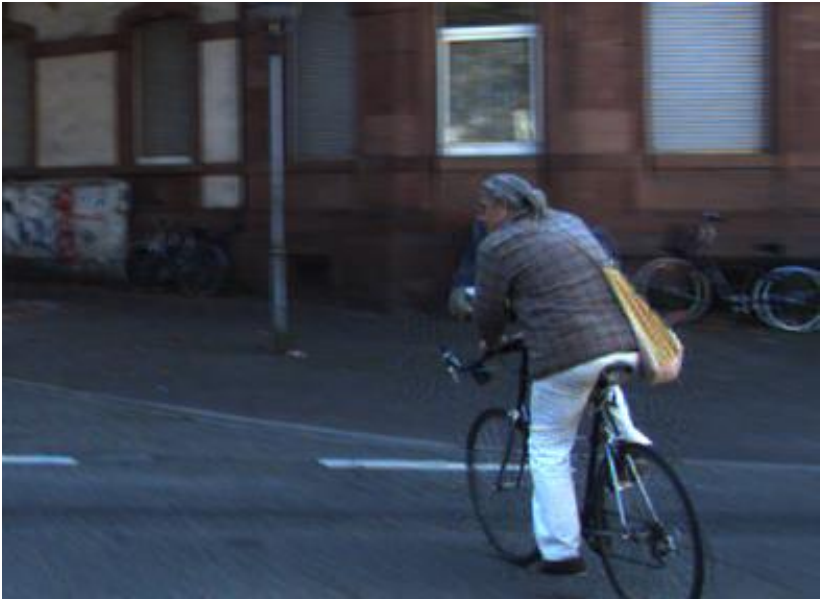}
&\hspace{-0.5cm}
\includegraphics[width=0.165\textwidth]{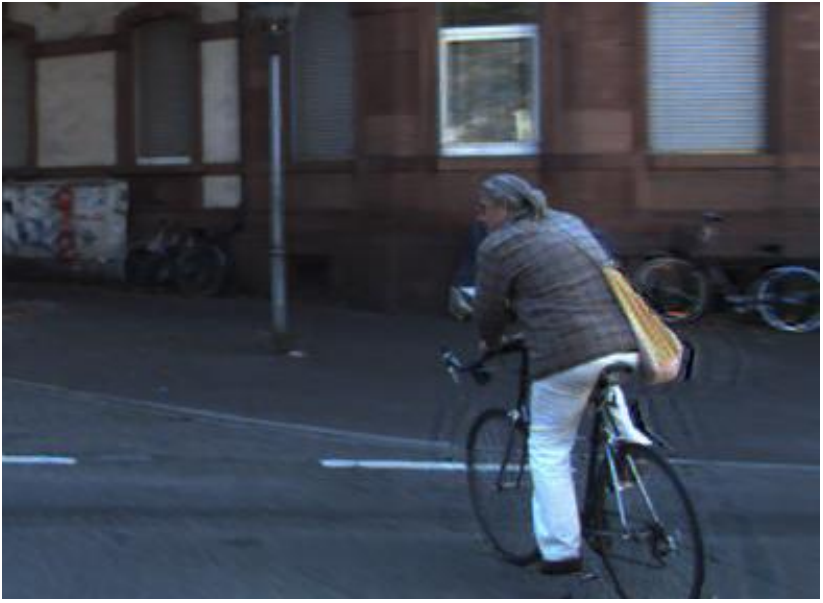}
&\hspace{-0.5cm}
\includegraphics[width=0.165\textwidth]{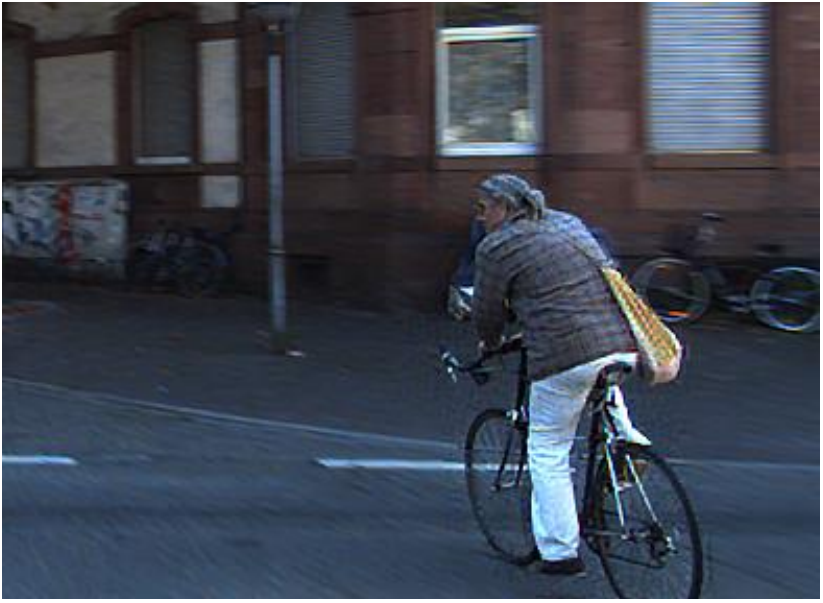}\\

\hspace{-0.4cm}  (g) Blurred image
&\hspace{-0.5cm} (h) Ground-truth
&\hspace{-0.5cm} (i) Kim and Lee~\cite{hyun2015generalized}
&\hspace{-0.5cm} (j) Sellent~\etal~\cite{sellent2016stereo}
&\hspace{-0.5cm} (k) Ours (one layer)
&\hspace{-0.5cm} (l) Ours (two layer) \\

\end{tabular}
\end{center}
% \vspace{-2 mm}
\caption{  
Scene flow results for an outdoor scenario. (a) and (g) The initial segmentation and blurred reference image from {\bf BlurData-1}. (b) Estimated flow by~\cite{menze2015object}. (c) Estimated flow by~\cite{hyun2015generalized}. (d)-(f) Our flow estimation result. (d) Without semantic segmentation. (e) With semantic segmentation, one layer StereoSLIC. (f) With semantic segmentation, two layer StereoSLIC. (h) The ground-truth latent image. (i) Deblurred result by~\cite{hyun2015generalized}. (j) Deblurred result by~\cite{sellent2016stereo}. (k) and (l) Our deblurred result. (k) Without semantic segmentation. (l) With semantic segmentation. The results show that, our two layer StereoSLIC could preserve edge information. Compared with both these state-of-the-art methods, our method achieves competitive performance. Best viewed in colour on the screen. %\PP{need to change (j)}
}
\label{fig:3methodevolution}
\end{figure*}

\vspace{-2 mm}
\subsection{Single view deblur}
Monocular based deblurring approaches often assume that the captured scene is static or has uniform blur kernel~\cite{gupta2010single}, or need user interaction~\cite{pan2016soft}.
A series of widely-used priors and regularizers are based on image gradient sparsity, such as the total variational regularizer~\cite{perrone2014total}, the Gaussian scale mixture prior~\cite{fergus2006removing}, the $l_1 \backslash l_2$ norm based prior~\cite{krishnan2011blind}, and the ${l}_0$-norm regularize~\cite{xu2013unnatural,pan2014deblurring}. Non-gradient-based priors have also been proposed, such as the edge-based patch prior~\cite{sun2013edge}, the colour line based prior~\cite{lai2015blur}, and the dark/white channel prior~\cite{pan2016blind, yan2017image}.
Hu \etal~\cite{hu2014joint} proposed to jointly estimate the depth layering and remove non-uniform blur caused by the in-plane motion from a single blurred image. While this unified framework is promising, user input for depth layers partition is required, and potential depth values should be known in advance.
Pan \etal~\cite{pan2016soft} proposed an algorithm to jointly estimate object segmentation and camera motion by incorporating soft segmentation, but require user input. In practical settings, it is still challenging to remove strongly non-uniform motion blur captured in complex scenes.

Since blur parameters and a latent image are difficult to be estimated from a single image, the monocular based approaches are extended to video to remove blurs in dynamic scenes.
In the work of Wulff and Black~\cite{wulff2014modeling}, a layered model is proposed to estimate the different motions of both foreground and background layers.
Kim and Lee~\cite{kim2014segmentation} proposed a method based on a local linear motion without segmentation. This method incorporates optical flow estimation to guide the blur kernel estimation and is able to deal with certain object motion blur. In~\cite{hyun2015generalized}, a new method is proposed to simultaneously estimate optical flow and tackle the case of general blur by minimization a single non-convex energy function. Park \etal~\cite{park2017joint} estimate camera poses and scene structures from severely blurred images and deblurring by using the motion information.

\vspace{-2 mm}
\subsection{Multi-view deblur}
As depth factor can significantly simplify the deblurring problem, multi-view deblurring methods have been proposed to leverage available depth information. Ezra and Nayar~\cite{nayar2004motion} proposed a hybrid imaging system, where a high-resolution camera captures the blurred frame and a low-resolution camera with faster shutter speed is used to estimate the camera motion. 
Xu \etal~\cite{xu2012depth} inferred depth from two blurred images captured by a stereo camera and proposed a hierarchical estimation framework to remove motion blur caused by the in-plane translation.
Sellent \etal~\cite{sellent2016stereo} proposed a video deblurring technique based on a stereo video, where 3D scene flow is estimated from the blurred images using a piecewise rigid 3D scene representation. Along the same line, Ren \etal~\cite{ren2017video} proposed an algorithm where accurate semantic segmentation is known. In their work, they also used the pixel-wise non-linear kernel model to approximate motion trajectories in the video. While the performance of their experiments shows limited effective for images which included multiple types of moving objects. We~\cite{Pan_2017_CVPR} proposed a single framework to jointly estimate the scene flow and deblur the images in CVPR 2017, where the motion cues from scene flow estimation and blur information could reinforce each other. These two methods represent the state-of-the-art in multi-view video deblurring and will be used for comparisons in the experimental section.

\vspace{-2 mm}
\subsection{Deep learning based deblurring methods}
Recently, deep learning-based methods have been used to restore clean latent images. Gong \etal~\cite{gong2017blur2mf} estimated flow from a single blurred image caused by camera motion through a fully convolutional deep neural network and recovered a clean image from the estimated flow. Su \etal~\cite{Su_2017_CVPR} introduced a deep learning solution to video deblurring, where a CNN is trained end-to-end to learn how to accumulate information across frames. However, they aimed to tackle motion blur from camera shake. Nah \etal~\cite{Nah_2017_CVPR} proposed a multi-scale convolutional neural network that restores latent images in an end-to-end manner without assuming any restricted blur kernel model. Kim~\etal~\cite{kim2017online,kim2016dynamic} proposed a novel network layer that enforces temporal consistency between consecutive frames by dynamic temporal blending which compares and adaptively shares features obtained at different time steps. 
%This deep learning method uses pixel-wise average kernel model in~\cite{kim2016dynamic} to generate blurred image dataset. 
Kupyn~\etal~\cite{Kupyn_2018_CVPR} presented an end-to-end learning approach for motion deblurring. The model they used is Conditional Wasserstein GAN with gradient penalty and perceptual loss based on VGG-19 activations. Tao~\etal~\cite{Tao_2018_CVPR} propose a light and compact network, SRN-DeblurNet, to deblur the image.
\rc{Jin~\etal~\cite{Jin_2018_CVPR} proposed to restore a video with fixed length from a single blurred image.}
However, deep deblurring methods generally need a large dataset to train the model and usually require sharp images provided as supervision. In practice, blurred images do not always have corresponding ground-truth sharp images.

\vspace{-2 mm}
\subsection{Moving object segmentation}
According to the level of supervision required, video segmentation techniques can be broadly categorized as unsupervised, semi-supervised and supervised methods.
Unsupervised methods~\cite{papazoglou2013fast} use a rapid technique to produce a rough estimate of which pixels are inside the object based on motion boundaries in pairs of subsequent frames. Then automatically bootstraps an appearance model based on the initial foreground estimate, and uses it to refine the spatial accuracy of the segmentation and to also segment the object in frames where it does not move. The works~\cite{faktor2014video, wang2015saliency, wang2017unified} extend the concept of salient objects detection~\cite{sundaram2010dense} as prior knowledge to infer the objects.
Semi-supervised video segmentation, which also refers to label propagation, is usually achieved via propagating human annotation specified on one or a few key-frames onto the entire video sequence~\cite{hariharan2015hypercolumns, shankar2015video, tsai2016video}. 
The idea of combining the best from both CNN model and MRF/CRF model is not new. A video object segmentation method by Jang and Kim~\cite{jang2017online} performs MRF optimization to fuse the outputs of a triple-branch CNN. However, the loosely-coupled combination cannot fully exploit the strength of MRF/CRF models. 
Supervised methods require tedious user interaction and iterative human corrections. \rc{These methods can attain high-quality boundaries while needing human supervision~\cite{wang2014touchcut, fan2015jumpcut}.}
Yan~\cite{yan2017weakly} proposed a multi-task ranking model for the higher-level weakly-supervised actor-action segmentation task.

\vspace{-2 mm}
\subsection{Semantic segmentation}
Another crucial factor to compute latent clean image is detecting moving objects boundaries. A general problem is that the object boundaries with mixed foreground and background pixels can lead to severe ringing artifacts. Semantic segmentation can help to provide objects information as initialization. He \etal~\cite{he2016deep} proposed the ResNets to combat the vanishing gradient problem in training very deep convolutional networks.~\cite{wu2016wider} obtain the semantic segmentation masks with the ResNet-38 network. Lin \etal~\cite{Lin_2017_CVPR} present RefineNet with multi-resolution fusion (MRF) to combine features at different levels, chained residual pooling (CRP) to capture background context, and residual convolutional units (RCUs) to improve end-to-end learning. Tsai \etal~\cite{tsai2016semantic} first generated the object-like tracklets and then adopted a sub-modular function to integrate object appearances, shapes and motions to co-select tracklets that belong to the common objects. Taking one step further, the Deep Parsing Network (DPN)~\cite{liu2018deep} is designed to approximate the mean-field inference for MRFs in one pass.

\vspace{-2 mm}
\subsection{\rc{Optical flow estimation}}
\rcs{
Menze \etal~\cite{menze2015object} proposed a novel model and dataset for 3D scene flow estimation with an application to autonomous driving. Pan \etal~\cite{Pan_2017_CVPR} proposed a single framework to jointly estimate the scene flow and deblur the images. 
Taniai \etal~\cite{Taniai_2017_CVPR} presented a multi-frame method for efficiently computing scene flow (dense depth and optical flow) and camera ego-motion for a dynamic scene observed from a moving stereo camera rig.
Yin \etal~\cite{Yin_2018_CVPR} proposed an unsupervised learning framework GeoNet for jointly estimating monocular depth, optical flow and camera motion from video.
Gong \etal~\cite{gong2017blur2mf} directly estimates the motion flow from the blurred image through a fully-convolutional deep neural network (FCN) and recovers the unblurred image from the estimated motion flow. 
%This is the first universal end-to-end mapping from the blurred image to the dense motion flow.
PWC-Net~\cite{Sun_2018_CVPR} uses the current optical flow estimate to warp the CNN features of the second image. It then uses the warped features and features of the first image to construct a cost volume, which is processed by a CNN to estimate the optical flow.
The FlowNet by Dosovitskiy \etal~\cite{dosovitskiy2015flownet} represented a paradigm shift in optical flow estimation. The work shows the feasibility of directly estimating optical flow from raw images using a generic U-Net CNN architecture. FlowNet 2.0~\cite{Ilg_2017_CVPR} develop a stacked architecture that includes warping of the second image with the intermediate optical flow which decreases the estimation error by more than 50\% than the original FlowNet.
}

%===========================================================
\section{Problem Formulation}\label{sec:model}
In this paper, we propose to solve the challenging and practical problem of stereo deblurring by using 
\rc{consecutive stereo image pairs of a calibrated camera}
%\YD{Do we use two or three? You illustrate with three somewhere, please confirm.} \PP{three}
in complex dynamic environments, where the blur is caused by the camera motion and the objects' motion. Under the problem setup, stereo deblurring and the scene flow estimation is already deeply coupled, \ie, stereo deblurring depends on the solution of the scene flow estimation while the scene flow estimation also needs the solution of stereo deblurring. \rcs{In addition}, with the multiple moving objects representation of the observed scene, Moving object segmentation also closely relates to both scene flow estimation and stereo deblurring, \ie, improper Moving object segmentation could result in dramatical changes in scene flow estimation and stereo deblurring especially along the object boundaries~\cite{Sevilla-Lara_2016_CVPR}. Therefore, we could conclude that the scene flow estimation, Moving object segmentation and video deblurring are deeply coupled under our problem setup.

To better exploit the deeply coupling nature of the problem, we propose to formulate our problem as a joint estimation of scene flow, Moving object segmentation and stereo image deblurring for complex dynamic scenes. In particular, we rely on the assumptions that the scene can be well approximated by a collection of 3D planes~\cite{yamaguchi2013robust} belonging to a finite number of objects \footnote{The background is regarded as a single 'object' due to the camera motion only.} performing rigid motions individually~\cite{menze2015object}. Therefore, the problem of scene flow estimation can be reformulated as the task of geometric and motion estimation for each 3D plane. The rigid motion is defined for each moving object, which naturally encodes the Moving object segmentation information. The blurred stereo images are generated due to the camera motion, multiple moving objects motion and the 3D scene structure, which are all characterized by the scene flow estimation and the Moving object segmentation. Specifically, our structured blur kernels are expressed with the geometry and motion of each 3D plane.

%%%%%%%%% figure pipeline %%%%%%%%%%%
\begin{figure}[t]
\begin{center}
\begin{tabular}{c}
\hspace{-0.3cm}
\includegraphics[width=0.45\textwidth]{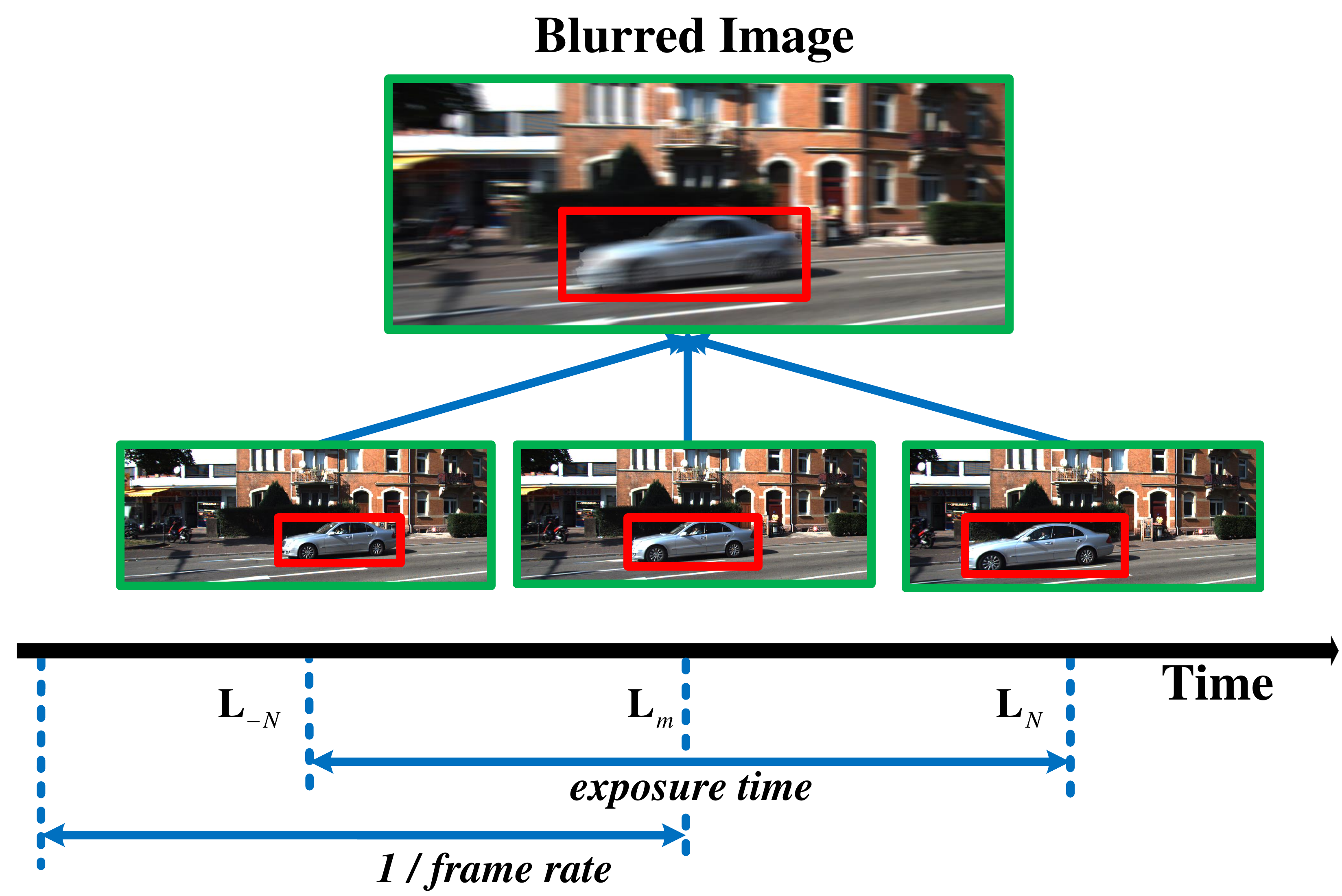}\\
\hspace{-0.3cm}
\includegraphics[width=0.43\textwidth]{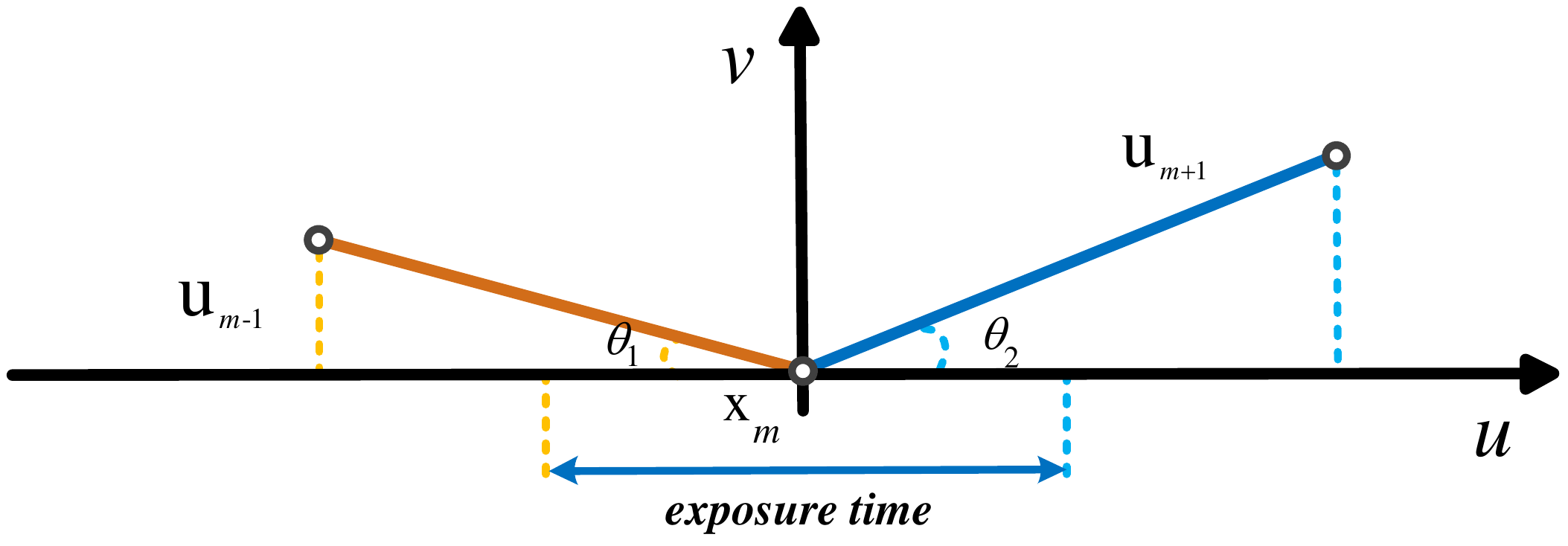}\\
\end{tabular}
\end{center}
% \vspace{-2 mm}
\caption{The pipeline of generating blurred images. We approximate the motion blur kernel as a piece-wise linear function based on bi-direction optical flows and generate blurred images by averaging consecutive frames whose relative motions between two neighbouring frames are known. Notably, ground truth sharp image is chosen to be the middle one.}
\label{fig:pipeline}
\end{figure}
% %%%%%%%%%%%%%%%%%%%% fig 4 %%%%%%%%%%%%%%%%%%%%%
\begin{figure*}
\begin{center}
\begin{tabular}{ccccccccc}
\hspace{-0.5cm}
%, height=1.23cm
\includegraphics[width=0.111\textwidth]{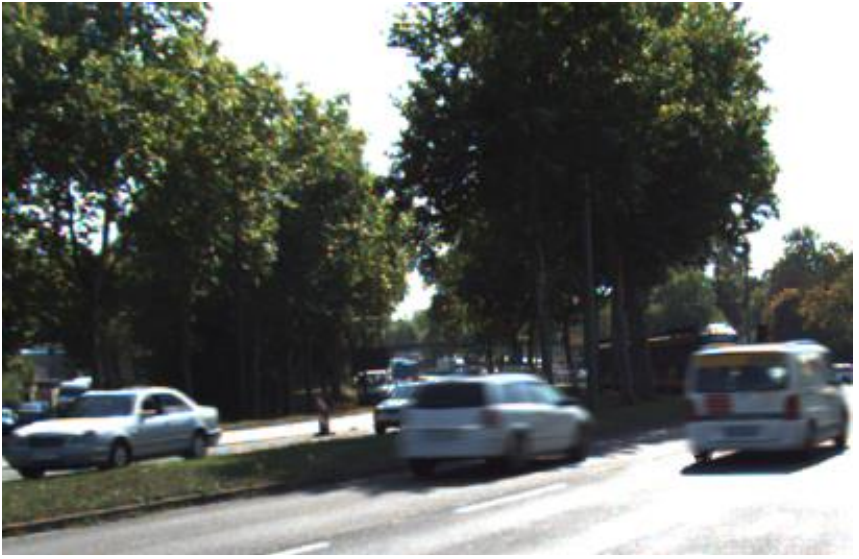}
&\hspace{-0.5cm}
\includegraphics[width=0.111\textwidth]{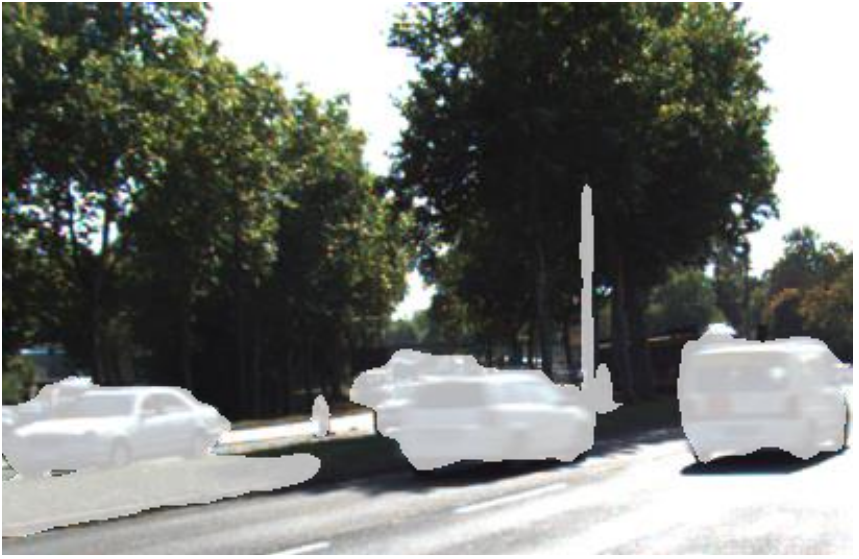}
&\hspace{-0.5cm}
\includegraphics[width=0.111\textwidth]{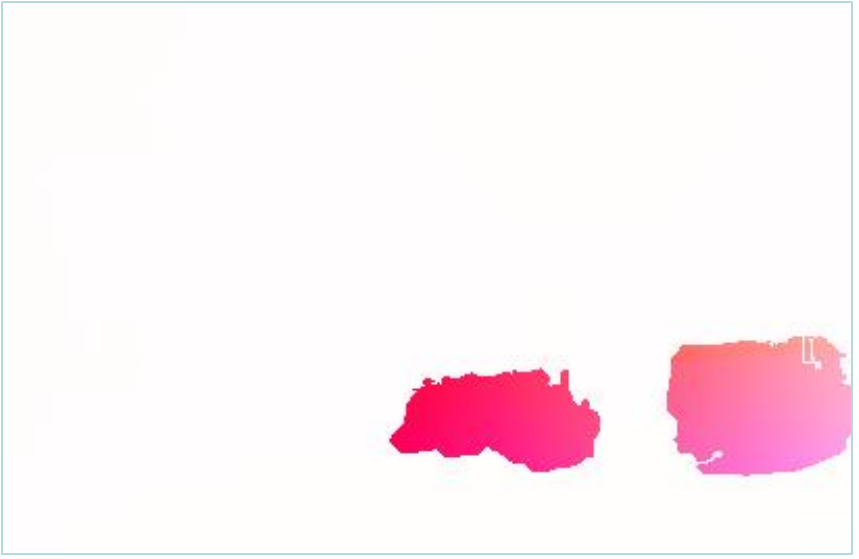}
&\hspace{-0.5cm}
\includegraphics[width=0.111\textwidth]{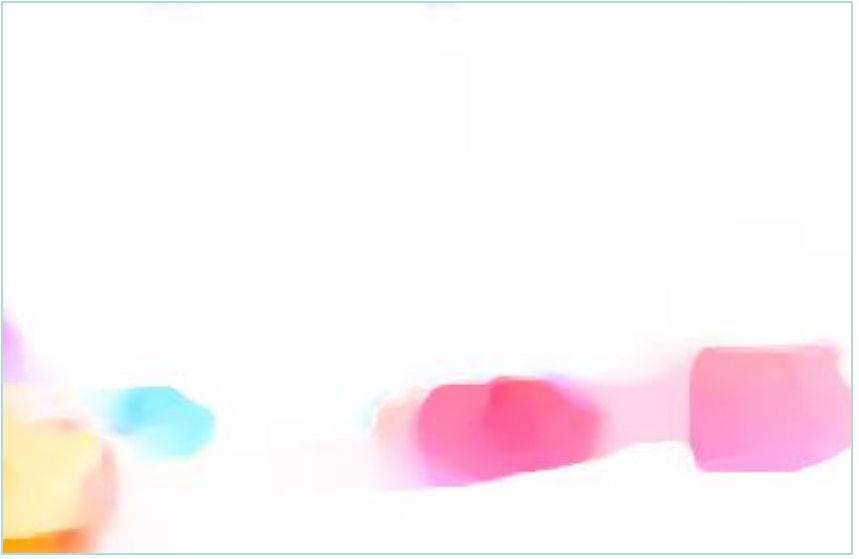}
&\hspace{-0.5cm}
\includegraphics[width=0.111\textwidth]{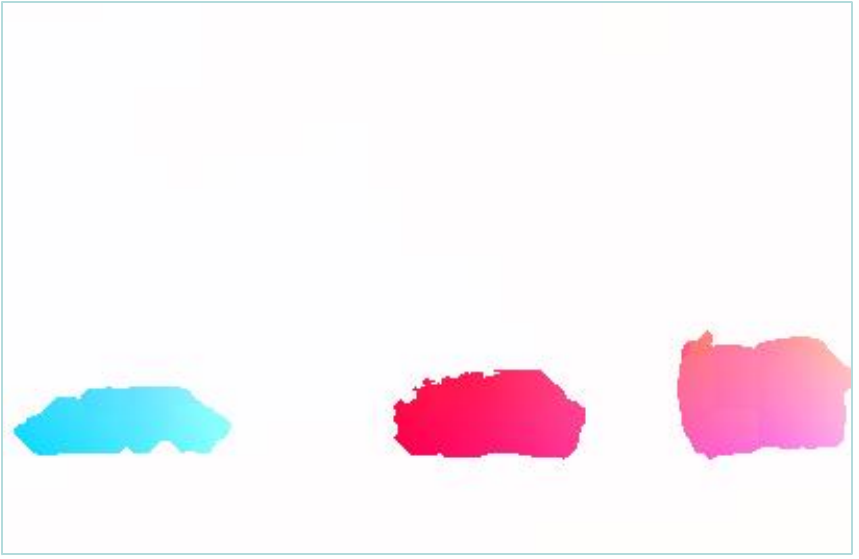}
&\hspace{-0.5cm}
\includegraphics[width=0.111\textwidth]{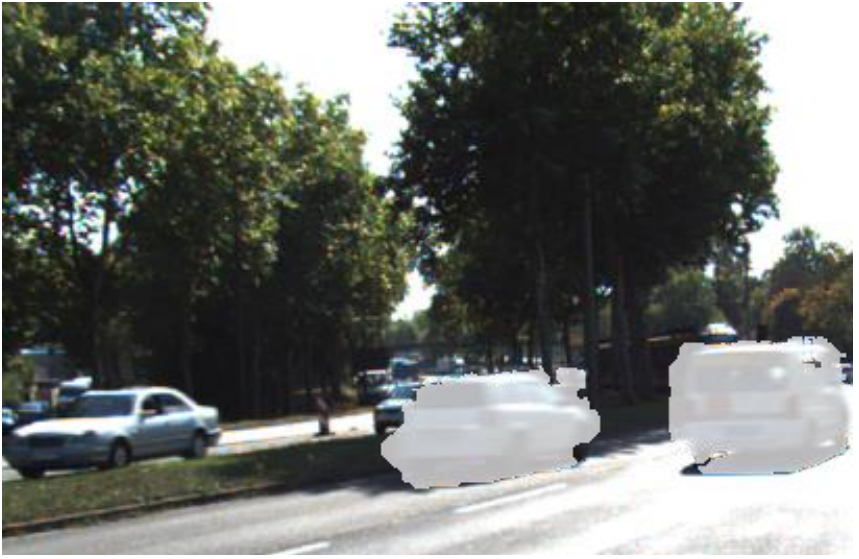}
&\hspace{-0.5cm}
\includegraphics[width=0.111\textwidth]{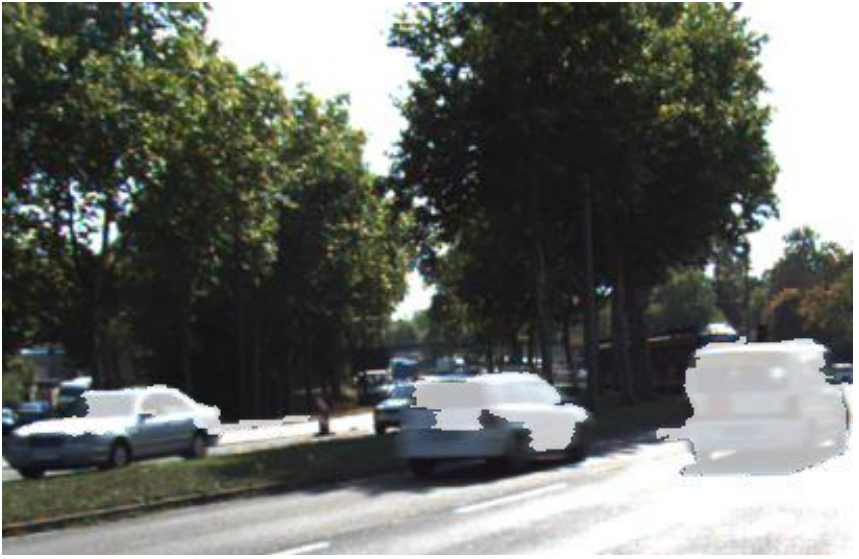}
&\hspace{-0.5cm}
\includegraphics[width=0.111\textwidth]{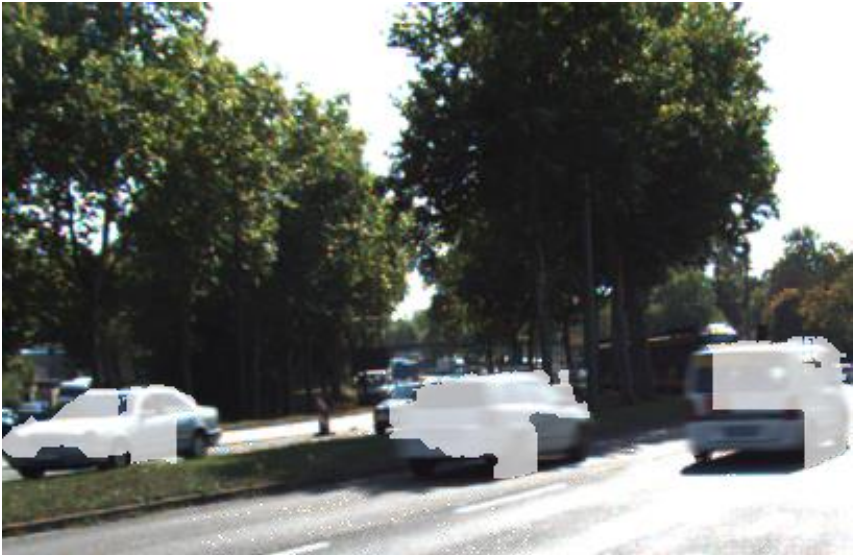}
&\hspace{-0.5cm}
\includegraphics[width=0.111\textwidth]{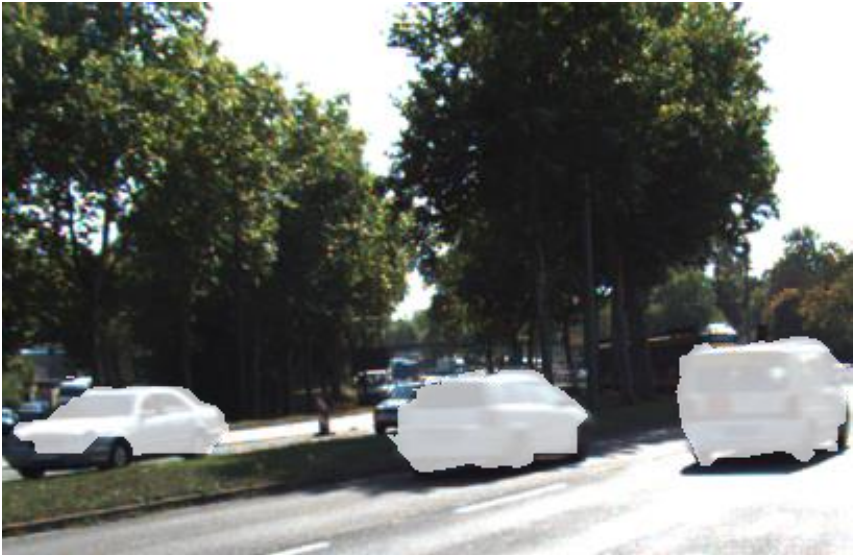}\\
\hspace{-0.5cm}
\includegraphics[width=0.111\textwidth]{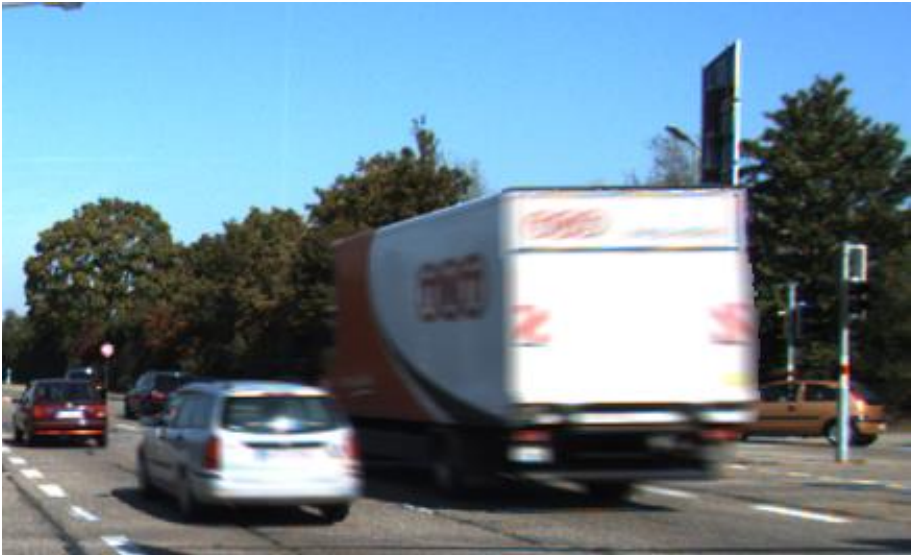}
&\hspace{-0.5cm}
\includegraphics[width=0.111\textwidth]{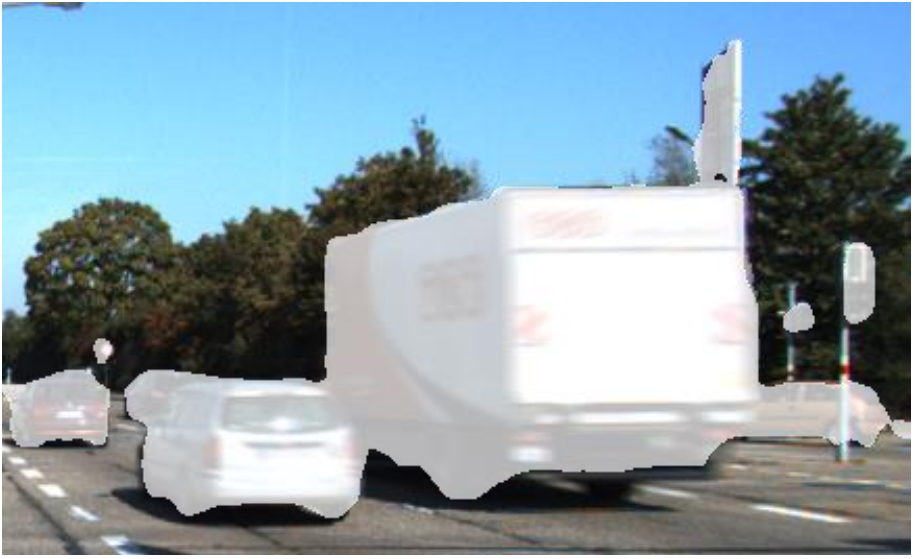}
&\hspace{-0.5cm}
\includegraphics[width=0.111\textwidth]{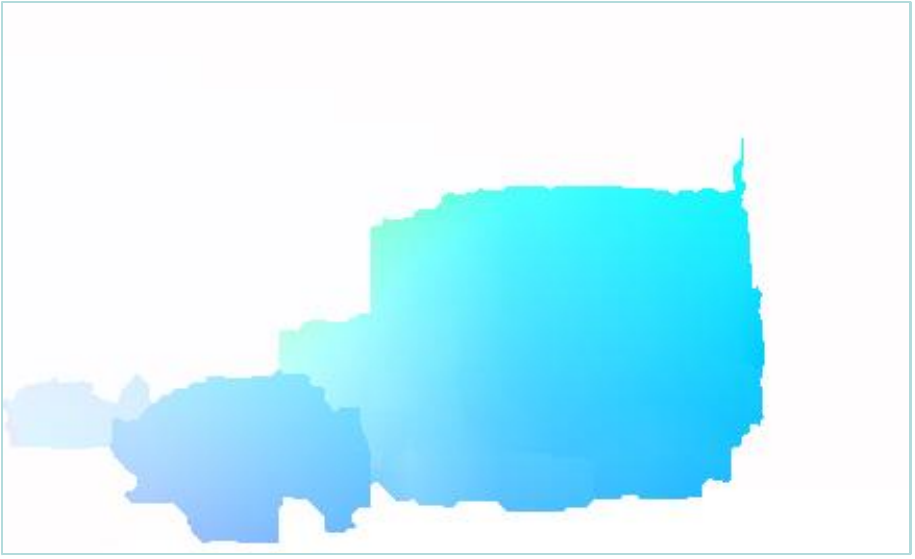}
&\hspace{-0.5cm}
\includegraphics[width=0.111\textwidth]{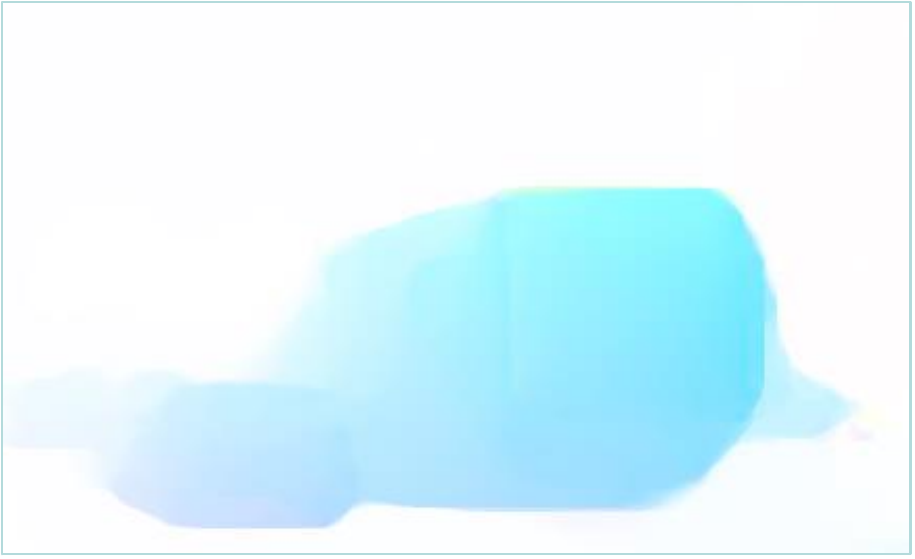}
&\hspace{-0.5cm}
\includegraphics[width=0.111\textwidth]{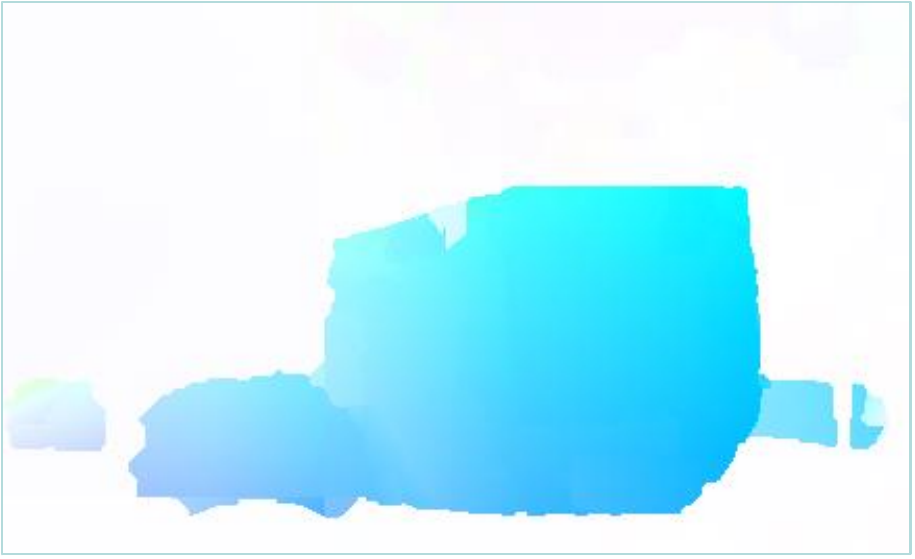}
&\hspace{-0.5cm}
\includegraphics[width=0.111\textwidth]{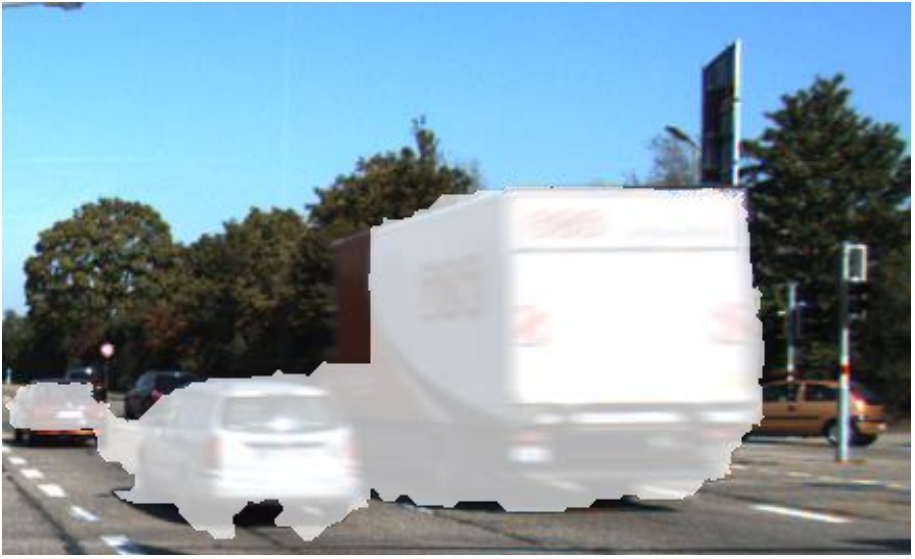}
&\hspace{-0.5cm}
\includegraphics[width=0.111\textwidth]{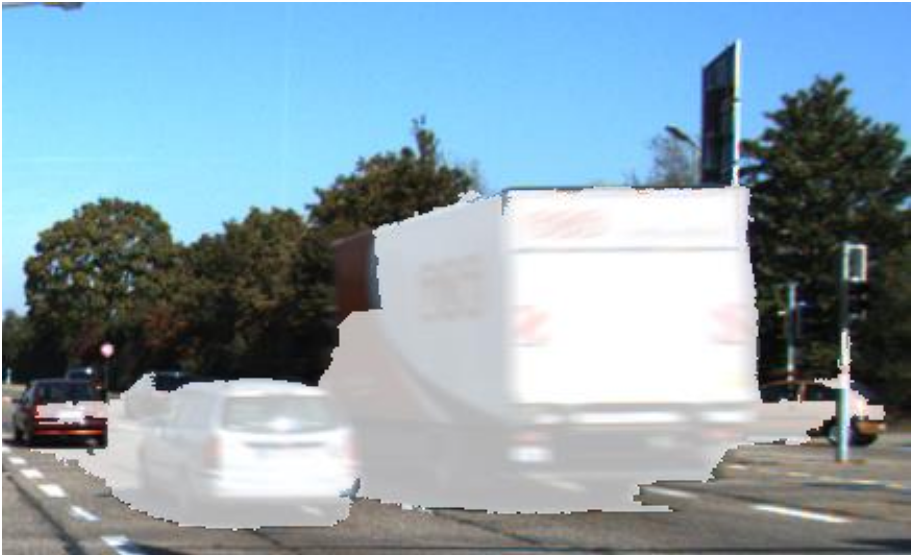}
&\hspace{-0.5cm}
\includegraphics[width=0.111\textwidth]{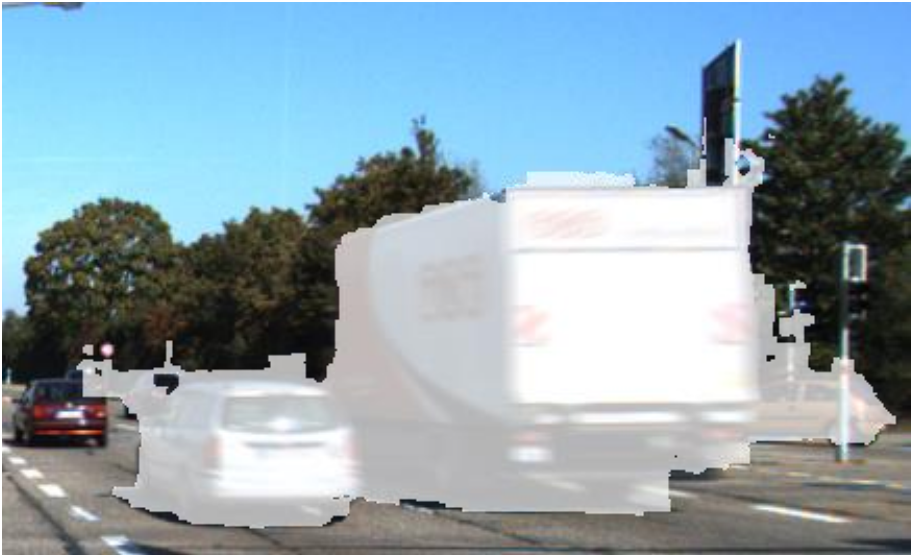}
&\hspace{-0.5cm}
\includegraphics[width=0.111\textwidth]{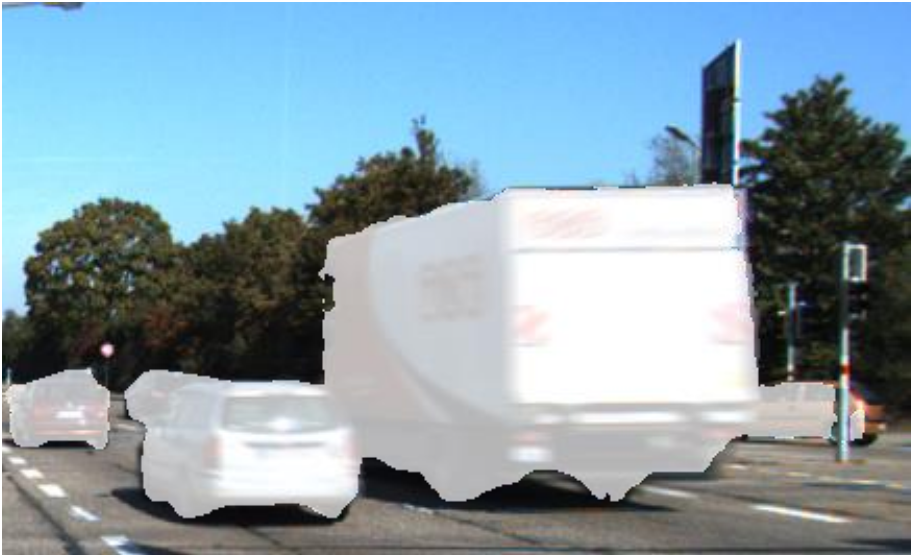}\\
\hspace{-0.5cm}
(a) Image
&\hspace{-0.5cm} (b) Segmentation
&\hspace{-0.5cm} (c) \cite{menze2015object}
&\hspace{-0.5cm} (d) \cite{hyun2015generalized}
&\hspace{-0.5cm} (e) Ours
&\hspace{-0.5cm} (f)  \cite{menze2015object}
&\hspace{-0.5cm} (g)  \cite{papazoglou2013fast}
&\hspace{-0.5cm} (h)  \cite{faktor2014video}
&\hspace{-0.5cm} (i) Ours\\
\end{tabular}
\end{center}
% \vspace{-2 mm}
\caption{
Scene flow and Moving object segmentation results for an outdoor scenario from {\bf BlurData-1}. (a) Input blurred image. (b) Input semantic segmentation. (c) Estimated flow by~\cite{menze2015object}. (d) Estimated flow by~\cite{hyun2015generalized}. (e) Our flow estimation result. (f) Segmentation result by~\cite{menze2015object}. (g) Segmentation result by~\cite{papazoglou2013fast}.  (h) Segmentation result by~\cite{faktor2014video}. (i) Our segmentation result. Compared with both these state-of-the-art methods, our method achieves competitive performance. Best viewed in colour on the screen.
}
\label{fig:4comparefs}
\end{figure*}
\subsection{Blurred Image Formation based on the Structured Pixel-wise Blur Kernel}
\label{sec:3.1}

Blurred images are formed by the integration of light intensity emitted from the dynamic scene over the aperture time interval of the camera. We assume that the blurred image $\vB$ can be generated by the integral of the latent high frame-rate image sequence $\{\vL_n\}$ during the exposure time. \rcs{This model follows by \cite{kim2014segmentation,gupta2010single,whyte2012non,dai2008motion}, which supposes the integration of light intensity happens in pixel colour space over the shutter time of the camera.}\footnote{\rcs{We notice that several methods model the integration in the raw sensor value and consider the effects of CRFs (camera response function) on motion deblurring. These yield a slightly different solution for deblurring~\cite{Nah_2017_CVPR,tai2013nonlinear}.}}
This defines the blurred image frame in the video sequence as
% {\small
\begin{equation} %\label{eq:convBlurKernel}
\mathbf{B}_m =\frac{1}{2N+1} \sum^{N}_{n=-N} \vL_n, 
\end{equation}
% }
\rc{where $\vB_m$ is the $m^{th}$ blurred image in the video sequence, ${\vL_n,\ n\in[-N,N]}$ denotes latent frames that generate the blurred image. The middle frame $\vL_m$ among the latent frames is defined as the deblurred image, which associated with $\vB_m$. This integration model has been widely used in the image/video deblurring literature~\cite{lee2013recent, kim2014segmentation, gong2017blur2mf}, which has also been used in~\cite{Nah_2017_CVPR, Su_2017_CVPR, kim2017online} to generate realistic blurred images from high frame-rate videos.
With optical flow, we can transform $\vL_n$ with $\vL_m$. Thus, the blur can be modelled by bi-directional optical flows. We approximate the kernel as piece-wise linear using bidirectional optical flows, where the kernel $\vA_m^\vx$ is spatially varying for each pixel.}
% {\small
\rc{
\begin{equation} \label{eq:convBlurKernel}
\mathbf{B}_m(\vx) = \mathrm{vec}(\vA_m^\vx)^{T} \mathrm{vec}(\vL_m), 
\end{equation}
}
% }
%
\rc{where $\vx \in \mathbb{R}^2$ denotes the pixel location in the image domain, $\mathrm{vec}$ denotes the vectorization operator,
%$\vL_n,\ n\in[-N,N]$ denotes un-exist neighbouring latent image of $\vL_m$. $\vL_m$ denotes the sharp image that associates with the blurred image $\vB_m$, which is also the middle frame among sequence $\vL_n$.
$\vA_m^\vx \in \mathbb{R}^{h \times w}$ is the blur kernel for each pixel $\vx$, where $h,\ w$ are the image size. In order to handle multiple types of blurs, we assumed that the blur kernel $\vA_m^\vx$ can be linearized in terms of a motion vector, which can be expressed as~\cite{kim2014segmentation}: }
%
% {\small
\begin{equation}
\begin{aligned} \label{eq:kimblurKernel}
\hspace{-0.1cm}& \vA_m^\vx(\tilde{u},\tilde{v})=\\
&\left\{
\begin{gathered}
\begin{aligned}
&\frac{\delta(\tilde{u}v_{m+1}-\tilde{v}u_{m+1})}{\tau||\mathbf{u}_{m+1}||},
&\;\rm{if}\;\tilde{\mathbf{u}}\in [\bf{0},\tau \mathbf{u}_{m+1}],\\
%\ and\ \frac{\tilde v}{\tilde u}=tan(\theta_1) \\
&\frac{\delta(\tilde{u}v_{m-1}-\tilde{v}u_{m-1})}{\tau||\mathbf{u}_{m-1}||},
&\;\rm{if}\;\tilde{\mathbf{u}}\in [\bf{0},\tau \mathbf{u}_{m-1}],\\
%\ and\ \frac{\tilde v}{\tilde u}=tan(\theta_2)\\
& \bf{0}, & \rm{otherwise},
\end{aligned}
\end{gathered}\right.
\end{aligned}
\end{equation}
% }
%
where $\tau$ = $\frac{1}{2}\ \times $ exposure time $\times$ frame rate, \rc{$\delta$ denotes the Kronecker delta function,}
$\mathbf{u}_{m+1}$ and $\mathbf{u}_{m-1}$ are the bidirectional optical flows at frame $m$. In particular, ~$\tilde{{\bf u}} = (\tilde{u},\tilde{v})$ which denotes the motion between exposure time, the kernel model is shown in Fig.~\ref{fig:pipeline}. \rc{We obtain the blur kernel matrix $\vA_m\in \mathbb{R}^{(h \times w)\times (h \times w)}$ by stacking $\mathrm{vec}(\vA_m^\vx )$ over the whole image domain. This leads to the blur model for the image as}
\rc{
\begin{equation}
    \mathrm{vec}(\vB_m) = \vA_m \mathrm{vec}(\vL_m).
\end{equation}
}
\rc{We omit the vectorize symbol in the following sections. We can cast the kernel estimation problem as a motion estimation problem.}

In our setup, the stereo video provides the depth information for each frame. Based on our piece-wise planar assumptions on the scene structure, optical flows for pixels lying on the same plane are constrained by a single homography. In particular, we represent the scene in terms of superpixels and finite number of objects with rigid motions. We denote $\calS$ and $\mathcal{O}$ as the set of superpixels and moving objects, respectively.
Each superpixel $i\in \calS$, is associated with a region  $\calS_i$ in the image, each region is denoted by a plane variable $\vn_{i,k_i} \in \mathbb{R}^3$ in 3D ($\vn_{i,k_i}^T\vx=1$ for $\vx \in \mathbb{R}^3$), where $k_i \in \left\{1,\cdots,|\mathcal{O}|\right\}$ denotes the $i^{th}$ superpixel associated with the $k^{th}$ object. Object inheriting its corresponding motion parameters $\vo_{k_i}=(\mathbf{R}_k, \mathbf{t}_k) \in \mathbb{SE}(3)$, where $\mathbf{R}_k \in \mathbb{R}^{3 \times 3}$ is the rotation matrix and $\mathbf{t}_k \in \mathbb{R}^{3}$ is the translation vector.
Note that $(\vn,\vo)$ encodes the scene flow information~\cite{menze2015object}, where $\vn = \{\vn_{\rc{i,k_i}}|i \in \calS\}$ and $\vo = \{\vo_{k_i}|k_i \in \mathcal{O}\}$. Given the motion parameters ${\bf o}_{k_i}$, we can obtain the homography defined by superpixel $i$ as
%,vogel2014view
% {\small
\begin{equation}
\mathbf{H}_{i} = {\bf K}(\mathbf{R}_k -\mathbf{t}_k \vn^{T}_{i,k_i}){\bf K}^{-1},
\end{equation}
% }
%
where ${\mathbf K} \in \mathbb{R}^{3 \times 3}$ is the camera calibration matrix. We note that, ${\bf H}_{i}$ relates corresponding pixels across two frames.

The optical flow is then defined as
\begin{equation}\label{eq:structuredflow}
\begin{aligned}
\mathbf{u}_{i} = \vx - \pi(\mathbf{H}_{i}\vx),
\end{aligned}
\end{equation}
%where ${\vx^*} = \mathbf{H}_{i}\vx$ is the coordinate of pixels in superpixel $i$.
where we denote $\vx^* = \pi(\mathbf{H}_{i}\vx)$. $\pi(\cdot)$ is the perspective division such that $\pi([x,y,z]^T) := [x/z, y/z]^T$. This shows that the optical flows for pixels in a superpixel are constrained by the same homography. Thus, it leads to a structured version of blur kernel defined in Eq.~(\ref{eq:kimblurKernel}).
%In Fig.~\ref{fig:flowkernel}, we compare our blur kernel with that from Kim~\cite{hyun2015generalized} and Sellent \etal~\cite{sellent2016stereo}. Our kernels are more structured, which also leads to more accurate scene flow estimation.

%%%%%%%%%%%%%%%%%%%%%%%%%%% Semantic segmentation %%%%%%%%%%%%%%%%%%%%%%%%%%%%%%
\subsection{Moving object segmentation}
\label{sec:3.2}
 
Semantic segmentation breaks the image into semantically consistent regions such as road, car, person, sky, \etc. Our algorithm computes each region independently based on the semantic class label, resulting in more precise Moving object segmentation and flow estimation, particularly at object boundaries. The provided additional information about object boundaries contributes to avoiding ringing and boundary artifact. 

A general problem in motion deblurring is that the moving object boundaries with mixed foreground and background pixels can lead to severe ringing artifacts (\rc{see Fig.~\ref{fig:fig1} for details}).
Most motion deblurring methods address this problem by segmenting blurred images into regions or layers where different kernels are estimated and applied for image restoration~\cite{tai2010correction, wulff2014modeling, pan2016soft}. Segmentation on blurred images is difficult due to ambiguous pixels between regions, but it plays an important role in motion deblurring.
% To deal with the multiple moving objects in the scenes, we propose to exploit both the motion segmentation cues and scene flow cues for stereo deblurring.

In our formulation, we use ResNet38~\cite{wu2016wider} to predict the semantic label map $\vM \in \mathbb{N}^{w \times h}$ as initialization for our ``generalized stereo deblur'' model. \rc{This approach ranks higher on Cityscapes~\cite{cordts2016cityscapes} where the image is captured on an urban street.} A $\vM $ determines the predicted semantic instance label for each pixel in each frame, which provides strong prior for boundary detection, motion estimation, and label classification for superpixels.

We first set roads, sky and trees are static background layer, and assume other things have a higher moving possibility to be the foreground layer. Here, a convincing background layer will provide the inline feature points on the background for ego-motion estimation. Then, we can estimate the disparity map and the 6-DOF camera motion using stereo matching and visual odometry with coarse background segmentation. 
We identify regions inconsistent with the estimated camera motion and estimate the motion at these regions separately. Each motion parameter $\vo$ is generated by moving clusters from sparse features points. In particular, the motion hypothesis is then generated using the 3-point RANSAC algorithm implemented in~\cite{geiger2011stereoscan}. These inconsistent regions can match with our prior $\vM$. This helps to maintain the boundary information for moving objects and avoid ringing artifacts (see Fig.~\ref{fig:3methodevolution} for details).

Each slanted plane in the image is labelled as moving or static according to the ego-motion estimation.
With the semantic segmentation masks, we can give each superpixel an additional label, foreground or background.
We then use the label map to initialize object label $k_i$ for each superpixel $i$. If most pixels' semantic label in $i^{th}$ superpixel are fore/background, the superpixel is more likely to belongs to the fore/background.
\rc{
% {\small
\begin{equation}
{k_i}(\vx) \in
\begin{cases}
\{1 \} &, \mathrm{if}\ \mathbf {M(x)} = \rm{Background} \\
\{2, \cdots, |\mathcal{O}| \} &, \mathrm{if}\ \mathbf {M(x)} = \rm{Foreground}.
\end{cases}
\end{equation}
% }
}
Although we provide over segmentation initially as shown in Fig.~\ref{fig:fig1}(a), our algorithm can precisely segment the moving objects after optimization (Fig.~\ref{fig:fig1}(b)) and provide more accurate motion boundaries information for optical flow estimation (Fig.~\ref{fig:fig1}(d)), and thereby facilitates stereo video deblurring (Fig.~\ref{fig:fig1}(h)).

With the semantic segmentation prior, we label each superpixel and objects more accurately, our approach obtains superior results in Moving object segmentation and scene flow estimation (see Fig.~\ref{fig:4comparefs} for details).

In the optimization part, instead of giving sample $k_i$ for every superpixel randomly, we use the semantic segmentation prior $\vM$ to give a more reliable sample for each superpixel (see Section~\ref{sec:sceneflow} for detail).

%===================================================================
%=================figure 2===========================================
\begin{figure*}
\begin{center}
\hspace{-0.3cm}
\includegraphics[width=0.99\textwidth]{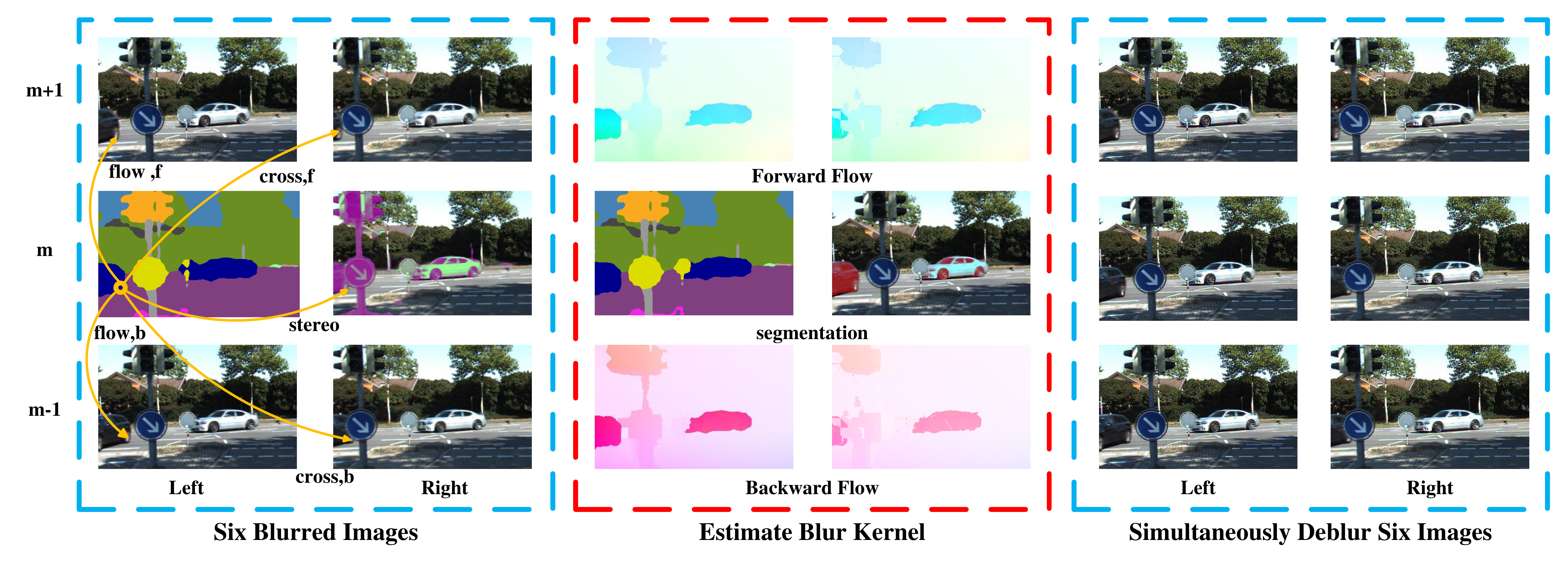}
\end{center}
\vspace{-2 mm}
\caption{ 
Illustration of our `generalized stereo deblurring' method. We simultaneously compute four scene flows (in two directions and in two views), Moving object segmentation and deblur six images. In case the input contains only two images, we use the reflection of the flow forward as the flow backward in the deblurring part.}
\label{fig:framework}
\end{figure*}
%%%%%% introduce the model %%%%%%%
\subsection{Energy Minimization}
We formulate the problem in a single framework as a discrete-continuous optimization problem to jointly estimate the scene flow, Moving object segmentation and deblur the stereo images. Specifically, our model is defined as

\begin{small}
\begin{equation}\label{eq:energy}
\begin{split}
\mathbf{E(n,o,L)}  =\underbrace{\sum_{i\in \calS} {\phi}_{i}(\vn_{i},\vo,\vL)}_{\text{data~term}} + \underbrace{\sum_{i,j \in \calS}{\phi}_{i,j }(\vn_{i},\vn_{j},\vo)}_{\substack{\text{scene flow}\\\text{smoothness term}}} + \underbrace{\sum_m \rc{\psi}_{m} (\vL)}_{\substack{\text{latent image}\\\text{regularisation}}},   
\end{split}
\end{equation}
\end{small}
where $i,\ j$ denotes the set of adjacent superpixels in $\calS$. The function consists of a data term, a smoothness term for scene flow, and a spatial regularization term for latent images.
%\textcolour{red}{Any term for moving object segmentation???}. 
Our model is initially defined on three consecutive pairs of stereo video sequences. It can also allow the input with two pairs of frames. Details are provided in Section~\ref{sec:experiments}. The energy terms are discussed in Section~\ref{sec:dataTerm}, Section~\ref{sec:smoothTerm}, and Section~\ref{sec:regularization}, respectively.

In Section~\ref{sec:optimization}, we perform the optimization in an alternative manner to handle mixed discrete and continuous variables, thus allowing us to jointly estimate scene flow, Moving object segmentation and deblur the images.

%%%%%%%%%%%%%%%%%%%% data term %%%%%%%%%%%%%%%%%%%%%
\subsection{Data Term}
\label{sec:dataTerm}
Our data term involves mixed discrete and continuous variables, and are of three different kinds. The first kind encodes the fact that the corresponding pixels across the six latent images should have a similar appearance, \ie, brightness constancy. This lets us write the term as

{\small
\begin{equation}
\phi_i^1(\vn_{i},\vo,\vL) = \theta_{1}\rc{\sum_{\vx \in \calS_i}} |\vL(\vx)-\vL^*(\mathbf{x^*})|_1,
\end{equation}
}
%\MM{$|\vM(\vx_i) = \vM(\vx^*_i)|$ should be $\delta$ function?}
where $\vL$ denotes the reference image, $\vL^*$ denotes the target image, the superscript $* \in \left\{ \mathbf{stereo}, \mathbf{flow}_{f,b}, \mathbf{cross}_{f,b} \right\}$ denote the warping direction to other images and $(\cdot)_{f,b}$ denotes the forward and backward direction, respectively (see Figure~\ref{fig:framework}). \rc{The terms is defined by summing the matching costs  of all pixels inside superpixel $i$.} We adopt the robust $\ell_1$ norm to enforce its robustness against noise and occlusions.

Our second potential, similar to one term used in~\cite{menze2015object}, is defined as

{\small
\usetagform{red}
\begin{equation}
\phi_i^2(\vn_{i},\vo)=
  \begin{cases}
  \theta_2 \rc{{\sum}_{\vx \in \calS_i}} \rho_{\alpha_1}(||\vx-\mathbf{x^*}||_2) & ,\mathrm{if} \  \vx\in \Pi_{\vx}, \\
  0 & ,\mathrm{otherwise}, \\
  \end{cases}
\end{equation}
}
where $\rho_{\alpha}(\cdot)=\min(|\cdot|,\alpha)$ denotes the truncated $l_1$ penalty function. More specifically, it encodes the information that the warping of feature points $x\in \Pi_{x}$ based on ${\bf H}^{*}$ should match its extracted correspondences $\vx^{*}$ in the target view. In particular, $\Pi_{\vx}$ is obtained in a similar manner as~\cite{menze2015object}.

The third data term, making use of the observed blurred images, is defined as
 
{\small
\begin{equation}
\mathbf{\phi}_i^3(\vn_{i},\vo,\vL)= \theta_{3}\sum_m\sum_{\partial}\left\|{\partial}\mathbf{A}_{m}(\vn_{i},\vo)\vL_m - {\partial}\mathbf{B}_m\right\|_2^2, 
\end{equation}}
where $\partial$ denotes the Toeplitz matrices corresponding to the horizontal and vertical derivative filters. This term encourages the intensity changes in the estimated blurred image to be close to that of the observed blurred image.
%{\colour{red} [It encourages the intensity changes in the estimated blur image to be close to that of the observed blur image. ?]}

%%%%%%%%%%%%%%%%%%%% smooth term %%%%%%%%%%%%%%%%%%%%%
\subsection{Smoothness Term for Scene Flow}
\label{sec:smoothTerm}
Our energy model exploits a smoothness potential that involves discrete and continuous variables. It is similar to the ones used in~\cite{menze2015object}. In particular, our smoothness term includes three different types. 

The first one is to encode the compatibility of two superpixels that share a common boundary by respecting the depth discontinuities. We define our potential function as
 
\usetagform{default} 
{\small
\begin{equation}
\phi_{i,j}^1(\vn_{i},\vn_{j})= \theta_{4}\sum_{\vx\in \mathcal{B}_{i,j}}\rho_{\alpha_2}(\omega_{i,j}(\vn_i,\vn_j,\vx)),
\end{equation}
}
where $d(\vn_i,\vx)$ is the disparity of pixel $\vx$ in superpixel $i$ in the reference disparity map, $\omega_{i,j}(\vn_i,\vn_j,\vx) = d(\vn_i,\vx)-d(\vn_j,\vx)$ are the dissimilarity value of disparity for pixel $\vx \in \mathcal{B}_{i,j}$ on the boundary.

The second potential is to encourage the neighbouring superpixels to orient in similar directions. It is expressed as
  
{\small
\begin{equation}
\phi_{i,j}^2(\vn_{i},\vn_{j}) = \theta_{5}\rho_{\alpha_3} \left(1-\frac{|\vn_{i}^T \vn_{j}|}{\left\| \vn_{i} \right\| \left\| \vn_{j} \right\|}\right).
\end{equation}
}

The shadows of moving objects have motion boundaries but no disparity discontinuities. However,  the motion boundaries are co-aligned with disparity discontinuities in general. Thus, we use the third and fourth potential encodes these discontinuities. This potential can be expressed as
 
\usetagform{red}
{\small
\begin{equation}
\begin{aligned}
& \phi_{i,j}^3(\vn_{i,k_i},\vn_{j,k_j})= \\
& \exp \left\{-\frac{\lambda}{|\mathcal{B}_{i,j}|}\sum\limits_{\vx \in \mathcal{B}_{i,j}} \omega_{i,j}(\vn_i,\vn_j,\vx)^2 \right\} \times \frac{|\vn_i^T \vn_j|}{\left\| \vn_i \right\| \left\| \vn_j \right\|} \times [k_i\neq k_j], \\
\end{aligned}
\end{equation}
}
where $|\mathcal{B}_{i,j}|$ denotes the number of pixels shared along boundary between superpixels $i$ and $j$. 
 
{\small
\begin{equation}
\begin{aligned}
&\phi_{i,j}^4(\vn_{i,k_i},\vn_{j,k_j},\vo_{k_i},\vo_{k_j}) =\\
&\exp \left\{-\frac{\lambda}{|\mathcal{B}_{i,j}|}\sum\limits_{\vx \in \mathcal{B}_{i,j}} G(\vo_{k_i},\vo_{k_j}) \right\} \times \frac{|\vn_{i,k_i}^T \vn_{j,k_j}|}{\left\| \vn_{i,k_i} \right\| \left\| \vn_{j,k_j} \right\|} \times [k_i\neq k_j], \\
%[\mathbf M(\vx_i)\neq \mathbf M(\vx_j)] \nonumber \\
\end{aligned}
\end{equation}
}
{\small
\begin{equation}\nonumber
    \begin{split}
        G(\vo_{k_i},\vo_{k_j}) = \theta_r(\mathrm{trace}(\mathbf{R}_{k_i}^{T}\mathbf{R}_{k_j})-1)/2%\\
        +\theta_t(\exp(-\left\|\mathbf{t}_{k_i}-\mathbf{t}_{k_j}\right\|)),\\
    \end{split}
\end{equation}}
where $\textbf{[} \mathbf{\cdot} \textbf{]}$ denotes the Iverson bracket. This encodes our belief that motion boundaries are more likely to occur at 3D folds or discontinuities than within smooth surfaces.

%%%%%%%%%%%%%%%%%%%% regularization term %%%%%%%%%%%%%%%%%%%%%
\subsection{Regularization Term for Latent Images}
\label{sec:regularization}
Several works~\cite{krishnan2009fast, krishnan2011blind} have studied the importance of spatial regularization in image deblurring. In our model, we use a total variation term to suppress the noise in the latent image while preserving edges, and penalize spatial fluctuations. Therefore, our potential takes the form
%=================function 11===========================================
\usetagform{default} 
\begin{equation} \label{Eregularization1}
\rc{\psi}_m = \sum_\vx|\nabla \vL_m|.
\end{equation}

Note that the total variation is applied to each colour channel separately.

%%%%%%%%%%%%%%%%%%%%%%%%%%% optimization %%%%%%%%%%%%%%%%%%%%%%%%%%%%
\section{Solution}\label{sec:optimization}
The optimization of our energy function defined in Eq.(\ref{eq:energy}), involving discrete and continuous variables, is very challenging to solve. Recall that our model involves two set of variables, namely scene flow variables and latent clean images. Fortunately, given one set of variables, we can solve the other efficiently. Therefore, we perform the optimization iteratively by the following steps,
\begin{itemize}
\item Fix latent clean image $\vL$, solve scene flow by optimizing Eq.(\ref{eq:sceneFlowEnergy}) (See Section~\ref{sec:sceneflow}).
\item Fix scene flow parameters, ${\bf n}$ and ${\bf o}$, solve latent clean images by optimizing Eq.(\ref{eq:latentImageEnergy}) (See Section~\ref{sec:deblurring}).
\end{itemize}

In the following sections, we describe the details for each optimization step.

%%%%%%%%%%%%%%%%%%%%%%%%%%% fix L optimize n o %%%%%%%%%%%%%%%%%%%%%%%%%%%%
\vspace{-2mm}
\subsection{Scene flow estimation}\label{sec:sceneflow}
We fix latent images, namely $\vL = \tilde{\vL}$. Eq.(\ref{eq:energy}) reduces to

%=================function 10===========================================
\rc{
{\small
\begin{equation}\label{eq:sceneFlowEnergy}
\min_{{\bf n},{\bf o}}\sum_{i\in \calS}\sum_{m=1}^{3} \mathbf{\phi}_{i}^{m}(\vn_{i},\vo,\tilde{\vL}) + \sum_{i, j \in \calS}\sum_{m=1}^{4}\mathbf{\phi}_{i,j}^{m}(\vn_{i},\vn_{j},\vo),
\end{equation}}}
which becomes a discrete-continuous CRF optimization problem.

We use the sequential tree-reweighted message passing (TRW-S) method in~\cite{menze2015object} to find an approximate solution.
Since the label $k$ of $\vn_{i}$ of each superpixel is drawing randomly, we use the semantic segmentation prior $\vM$ to give a more reliable sample of each superpixel. We modify their sampling strategy as shown in Algorithm 1.

%%%%%%%%%%%%%%%%%%%%%%%%% algorithm %%%%%%%%%%%%%%%%%%%%%%%%%%%
\begin{algorithm}
    \SetKwInOut{Objective}{Input}
    \SetKwInOut{Algorithm}{Output}
\caption{TRW-S Optimization}
\Objective{$\tilde{\vL}$, $\vM$, $\vB$.}

Initialize $\vn$ and $\vo$ as described in `Initialization'.

Iteration times = 3

\hspace{2mm} For all $i \in S$

\hspace{5mm} Draw sample for $\vn_{i}$ (Gaussian)

\hspace{5mm} Draw sample for $\mathbf{k}_{i}$(M)

\hspace{2mm} For all $k \in \mathcal{O}$

\hspace{5mm} Draw sample for $\vo_k$ (MCMC)

\hspace{2mm} Run TRW-S~\cite{kolmogorov2006convergent} on discretized problem

\Algorithm{$\vn_{i,k_i}, \vo_{k_i}$}
\label{Algorithm:TRW}
\end{algorithm}
\vspace{-2 mm}
%%%%%%%%%%%%%%%%%%%%%%%%%%% fix n o optimize L %%%%%%%%%%%%%%%%%%%%%%%%%%%%
\vspace{-2mm}
\subsection{Debblurring}\label{sec:deblurring}
Given the scene flow parameters, namely \rc{$\vn=\tilde{\vn}$, and $\vo=\tilde{\vo}$}, the blur kernel matrix, $\vA_m$ is derived based on Eq.(\ref{eq:kimblurKernel}), and Eq.(\ref{eq:structuredflow}). The objective function in Eq. (\ref{eq:energy}) becomes convex with respect to $\vL$ and is expressed as

%=================function 11===========================================
{\small
\begin{equation}\label{eq:latentImageEnergy}
\min_{{\bf L}}\sum_{\calS_i\in \calS} \mathbf{\phi}_{i}^1(\tilde{\vn}_{i},\tilde{\vo},\vL) +\mathbf{\phi}_{i}^3(\tilde{\vn}_{i},\tilde{\vo},\vL)+\rc{\psi}_{m}(\vL).
\end{equation}}

In order to obtain sharp image $\vL$, we adopt the conventional convex optimization method~\cite{chambolle2011first} and derive the primal-dual updating scheme as follows

%=================function 12===========================================
{\small
\begin{equation}
\left\{
\begin{gathered}
\begin{aligned}
& \vp^{r+1}=\frac{\vp^{r}+\gamma \nabla \vL_m^r}{\max(1,\text{abs}(\vp^{r}+\gamma\nabla \vL_m^r))}\\
& \vq^{r+1}=\frac{\vq^{r}+\gamma\theta_1 (\vL_m^r-\vL_{*}^{r})}{\max(1,\text{abs}(\vq^{r}+\gamma\theta_1 (\vL_m^r-\vL_{*}^{r}))}\\
& \vL_{m}^{r+1} =\arg \min_{\vL_m} \sum_i \theta_{3}\sum_{\partial}\left\|{\partial}\mathbf{A}_{m}\vL_m - {\partial}\mathbf{B}_m\right\|_2^2 +\\
& \frac{\left\|[\vL_m-\eta ((\nabla {\bf p}_{m}^{r+1})^{T}+\theta_1({\bf q}^{r+1}-{\bf q}_{*}^{r+1})^{T})]-\vL_m^r\right\|^2}{2\eta},
\end{aligned}
\end{gathered}\right.
\end{equation}
}
where $\vp_m$, $\vq_{m,*}$ are the dual variables, $\gamma$ and $\eta$ are the step variants which can be modified at each iteration, and $r$ is the iteration number.

%%%%%%%%%%%%%%%%%%%%%%%%% algorithm %%%%%%%%%%%%%%%%%%%%%%%%%%%
\begin{algorithm}[h]
    \SetKwInOut{Objective}{Input}
    \SetKwInOut{Algorithm}{Output}

\Objective{Stereo Blurred Image Sequences $\vB$, Semantic Segmentation of Reference Image Pair.}

Initialize $\vn$ and $\vo$ as described in `Initialization'.

\hspace{2mm} Run Algorithm 1 minimize Eq.~(\ref{eq:sceneFlowEnergy}).
Estimate scene flow and moving object segmentation map.

\hspace{2mm} Run Primal-Dual~\cite{chambolle2011first} minimize Eq.~(\ref{eq:latentImageEnergy}).
Restoration clean image.

Repeat steps 2,3 \rc{until reaches a preset iteration number (3 in our experiment).}

\Algorithm{Latent Images $\vL$, Moving object Segmentation Map, Scene Flow}
\caption{Proposed deblurring system}
\end{algorithm}
%%%%%%%%%%%%%%%%%
%%%%%%%%%%%% Experimental %%%%%%%%%%%%%%
\input{experiments}

%%%%%%%%%%%%%%%%%%%Conclusion%%%%%%%%%%%%%%%%%%%%%
\section{Conclusion}
In this paper, we present a joint optimization framework to tackle the challenging task of stereo video deblurring where scene flow estimation, Moving object segmentation and video deblurring are solved in a coupled manner. Under our formulation, the motion cues from scene flow estimation and blur information could reinforce each other, and produce superior results than conventional scene flow estimation or stereo deblurring methods. We have demonstrated the benefits of our framework on extensive synthetic and real stereo sequences. In future, we plan to extend our method to deal with multiple frames to achieve better stereo deblurring.

%==================================================================
\section*{Acknowledgements}
This research was supported in part by Australia Centre for Robotic Vision, the Natural Science Foundation of China grants (61871325, 61420106007, 61671387, 61603303) and the Australian Research Council (ARC) grants (DE140100180, DE180100628, DP150104645).

\flushbottom

{\small
\bibliographystyle{IEEEtran}
%\bibliography{IEEEabrv,Flow_Deblur_Reference.bib}
\bibliography{Flow_Deblur_Reference}
}%IEEEbiography

\begin{IEEEbiographynophoto}{Liyuan Pan}
 is currently pursuing the Ph.D. degree in the College of Engineering and Computer Science, Australian National  University, Canberra, Australia. She received her B.E degree from Northwestern Polytechnical University, Xian, China in 2014. Her interests include deblurring, flow estimation, depth completion, and event camera. 
\end{IEEEbiographynophoto}
\vspace{-3mm}
\begin{IEEEbiographynophoto}{Yuchao Dai} 
is currently a Professor with School of Electronics and Information at the Northwestern Polytechnical University (NPU). He received the B.E. degree, M.E degree and Ph.D. degree all in signal and information processing from NPU, Xian, China, in 2005, 2008 and 2012, respectively. 
He was an ARC DECRA Fellow with the Research School of Engineering at the Australian National University, Canberra, Australia from 2014 to 2017and a Research Fellow with the Research School of Computer Science at the Australian National University, Canberra, Australia from 2012 to 2014. 
His research interests include structure from motion, multi-view geometry, low-level computer vision, deep learning, compressive sensing and optimization. He won the Best Paper Award in IEEE CVPR 2012, the DSTO Best Fundamental Contribution to Image Processing Paper Prize at DICTA 2014, the Best Algorithm Prize in NRSFM Challenge at CVPR 2017, the Best Student Paper Prize at DICTA 2017 and the Best Deep/Machine Learning Paper Prize at APSIPA ASC 2017. He served as Area Chair for WACV 2019/2020 and ACM MM 2019.
\end{IEEEbiographynophoto}
\vspace{-3mm}
\begin{IEEEbiographynophoto}{Miaomiao Liu} is a Lecturer and an ARC DECRA Fellow in the Research School of Engineering, the Australian National University. She was a Research Scientist at Data61/CSIRO from 2016-2018. Prior to that she was a researcher in NICTA. She received the BEng, MEng, and PhD degrees from Yantai Normal University, Yantai, China, Nanjing University of Aeronautics and Astronautics, Nanjing, China, and the University of Hong Kong, Hong Kong SAR, China, in 2004, 2007, and 2012, respectively. Her research interests include 3D vision, 3D reconstruction and 3D scene modeling and Understanding. She is a member of the IEEE.
\end{IEEEbiographynophoto}
\vspace{-3mm}
\begin{IEEEbiographynophoto}{Fatih Porikli}
is an IEEE Fellow and a Professor in the Research School of Engineering, Australian National University. He is acting as the Chief Scientist at Huawei, Santa Clara. He received his Ph.D. from New York University in 2002. 
%Previously he served Distinguished Research Scientist at Mitsubishi Electric Research Laboratories. 
His research interests include computer vision, pattern recognition, manifold learning, image enhancement, robust and sparse optimization and online learning with commercial applications in video surveillance, car navigation, intelligent transportation, satellite, and medical systems.
\end{IEEEbiographynophoto}
\vspace{-3mm}
\begin{IEEEbiographynophoto}{Quan Pan} is the Dean of the Automation School of Northwestern Polytechnical University (NPU). He received the Ph.D. degree from NPU, Xi’an, China, in 1997. %Since 1998, he has been a Professor with the Automation School, and the Director of the Research Institute of Control and Information, NPU,1998. From 1996 to 2002, he was the Duty Dean with the Graduate School, NPU. From 2002 to 2004, he was the Duty Dean with the Management School, NPU. From 2004 to 2009, he was the Director of the Office of Development and Planning, NPU. Since 2009, he has been the Dean with the Automation School, NPU. His research interests include information fusion, hybrid system estimation theory, multiscale estimation theory, target tracking, and image processing. 
He is a Member of IEEE, a Member of the International Society of Information Fusion, a Board Member of the Chinese Association of Automation, and a Member of Chinese Association of Aeronautics and Astronautics. 
He obtained the 6th Chinese National Youth Award for Outstanding Contribution to Science and Technology in 1998 and the Chinese National New Century Excellent Professional Talent in 2000.
\end{IEEEbiographynophoto}

\end{document}

%% file: experiments.tex
\section{Experiments}\label{sec:experiments}

To demonstrate the effectiveness of our method, we evaluate it based on two datasets: the synthetic chair sequence~\cite{sellent2016stereo} and the KITTI dataset~\cite{geiger2013vision}. We report our results on both datasets in the following sections.

%=================Table our blur model compare with others===============
\begin{table}[ht]\footnotesize
\centering
\vspace{-2 mm}
\caption{Quantitative comparisons on disparity, optical flow and deblurring results on the KITTI dataset (BlurData-1).%\MM{Liyuan: CHECK WHETHER THE SEQUENCE IS BLURDATA-1??? TOGETHER WITH THE DESCRIPTION IN THE TEXT.}. %(the best results are in bold)
}
\label{all_all}
\begin{tabular}{c|c|c|c|c|c|c|c}
\hline
\multicolumn{2}{c|}{\multirow{2}{*}{KITTI Dataset}} & \multicolumn{2}{c|}{Disparity} & \multicolumn{2}{c|}{Flow} & \multicolumn{2}{c}{PSNR} \\ \cline{3-8}
\multicolumn{2}{c|}{}                                      & m              & m+1           & Left        & Right       & Left        & Right    \\ \hline
\multicolumn{2}{c|}{\rc{Vogel} \etal ~\cite{vogel20153d}}        & 8.20           & 8.50          & 13.62       & 14.59       & /           & /        \\ \hline
\multicolumn{2}{c|}{Kim and Lee ~\cite{hyun2015generalized}}& /              & /             & 38.89       & 39.45       & 28.25       & 29.00    \\ \hline
\multicolumn{2}{c|}{Sellent \etal ~\cite{sellent2016stereo}}& 8.20           & 8.50          & 13.62       & 14.59       & 27.75       & 28.52    \\ \hline
\multicolumn{2}{c|}{\rc{Kupyn \etal ~\cite{Kupyn_2018_CVPR}}}& /           & /          & /       & /       & \rc{28.34}       & \rc{28.73}    \\ \hline
\multicolumn{2}{c|}{\rc{Tao \etal ~\cite{Tao_2018_CVPR}}}& /           & /          & /       & /       & \rc{29.55}       & \rc{29.95}    \\ \hline
\multicolumn{2}{c|}{Pan \etal~\cite{Pan_2017_CVPR}}         & 6.82           &  8.36         & 10.01        & 11.45       & 29.80       & 30.30    \\ \hline
\multicolumn{2}{c|}{Ours}                                  & \bf 6.18       &  \bf 7.49     & \bf 9.83   & \bf 11.14    & \bf 29.85       & \bf 30.50    \\ \hline
\multicolumn{8}{c}{Baseline}                \\ \hline
\multicolumn{2}{c|}{\cite{vogel20153d} and~\cite{hyun2015generalized} }     & /   & /   & 22.42  & /           & 28.11       & /        \\ \hline
%\multicolumn{2}{c|}{Ours without iteration }               & /              & /             & /           & /           & 29.57       & /        \\ \hline
\end{tabular}
\end{table}
\vspace{-2mm}
\subsection{Experimental Setup}
\noindent{\bf Initialization.} Our model in Section~\ref{sec:model} is formulated on three consecutive stereo pairs. In particular, we treat the middle frame in the left view as the reference frame. We adopt StereoSLIC~\cite{yamaguchi2013robust} to generate superpixels. Given the stereo images, we apply the approach in~\cite{geiger2011stereoscan} to obtain sparse feature correspondences. The traditional SGM~\cite{hirschmuller2008stereo} method is applied to obtain the disparity map.  We further leverage the semantic segmentation results to provide priors for motion segmentation. \rc{In particular, we applied the pre-trained model from the high-accuracy method~\cite{wu2016wider} on our blurred image.} Based on the obtained semantics, we generate a binary map ${\bf M}$ which indicates the foreground as 1 and background as 0 by grouping the estimated semantics (see Section~\ref{sec:3.2} for details.)
%The segmentation mask is obtained by
%We use approach in~\cite{wu2016wider} to get semantic segmentation masks.
The motion hypotheses are first generated using RANSAC algorithm implemented in~\cite{geiger2011stereoscan}.
% and then refined based on Eq.~(\ref{eq: refinedK}).
%In order to obtain the model parameters $\{\theta\}_{1,2,4,5,6}$ and $\{\alpha\}$, we perform block-coordinate-descent on a subset of 30 randomly selected training images, where $\theta_3 = \gamma = 250$.
Regarding the model parameters, we perform grid search on 30 reserved images. In our experiments, we fix the model parameters as \rc{$\theta_1 = 0.7$, $\theta_2 = 5.5$,  $\theta_3 =0.7$, $\gamma = 250$, $\theta_4 = 0.37$, $\theta_5 = 17$, $\lambda=0.13$, $\alpha_{1}=3.39$, $\alpha_{2}=2.5$, $\alpha_{3}=0.25$, $\theta_{r}=0.05$, $\theta_{t}= 0.1$.}

\vspace{1mm}
\noindent{\bf Evaluation metrics.} Since our method estimates the scene flow, segments moving objects and deblurs images, we thus evaluate multiple tasks separately. As for the scene flow estimation results, we evaluate both the optical flow and disparity map by the same error metric, which is by counting the number of pixels having errors more than $3$ pixels and $5\%$ of its ground-truth. We adopt the PSNR to evaluate the deblurred image sequences for left and right view separately.~We report precision (P), recall (R) and F-measure (F) for our motion segmentation results.~Those metrics are defined as:
% {\small
\begin{equation}
\begin{aligned}
P = \frac{t_p}{t_p+f_p},\;\;R = \frac{t_p}{t_p+f_n},\;\;F = \frac{2R*P}{R+P},
\end{aligned}
\end{equation}
%}
where the true positive $t_p$ represents the number of pixels that have been correctly detected as moving objects; false positive $f_p$ are defined as pixels that have been mis-detected as moving pixels; false negative $f_n$ are denoted as moving pixels that have not been detected correctly.
Thus, for each sequence, we report disparity errors for three stereo image pairs, flow errors in forward and backward directions, and PSNR values for six images, and precision, recall and F-measure for the Moving object segmentation results.

\begin{figure}[!t]\begin{center}
\includegraphics[width=0.225\textwidth]{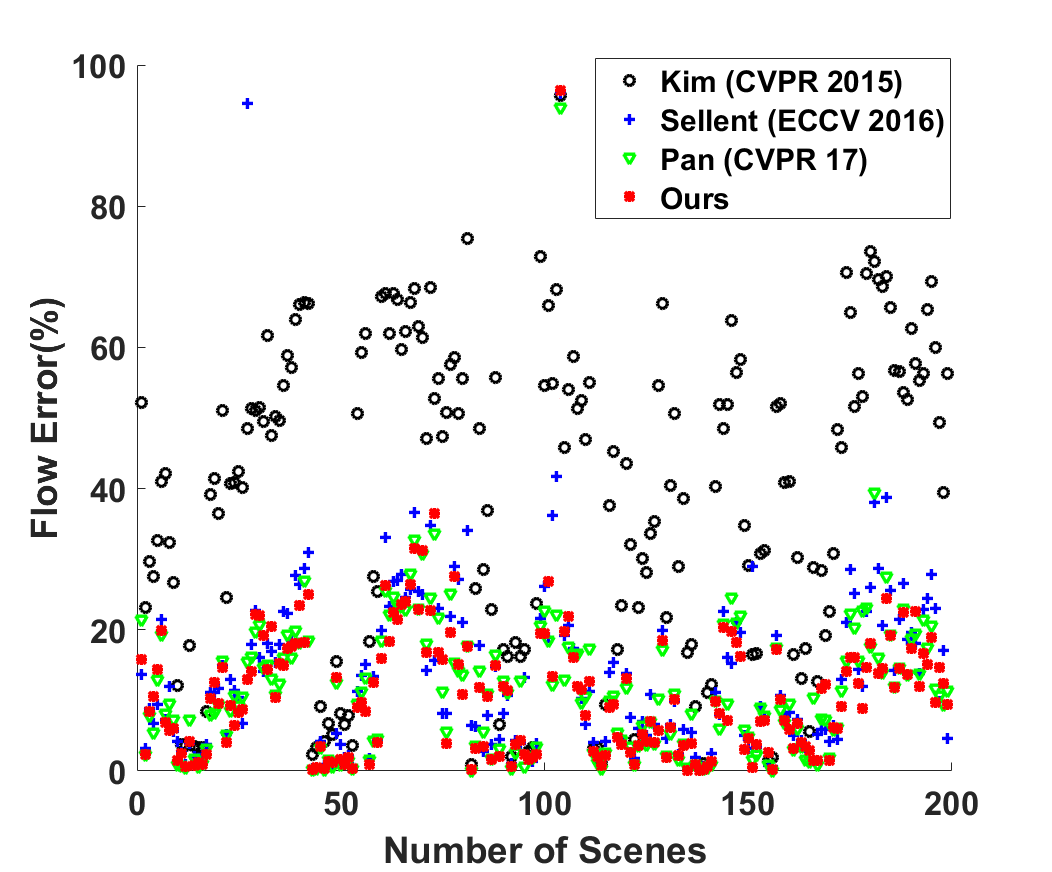}
\includegraphics[width=0.225\textwidth]{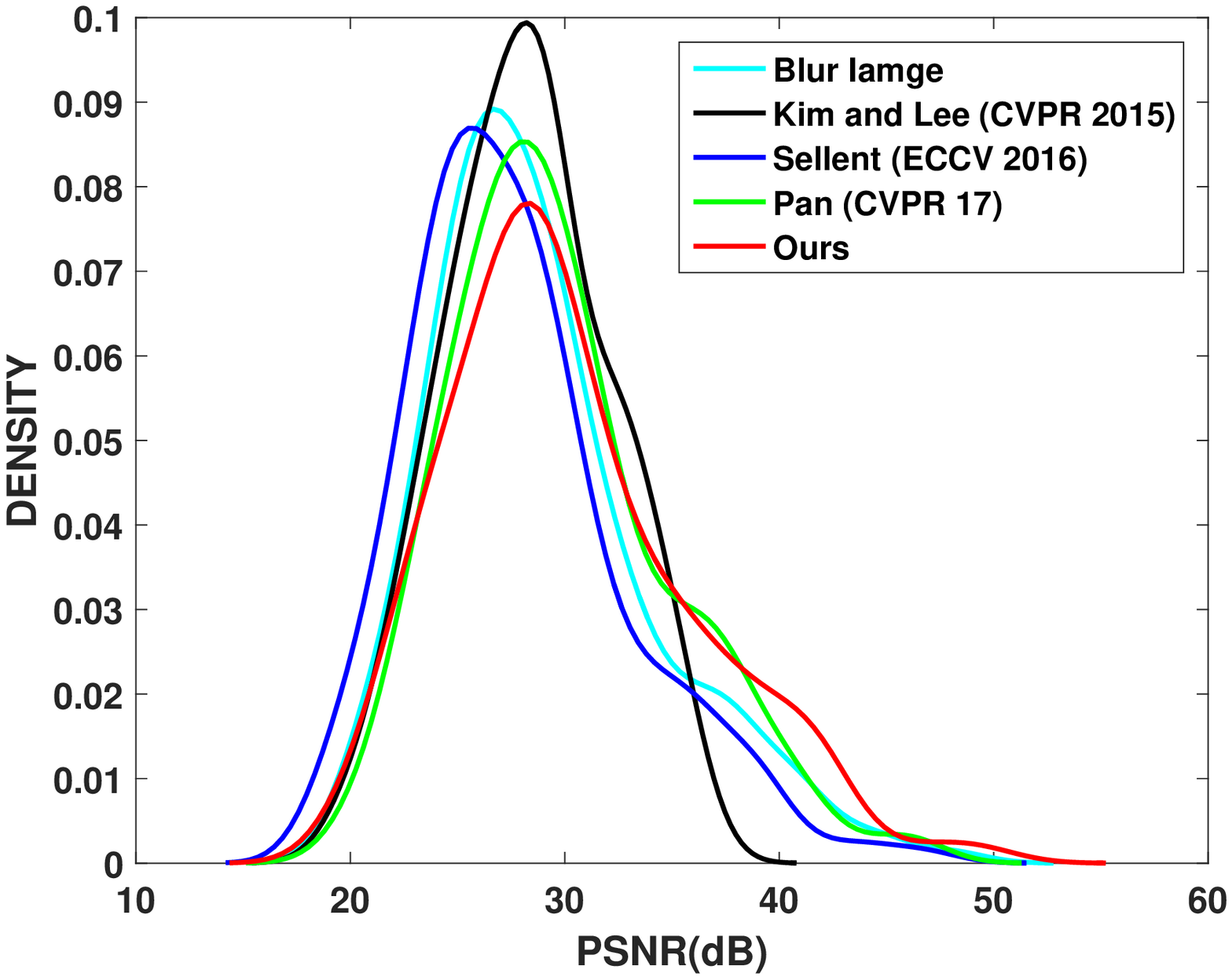}
\end{center}
\vspace{-2.5 mm}
\caption{Left: The flow estimation errors for 199 scenes in the KITTI dataset. Our method clearly outperforms the monocular and stereo video deblurring methods. Right: The distribution of the PSNR scores for 199 scenes in the KITTI dataset(BlurData-1). The probability distribution function for each PSNR was estimated using kernel density estimation with a normal kernel function. The heavy tail of our method means larger PSNR can be achieved using our method.}
\label{fig:err_compare}\label{fig:psnr_compare}
\end{figure}

\vspace{1mm}
\noindent\textbf{Baselines.} We first compare our scene flow results with piece-wise rigid scene flow method (PRSF)~\cite{vogel20153d}, whose performance ranks as one of the top 3 approaches on KITTI optical flow benchmark~\cite{geiger2013vision}. \rc{We then compare our results with the state-of-the-art stereo deblurring approach~\cite{sellent2016stereo}, monocular deblurring approach~\cite{kim2014segmentation} and deep-learning-based deblurring approaches~\cite{Tao_2018_CVPR,Kupyn_2018_CVPR}. }We compare our moving object segmentation results with the state-of-the-art approach using sharp stereo video sequences~\cite{zhou2017moving}. In addition, we further choose NLC~\cite{faktor2014video} and FST~\cite{papazoglou2013fast} as baselines since they are more robust to occlusions, motion blur and illumination changes according to the comprehensive evaluations in~\cite{Perazzi2016}. We make the quantitative comparison of our model w/o explicitly imposing semantics priors for our flow and deblurring results in Fig~\ref{fig:err_compare}. \rc{In addition, we compare with our previous method (Pan \etal CVPR 17) that has no semantics priors. The comparison clearly shows that the performance is improved significantly with the introduction of semantics as priors.} %In Fig.~\ref{fig:adddeblurresult}(d-e) and (j-k) qualitatively show that imposing the semantics prior can lead to better flow estimation (see the first row for the flow estimation).

\vspace{1mm}
\noindent\textbf{Runtime:}
In all experiments, we simultaneously compute two directions, namely forward and backward, scene flows, restore six blurred images and segment all moving objects. Our MATLAB implementation with C++ wrappers requires a total runtime of 35 minutes for processing one scene (6 images, 3 iterations) on a single i7 core running at 3.6 GHz.

%======================================================
%=================figure compare with others (psnr and flow err)============
% \begin{figure}[!t]\begin{center}
% \includegraphics[width=0.35\textwidth]{./a_figure/i_err_compare.eps}
% \end{center}
% \vspace{-2 mm}
% \caption{The flow estimation errors for 199 scenes in the KITTI dataset. Our method clearly outperforms the monocular and stereo video deblurring methods.}
% \label{fig:err_compare}
% \end{figure}

% \begin{figure}
% \begin{center}
% \includegraphics[width=0.35\textwidth]{./a_figure/psnr_d.eps}
% \end{center}
% \vspace{-2 mm}
% \caption{The distribution of the PSNR scores for 199 scenes in the KITTI dataset(BlurData-1). The probability distribution function for each PSNR was estimated using kernel density estimation with a normal kernel function. The heavy tail of our method means larger PSNR can be achieved using our method.}
% \label{fig:psnr_compare}
% \end{figure}

%=================figure iter===========================================
\begin{figure}[!t]
\begin{center}
\includegraphics[width=0.225\textwidth]{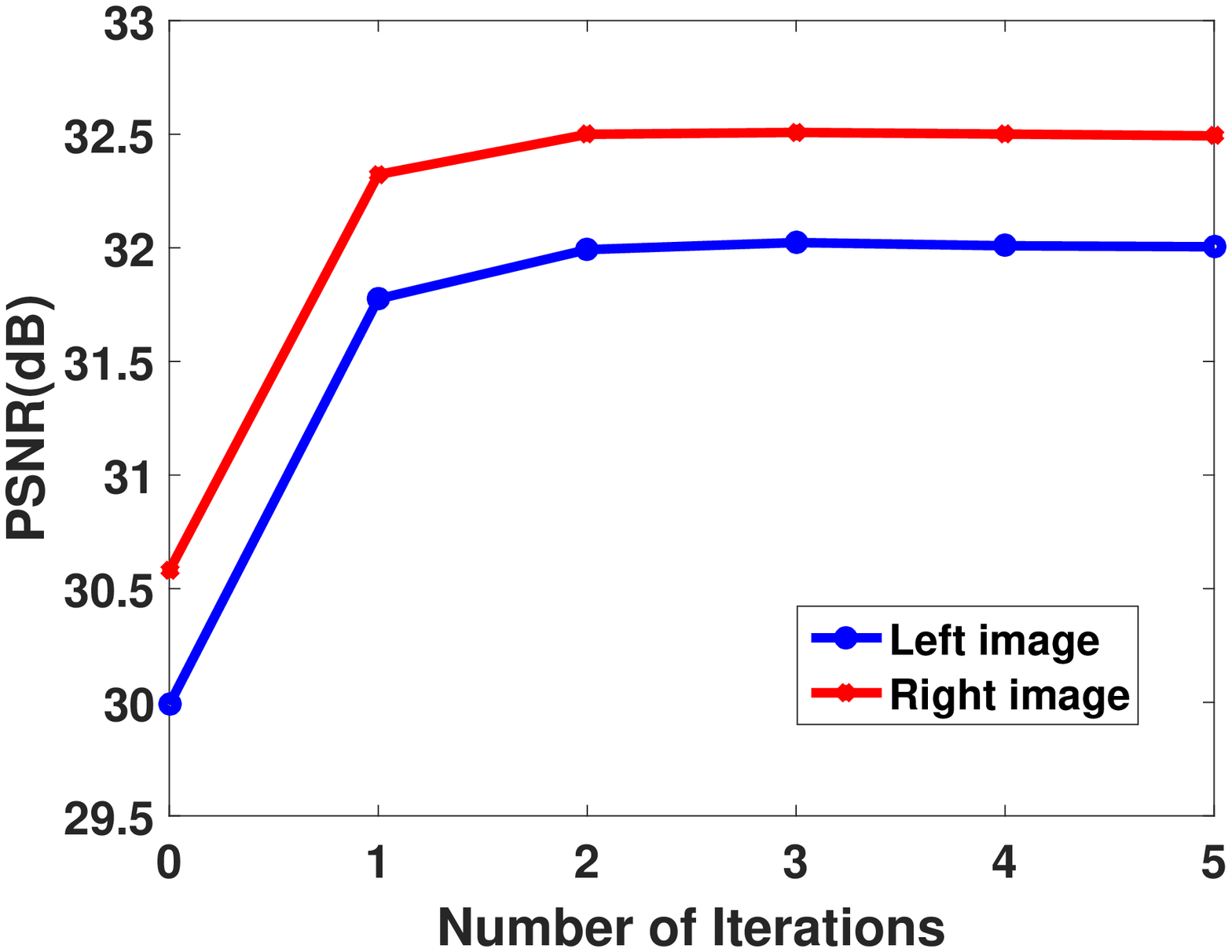}
\includegraphics[width=0.222\textwidth]{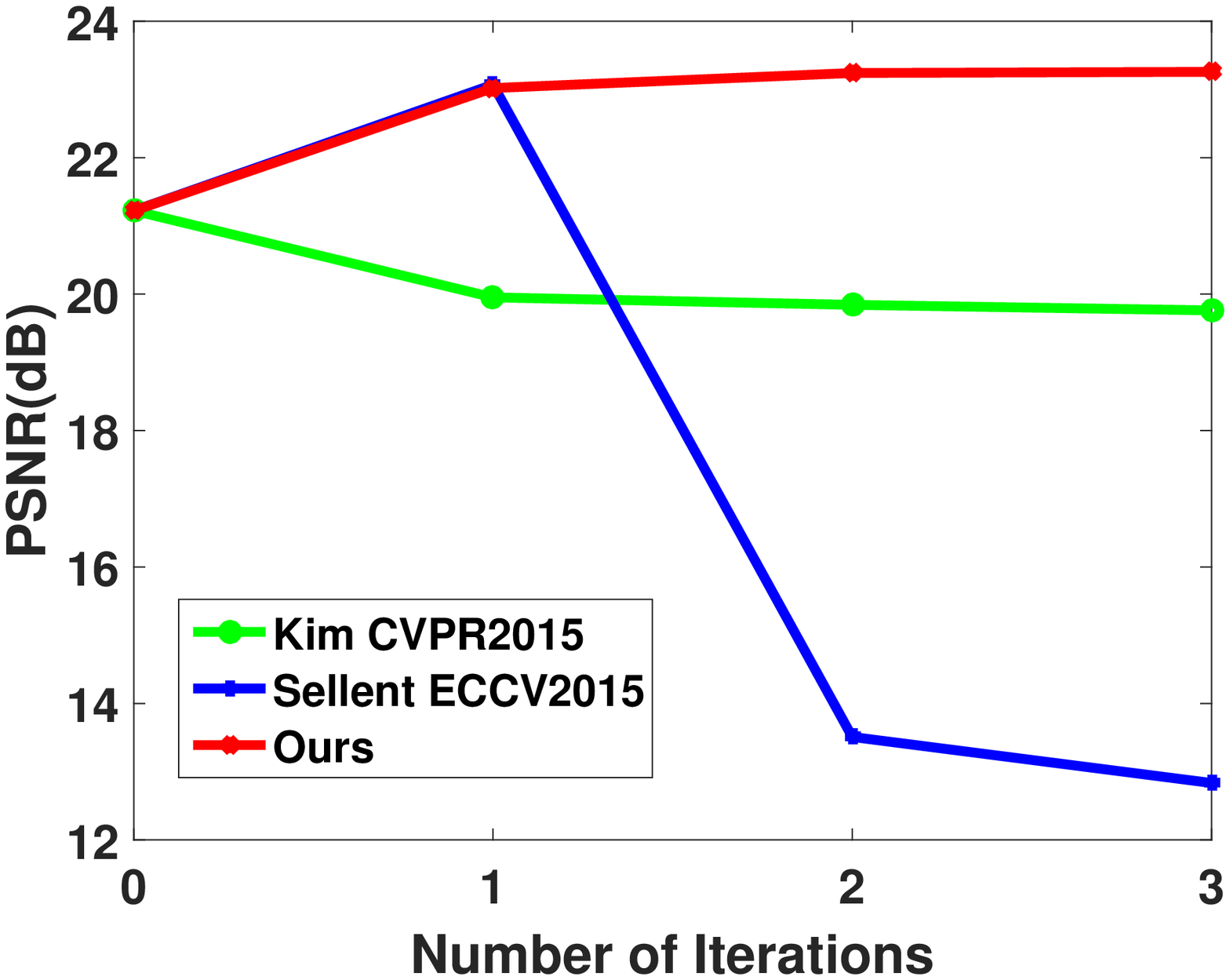}
\end{center}
\vspace{-2.5 mm}
\caption{The deblurring performance of our approach with~respect to the number of iterations. (left) Our method on our dataset with the gap of 0.3 dB between the first and the last iteration. (right) Several baselines on 'Chair'.}
\label{fig:iter_psnr}
\end{figure}

\begin{figure}[!t]
\begin{center}
\begin{tabular}{ccc}
\hspace{-4 mm}
\includegraphics[width=0.155\textwidth]{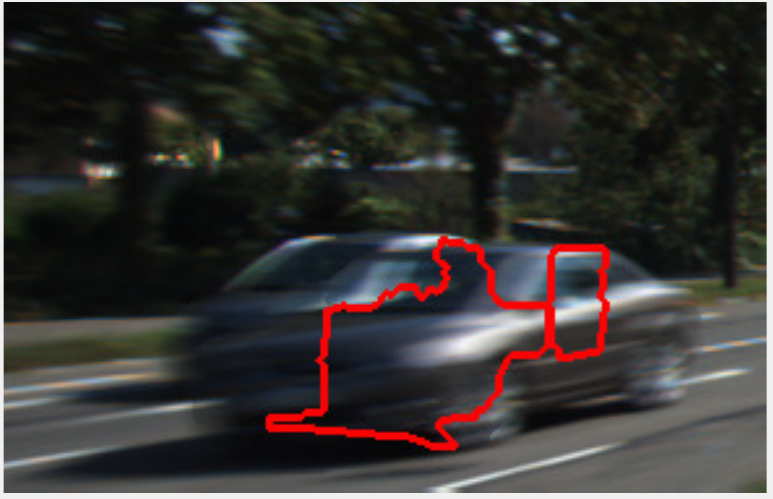}
\hspace{-4 mm}
&\includegraphics[width=0.155\textwidth]{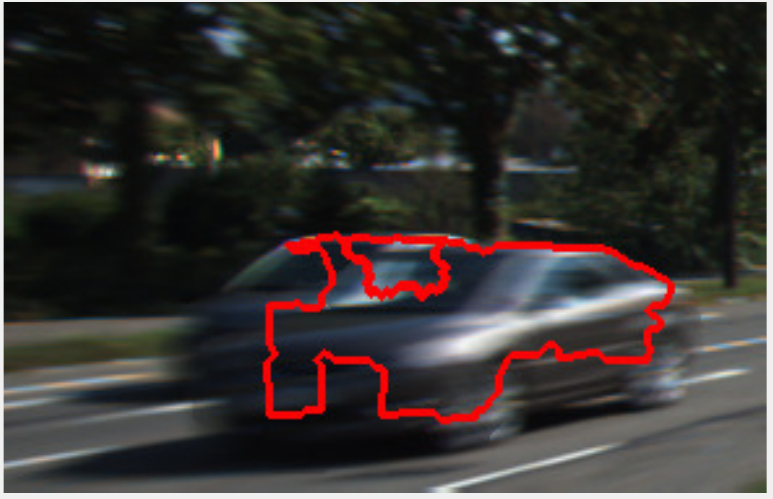}
\hspace{-4 mm}
&\includegraphics[width=0.155\textwidth]{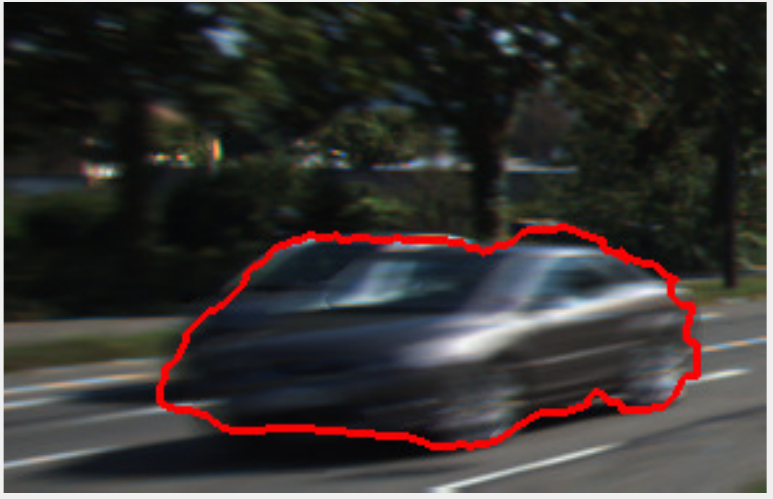}\\
\hspace{-4 mm}
\rcs{iterations = 1}
\hspace{-4 mm}
&\rcs{iterations = 2}
\hspace{-4 mm}
&\rcs{iterations = 3}
\end{tabular}
\end{center}
\vspace{-2.5 mm}
\caption{The moving object segmentation result with respect to the number of iterations}
\label{fig:iter_seg}
\end{figure}
%===============================================

\begin{table}\footnotesize
\centering
\caption{Moving object segmentation evaluation on the KITTI dataset BlurData-1.}
\label{tab:movingComparison}
\begin{tabular}{l|c|c|c}
\hline
Methods                                    & Recall(R)     & \rc{Precision (P)}     & F-measure (F)     \\ \hline
Menze \etal~\cite{menze2015object}        & 0.7995        & 0.5841          & 0.6045            \\ \hline
Zhou \etal~\cite{zhou2017moving}          & 0.7641        & 0.6959          & 0.7284            \\ \hline
Papazoglou \etal~\cite{papazoglou2013fast} & 0.5945        & 0.3199          & 0.2938            \\ \hline
Faktor \etal~\cite{faktor2014video}        & 0.4761        & 0.3148          &  0.3339           \\ \hline
\rc{Baseline}         & \rc{0.7633}        & \rc{0.6113}          & \rc{0.6789}           \\ \hline
Our                                        & \bf 0.8520    & \bf 0.7281      & \bf 0.7426        \\ \hline
\end{tabular}
\end{table}
%================additional result==============
\begin{figure*}
\begin{center}
%\begin{small}
% \rule{\linewidth}{1cm}
% \resizebox{\textwidth}{!}{
\begin{tabular}{cccccc}
\hspace{-0.4cm}
\includegraphics[width=0.164\textwidth]{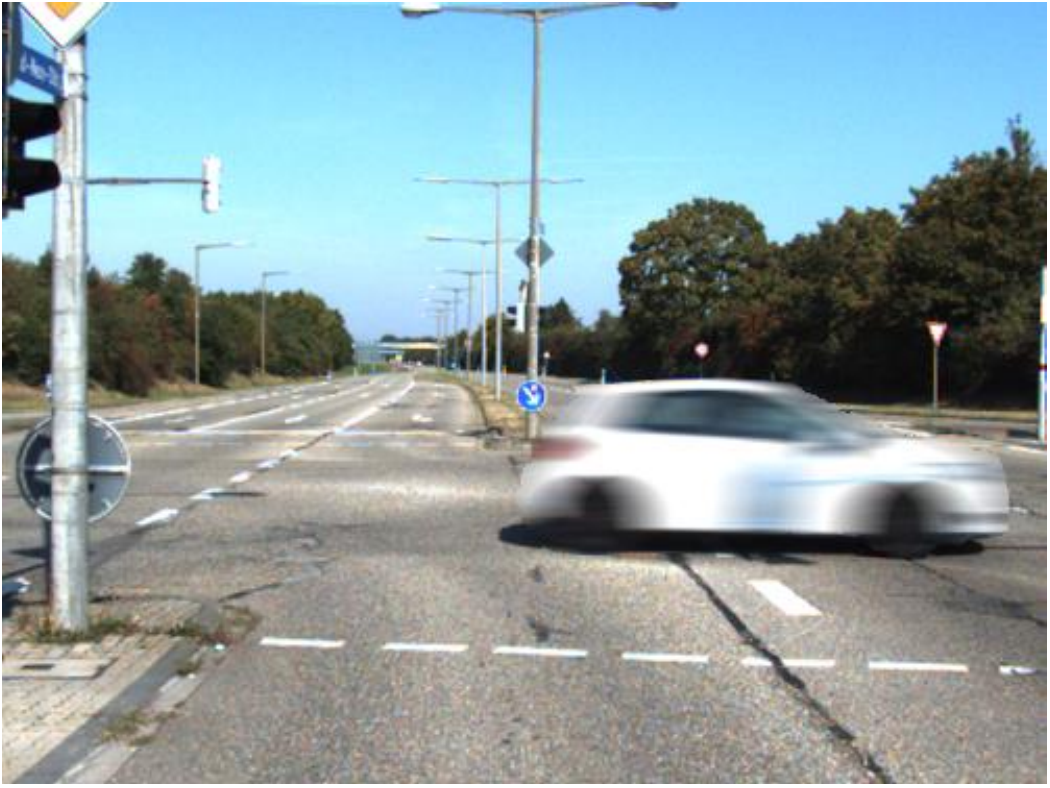}
&\hspace{-0.45cm}
\includegraphics[width=0.164\textwidth]{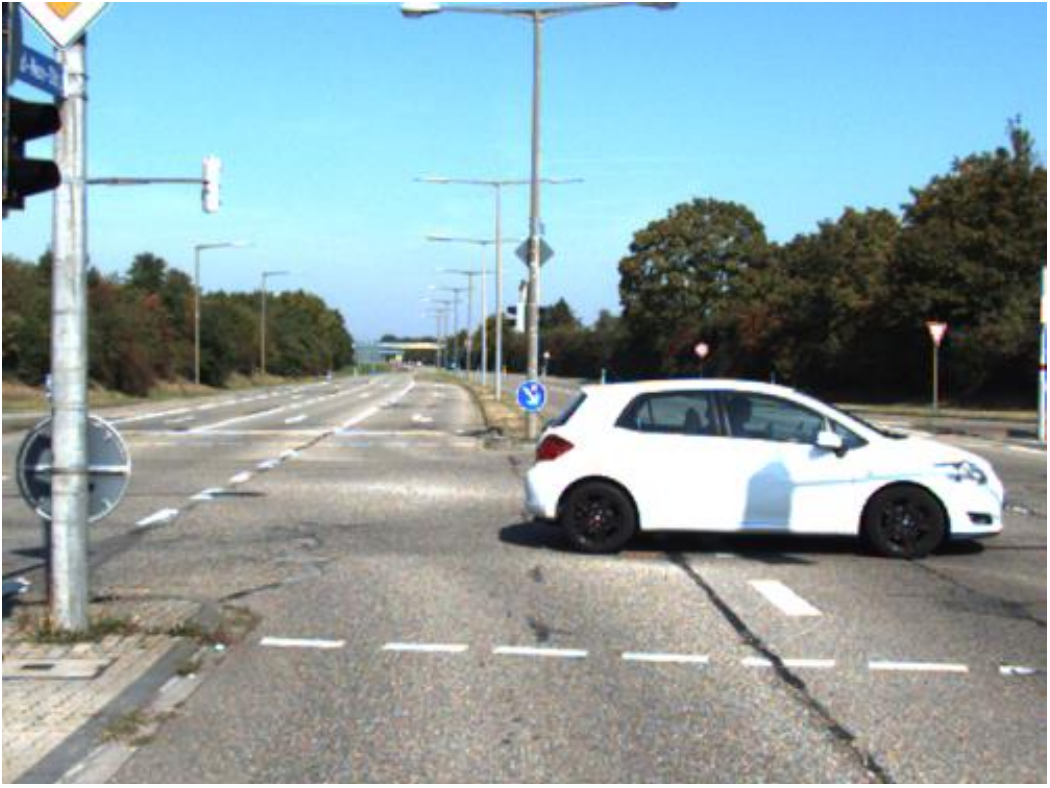}
&\hspace{-0.4cm} \includegraphics[width=0.164\textwidth]{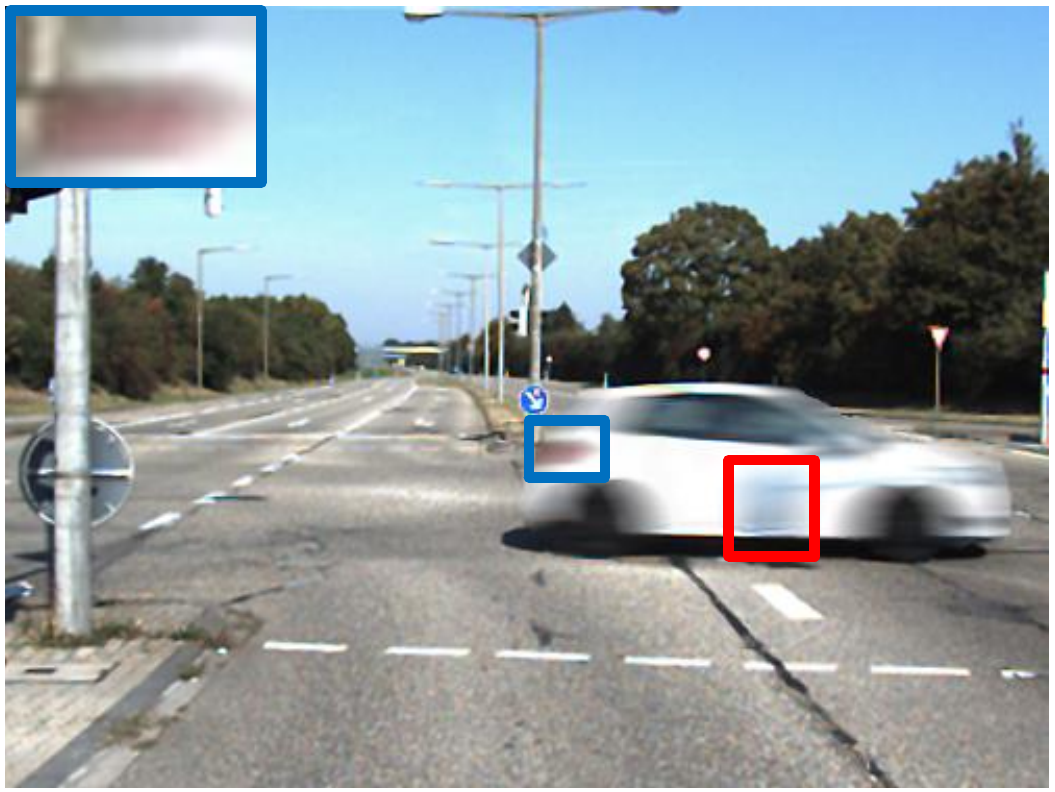}
&\hspace{-0.4cm}\includegraphics[width=0.164\textwidth]{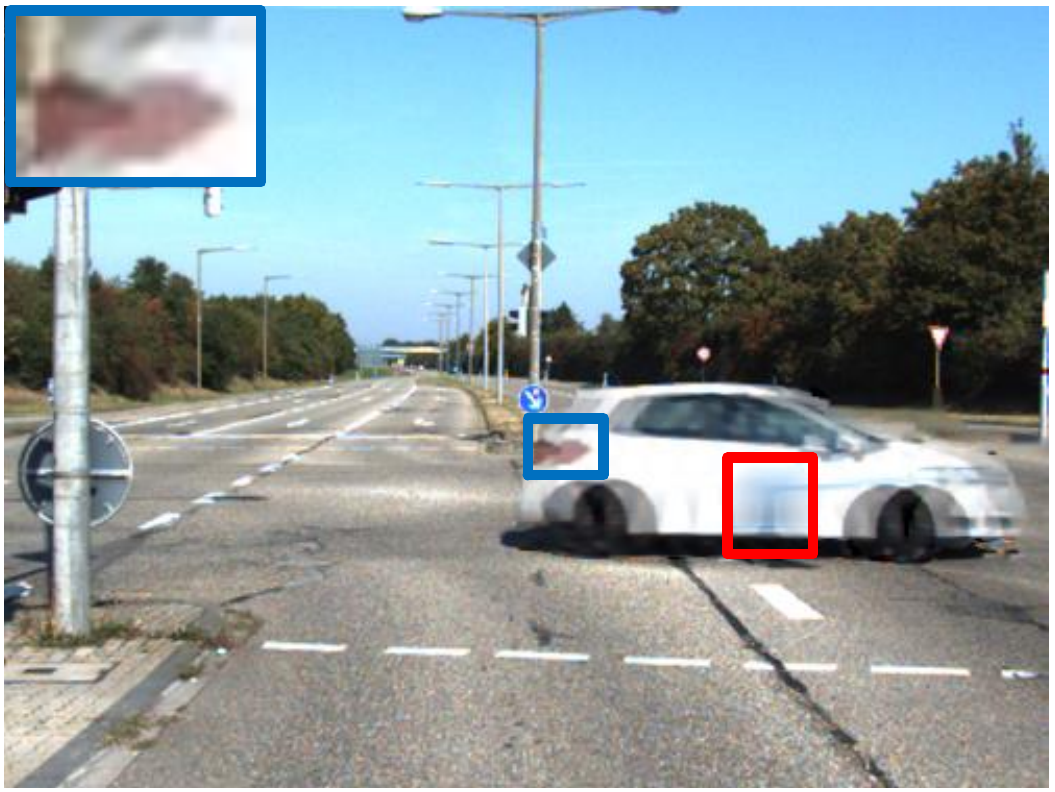}
&\hspace{-0.4cm}\includegraphics[width=0.164\textwidth]{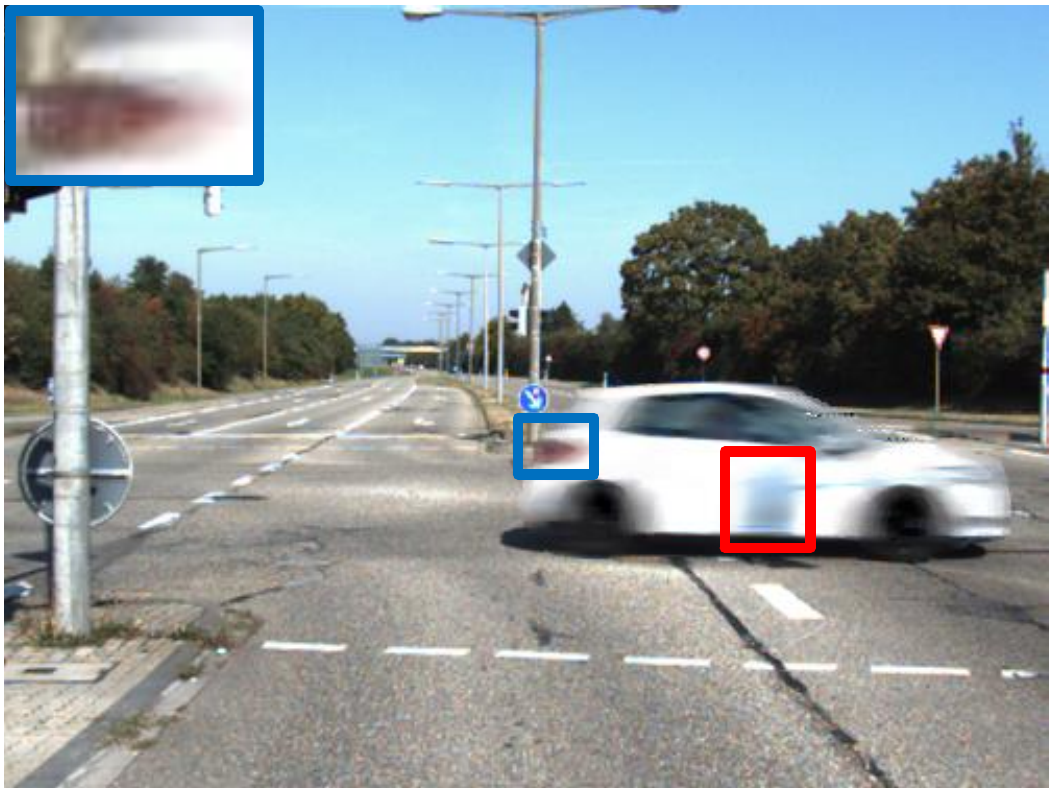}
&\hspace{-0.4cm}\includegraphics[width=0.164\textwidth]{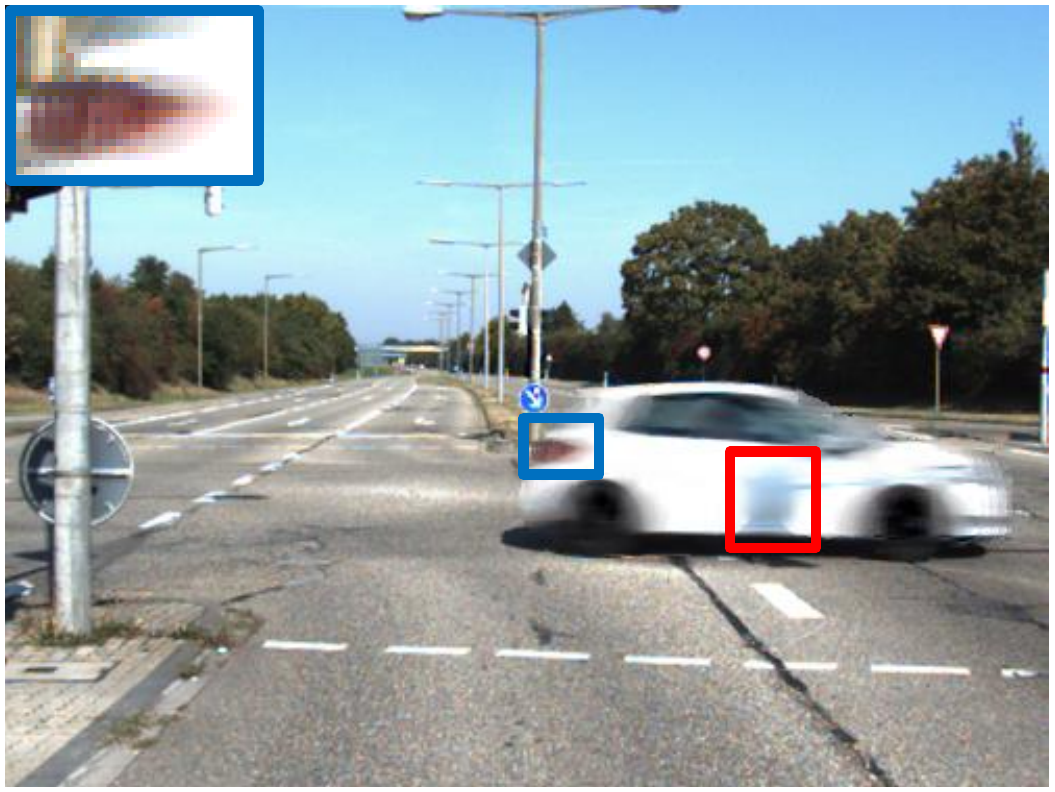}\\%[0.1in]
\hspace{-0.4cm}
\includegraphics[width=0.164\textwidth]{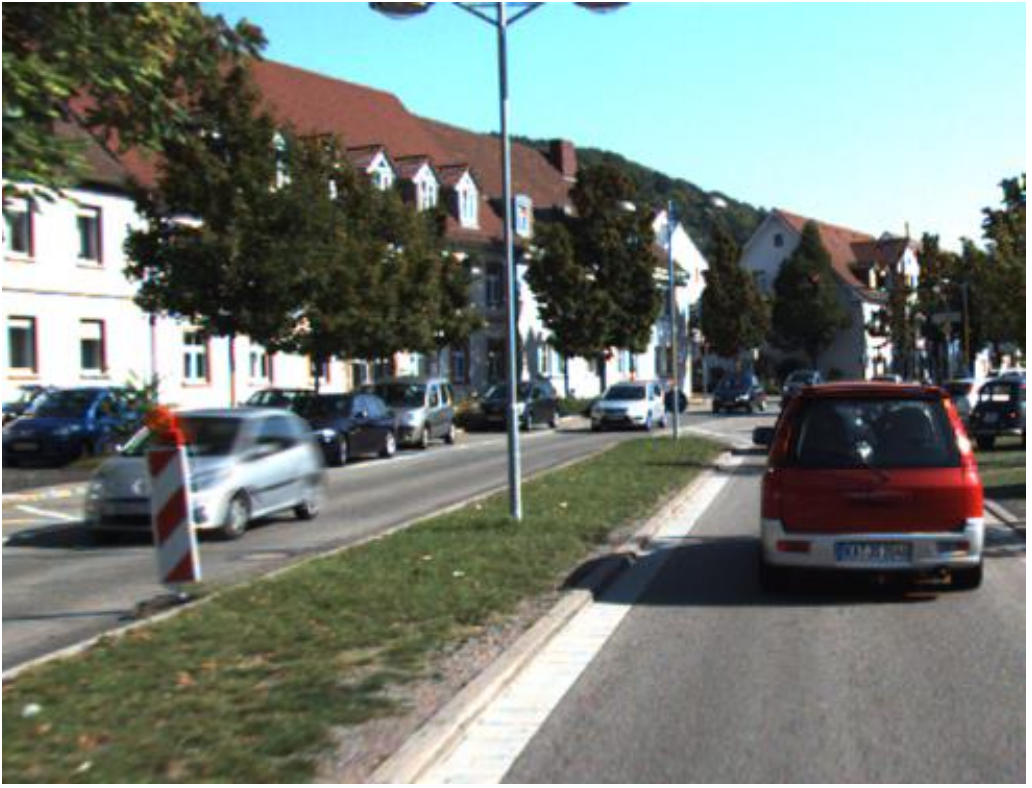}
&\hspace{-0.45cm}
\includegraphics[width=0.164\textwidth]{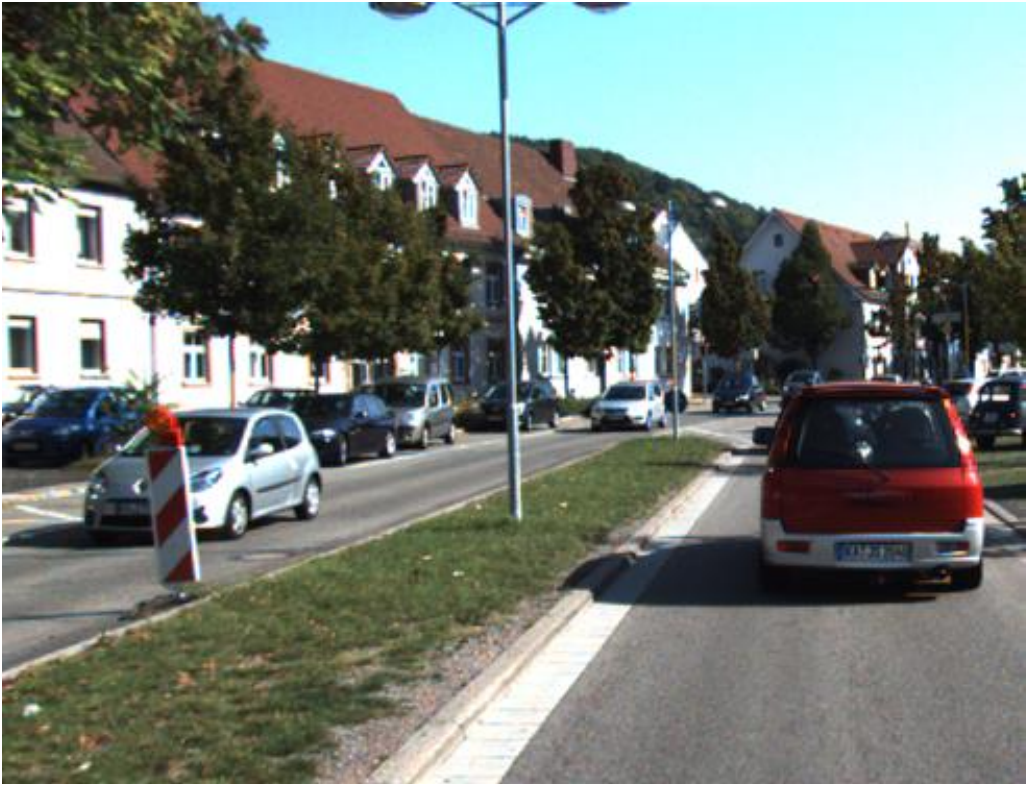}
&\hspace{-0.4cm}\includegraphics[width=0.164\textwidth]{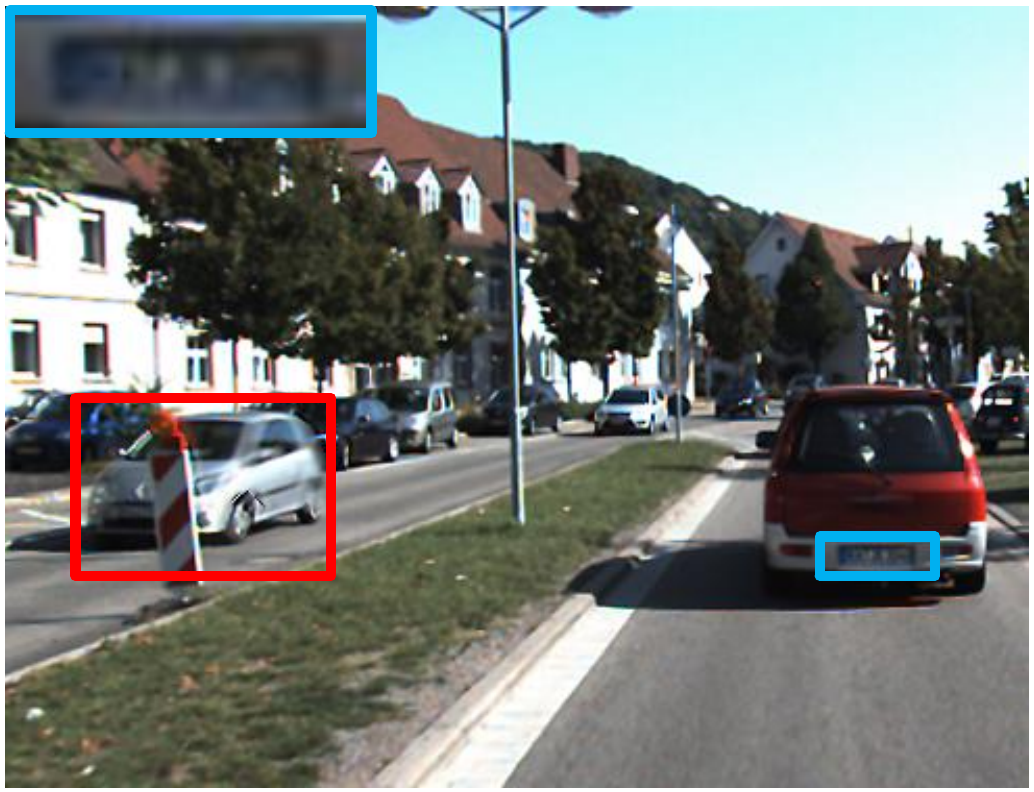}
&\hspace{-0.4cm}\includegraphics[width=0.164\textwidth]{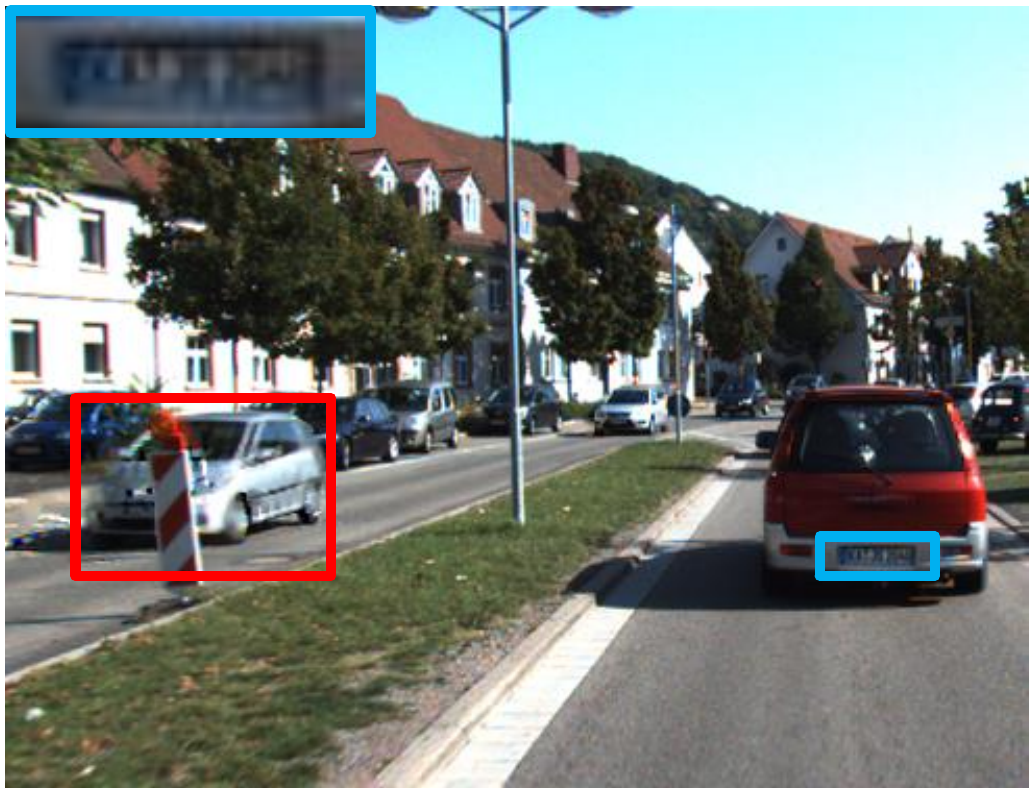}
&\hspace{-0.4cm}\includegraphics[width=0.164\textwidth]{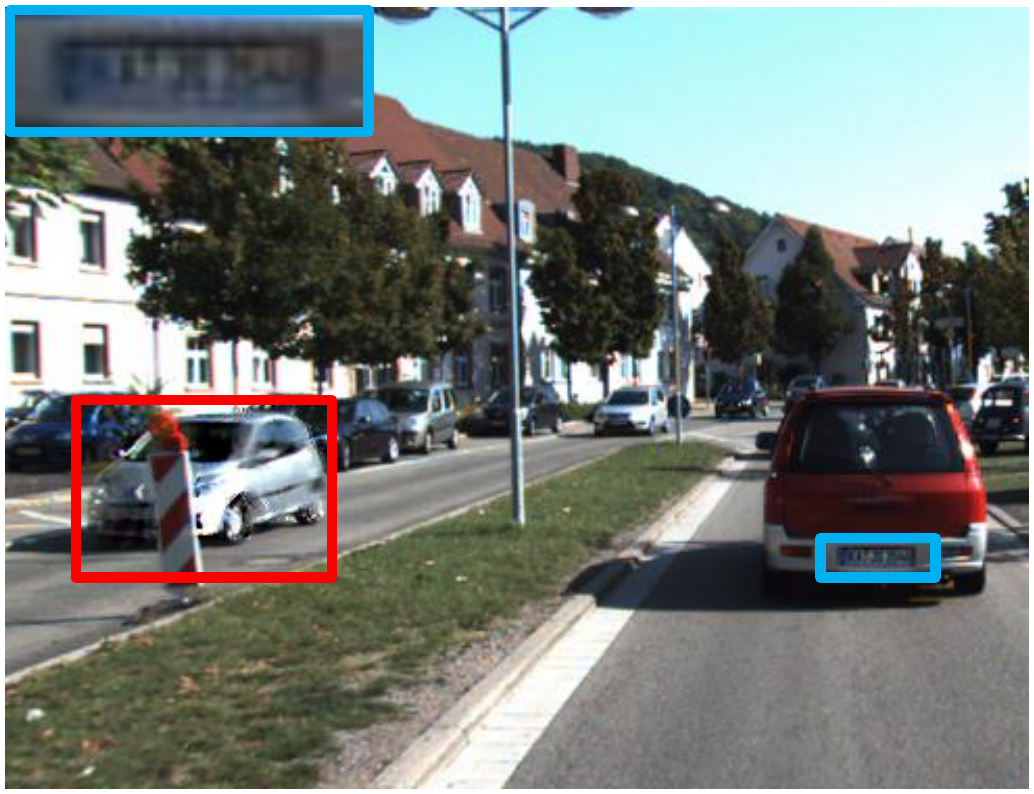}
&\hspace{-0.4cm}\includegraphics[width=0.164\textwidth]{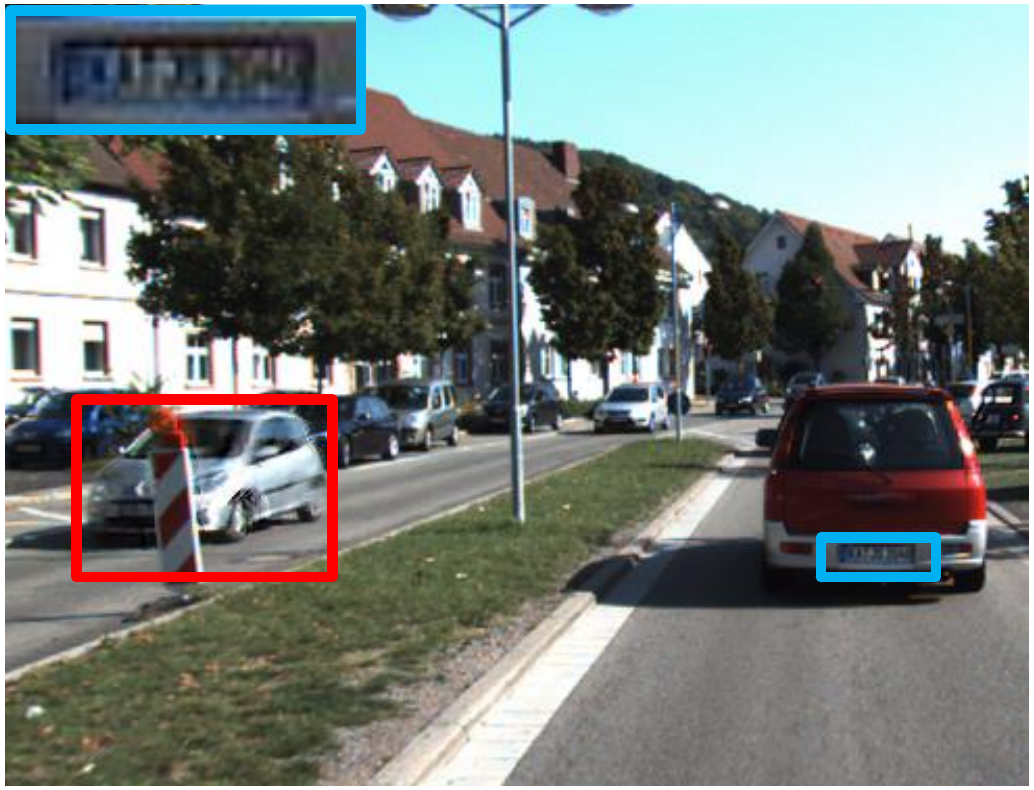}\\%[0.1in]
\hspace{-0.4cm}
\includegraphics[width=0.164\textwidth]{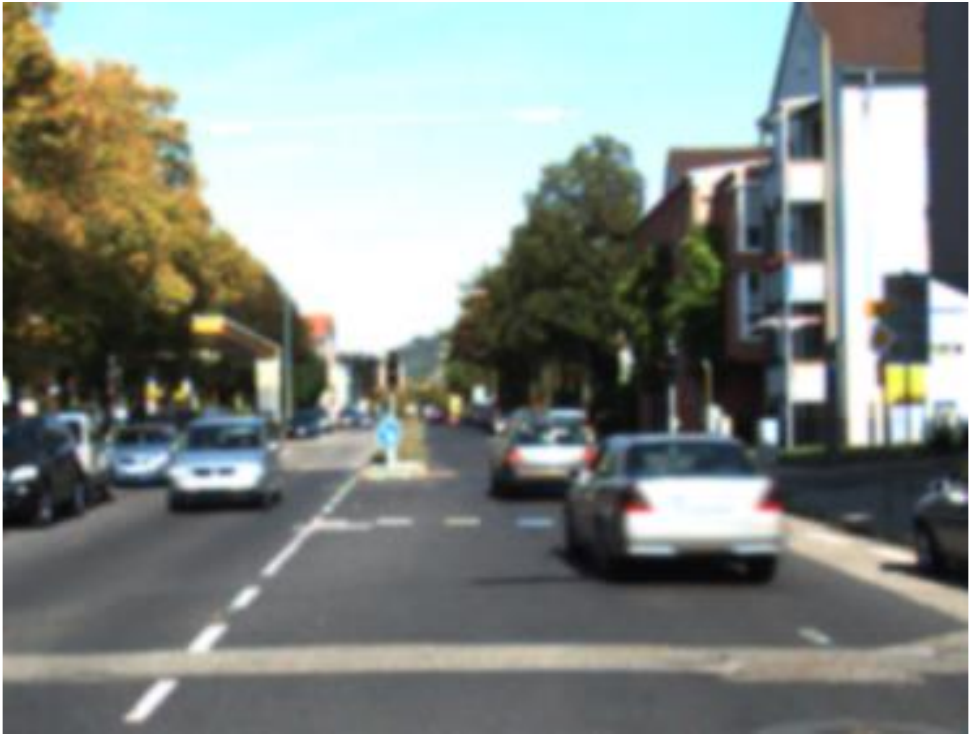}
&\hspace{-0.45cm}
\includegraphics[width=0.164\textwidth]{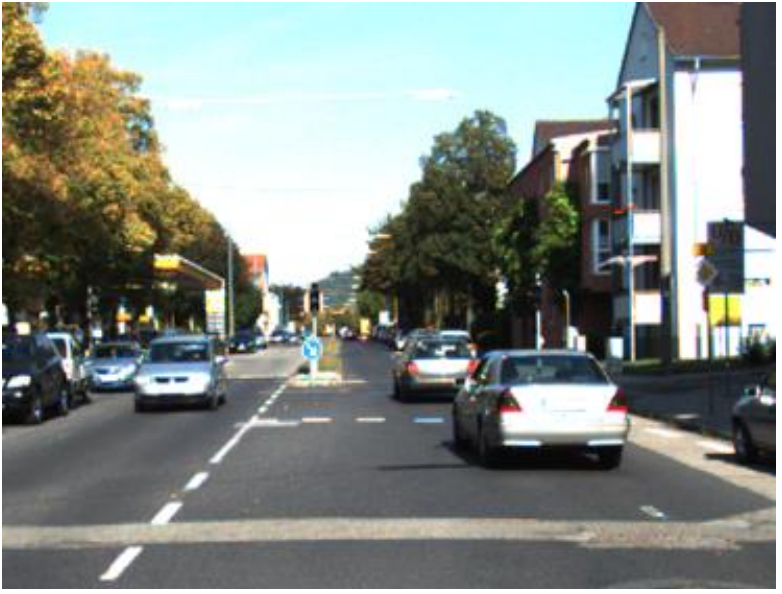}
&\hspace{-0.4cm}\includegraphics[width=0.164\textwidth]{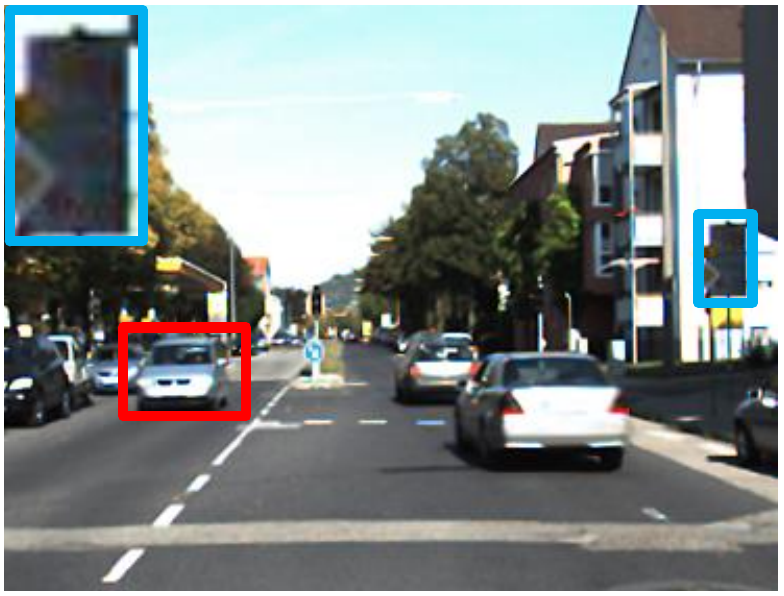}
&\hspace{-0.4cm}\includegraphics[width=0.164\textwidth]{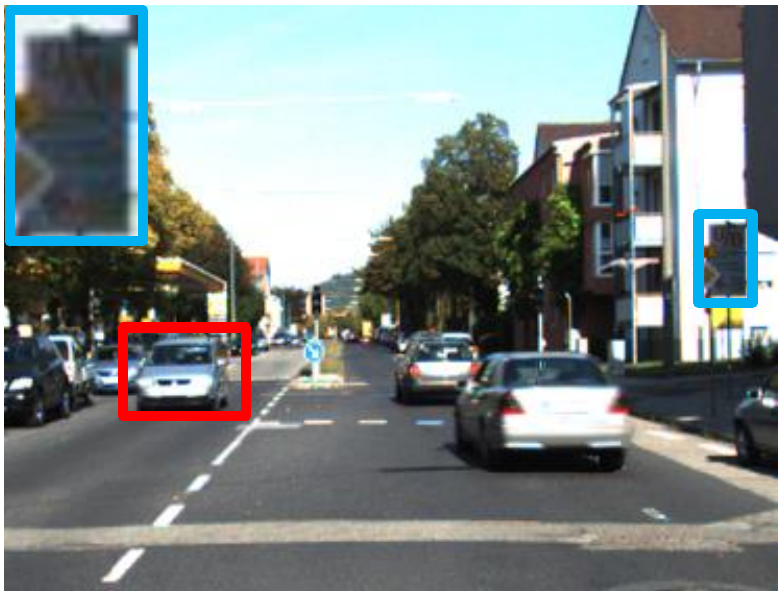}
&\hspace{-0.4cm}\includegraphics[width=0.164\textwidth]{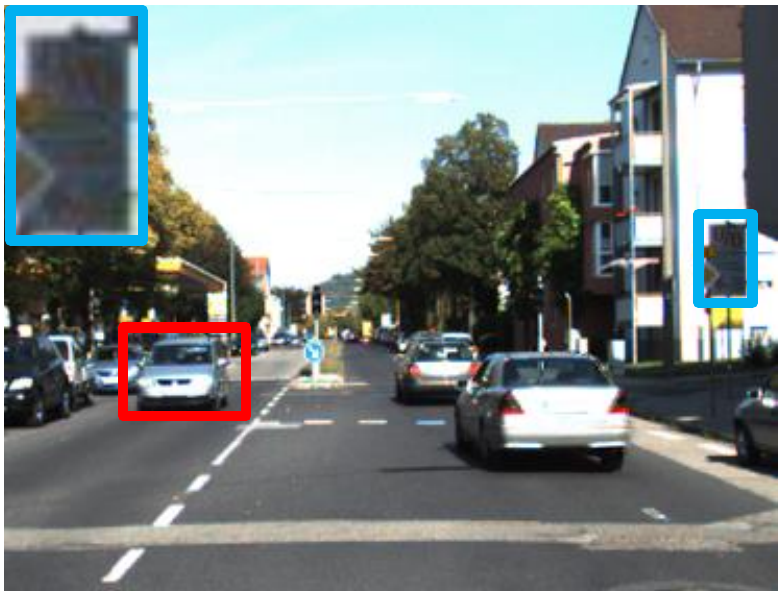}
&\hspace{-0.4cm}\includegraphics[width=0.164\textwidth]{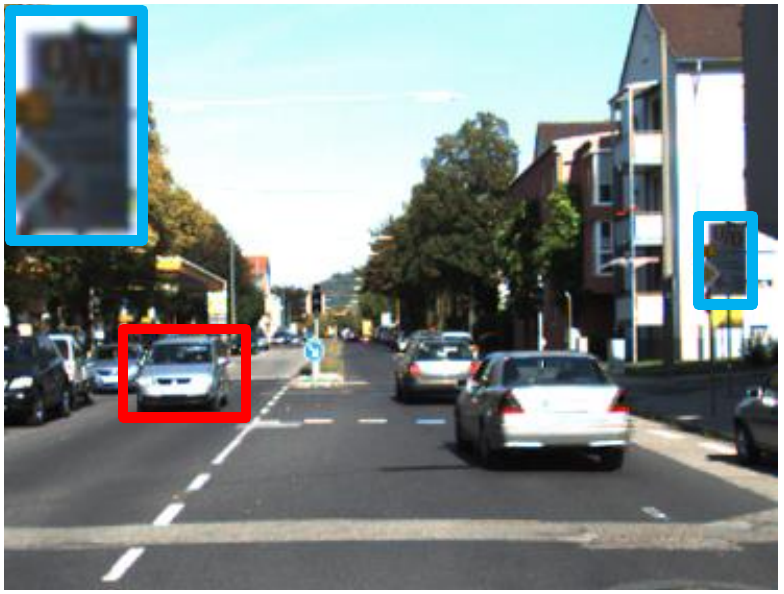}\\
\hspace{-0.4cm}
(a) Blurred image
&\hspace{-0.45cm}(b) GT image
&\hspace{-0.4cm}(c) \rc{Kim and Lee~\cite{hyun2015generalized}}
&\hspace{-0.4cm}(d) \rc{Sellent~\etal~\cite{sellent2016stereo}}
&\hspace{-0.4cm}(e) \rc{Pan~\etal~\cite{Pan_2017_CVPR}}
&\hspace{-0.4cm}(f) \rc{Ours}\\
\hspace{-0.4cm}
\includegraphics[width=0.164\textwidth]{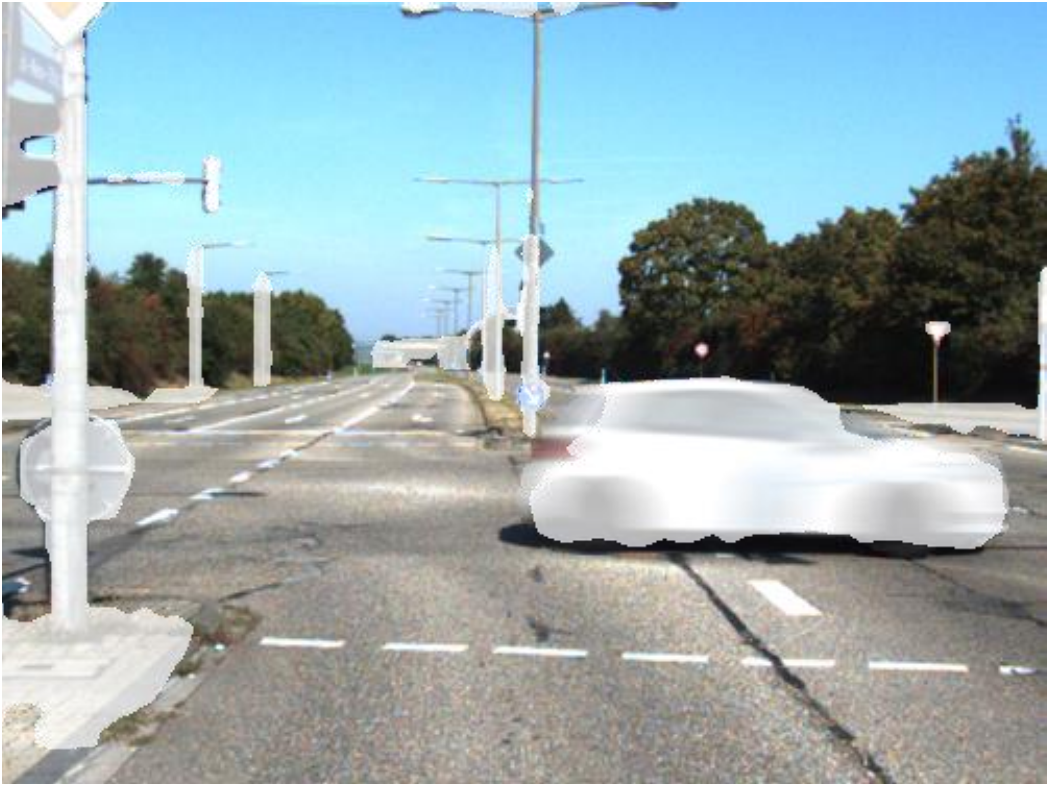}
&\hspace{-0.45cm}
\includegraphics[width=0.164\textwidth]{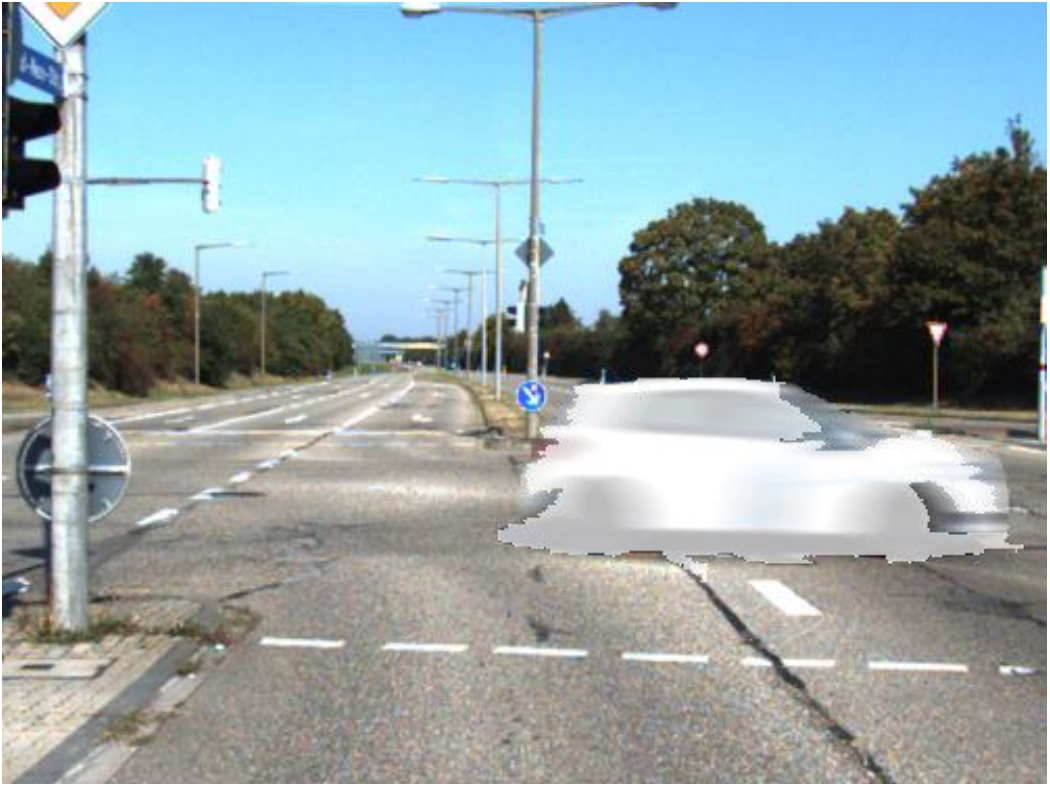}
&\hspace{-0.4cm}\includegraphics[width=0.164\textwidth]{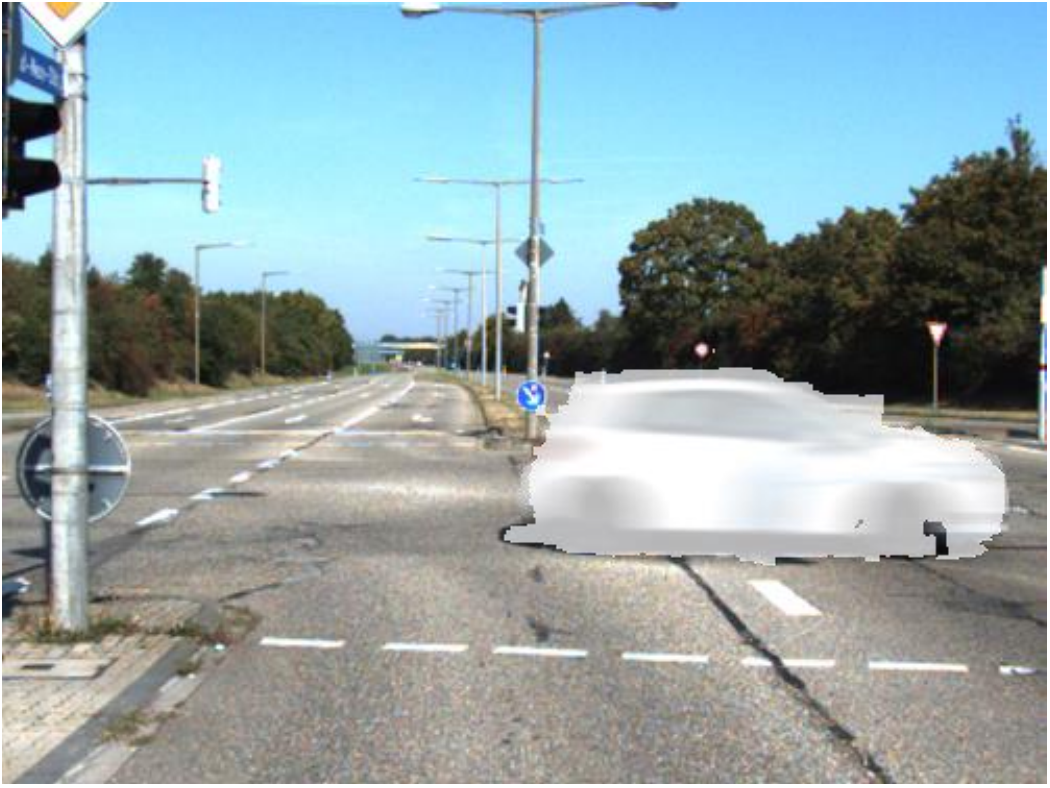}
&\hspace{-0.4cm}\includegraphics[width=0.164\textwidth]{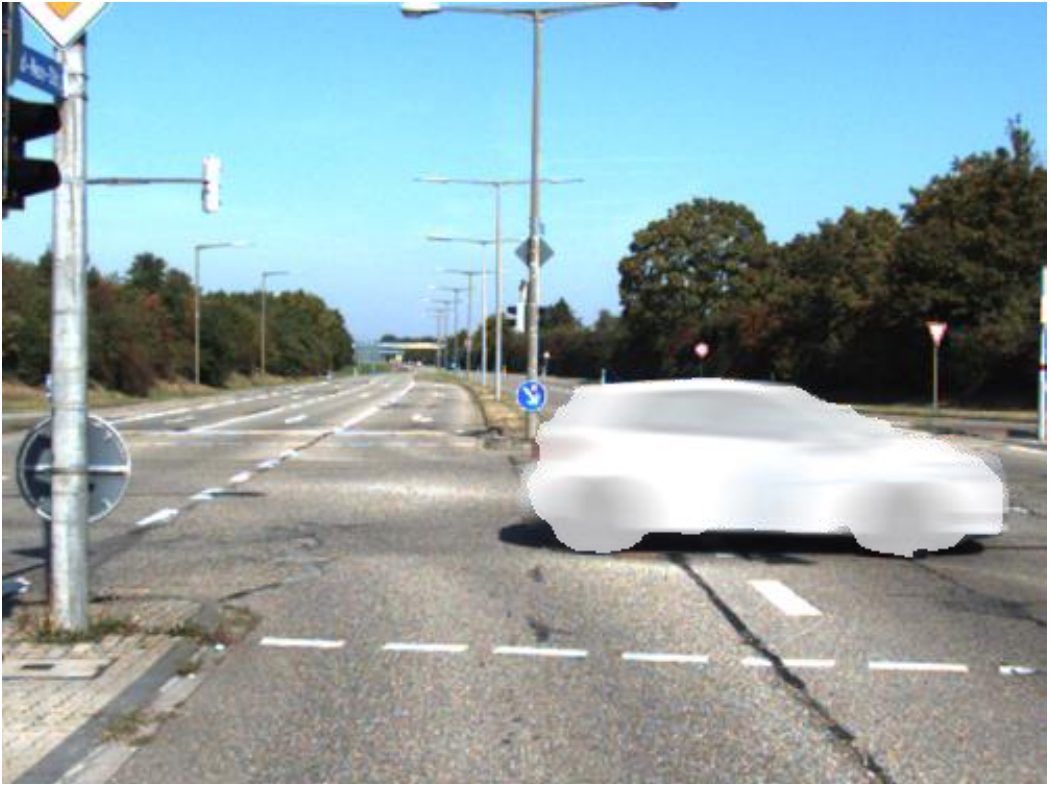}
&\hspace{-0.4cm}\includegraphics[width=0.164\textwidth]{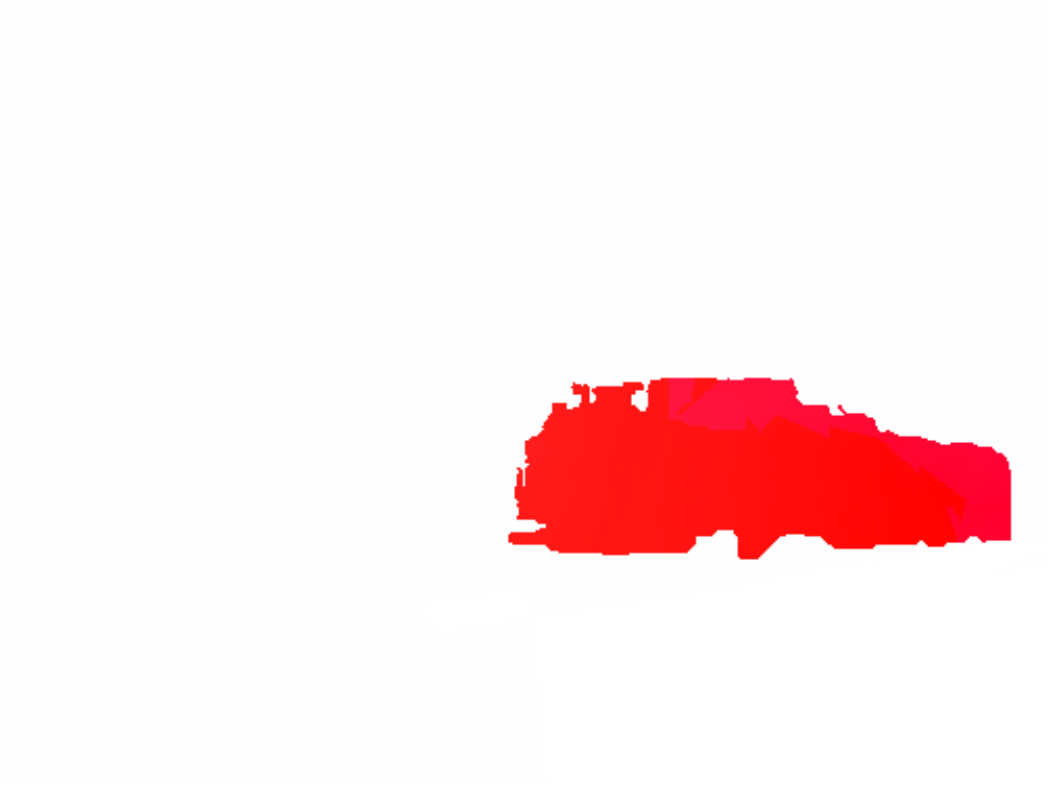}
&\hspace{-0.4cm}\includegraphics[width=0.164\textwidth]{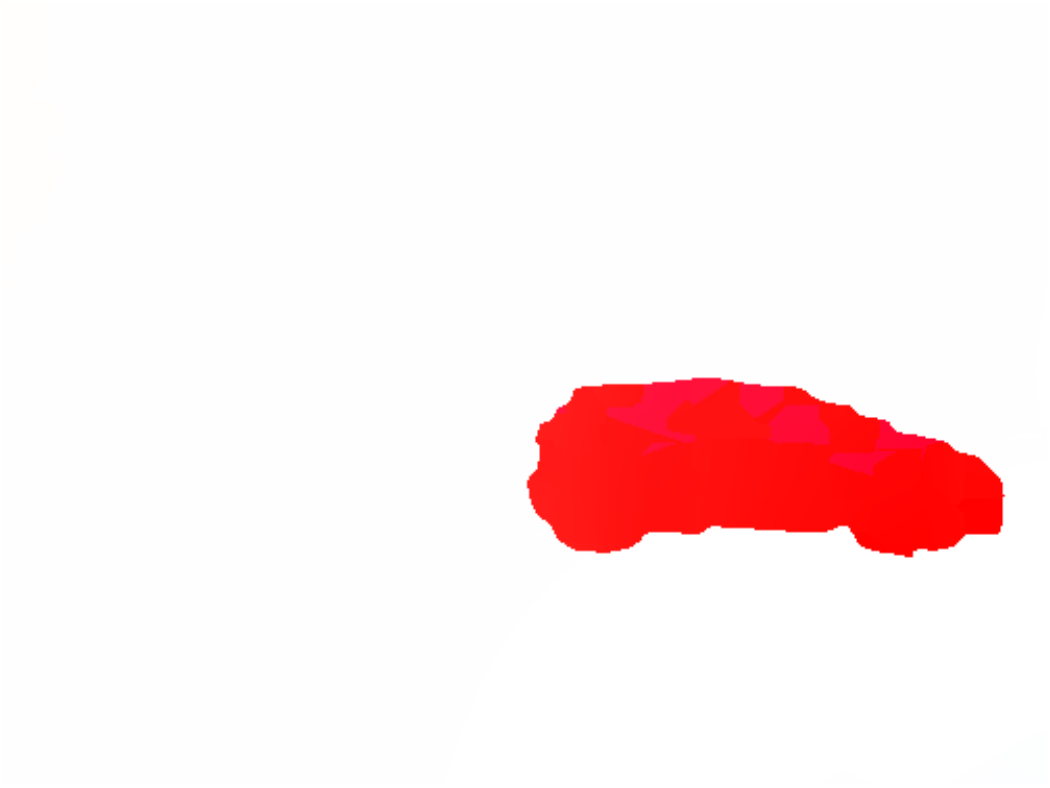}\\%[0.1in]
\hspace{-0.4cm}
\includegraphics[width=0.164\textwidth]{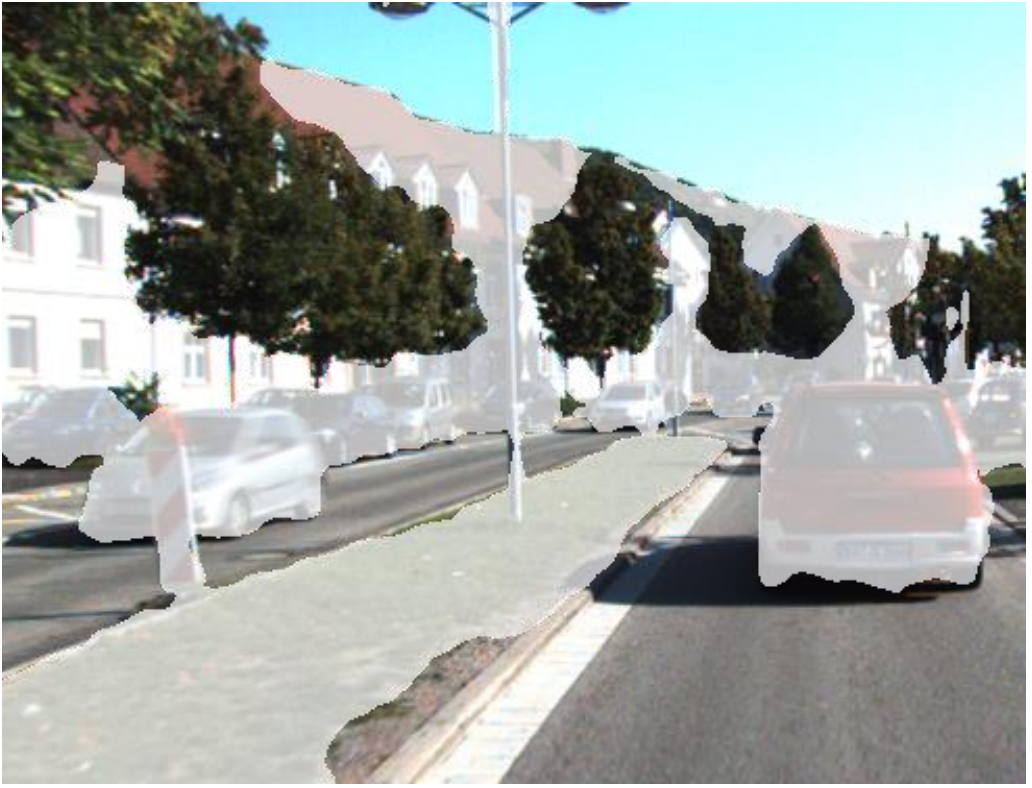}
&\hspace{-0.45cm}
\includegraphics[width=0.164\textwidth]{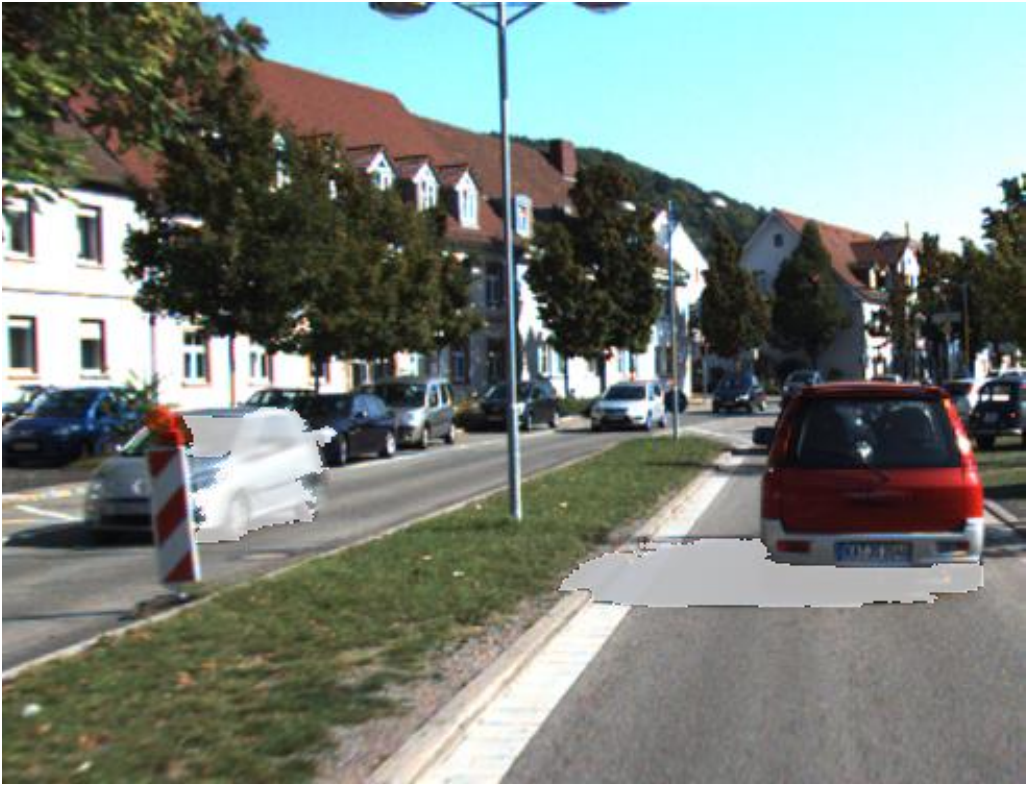}
&\hspace{-0.4cm}\includegraphics[width=0.164\textwidth]{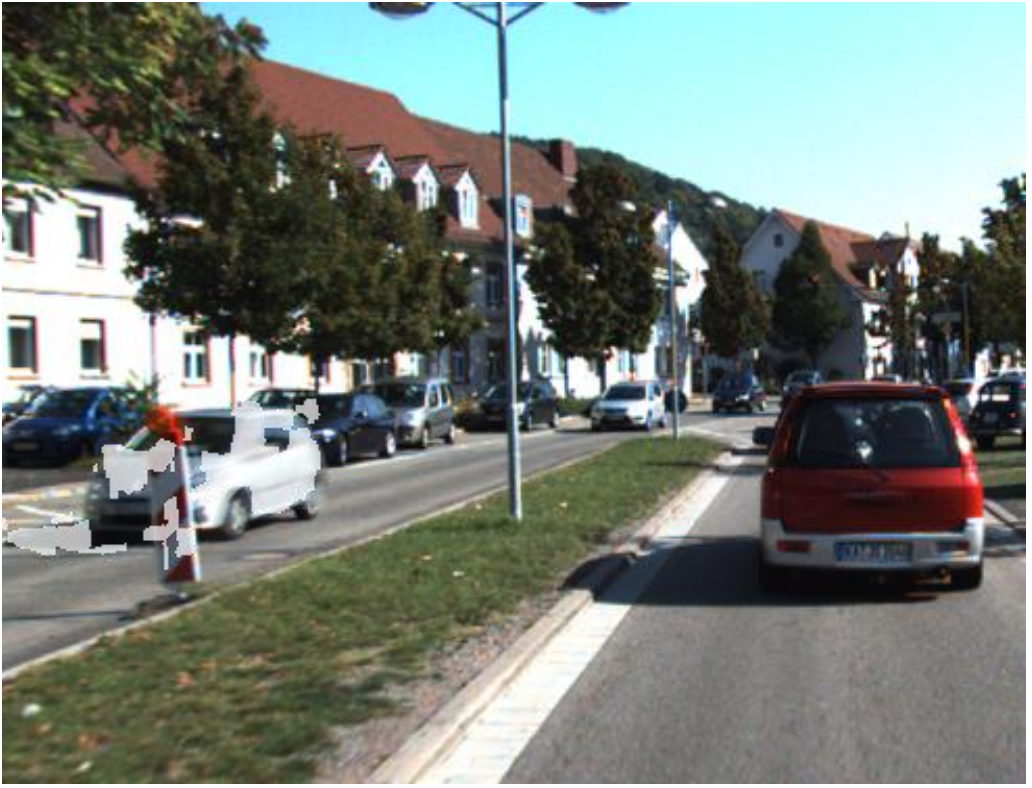}
&\hspace{-0.4cm}\includegraphics[width=0.164\textwidth]{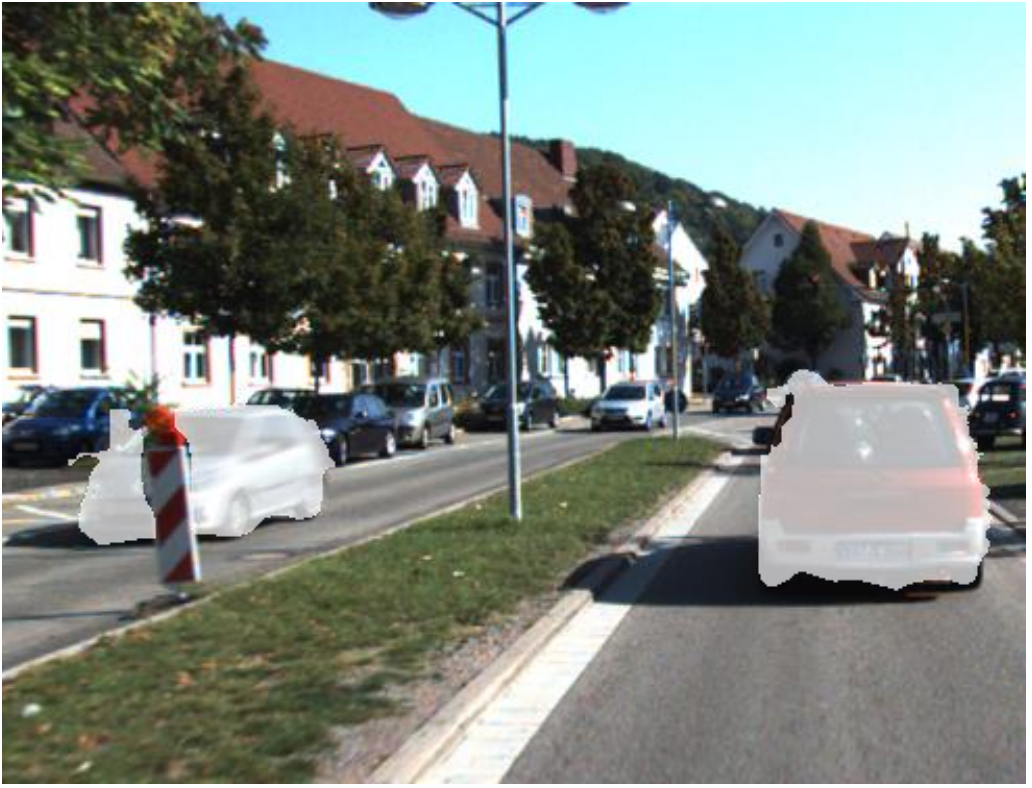}
&\hspace{-0.4cm}\includegraphics[width=0.164\textwidth]{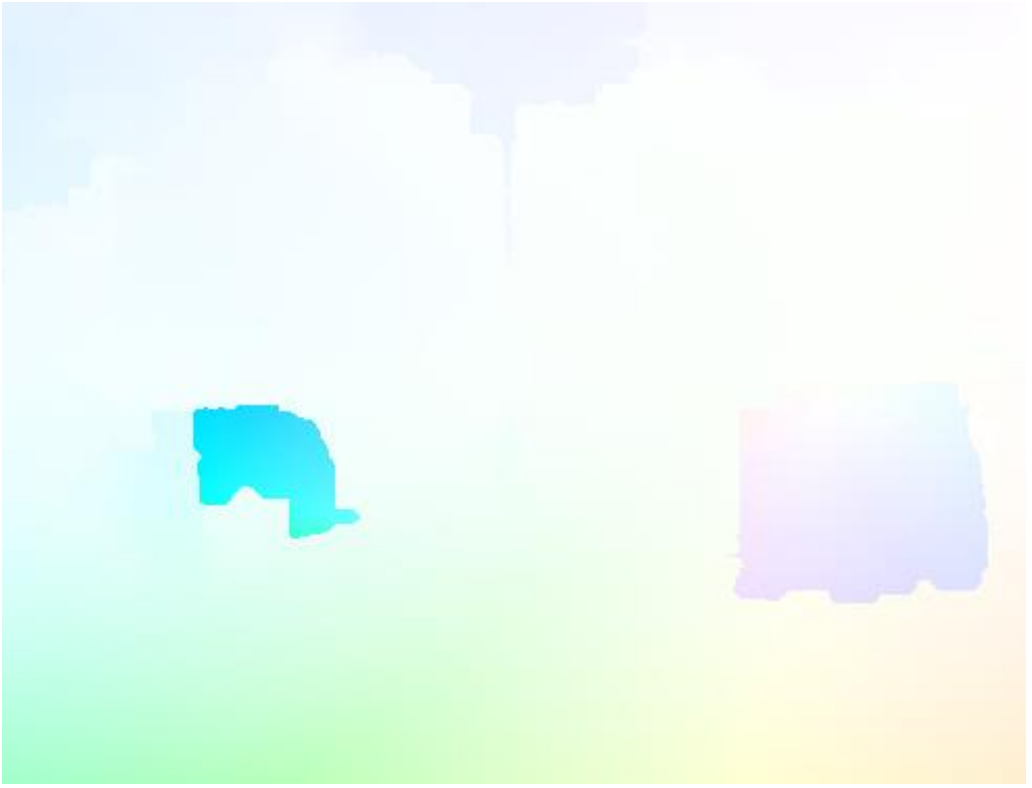}
&\hspace{-0.4cm}\includegraphics[width=0.164\textwidth]{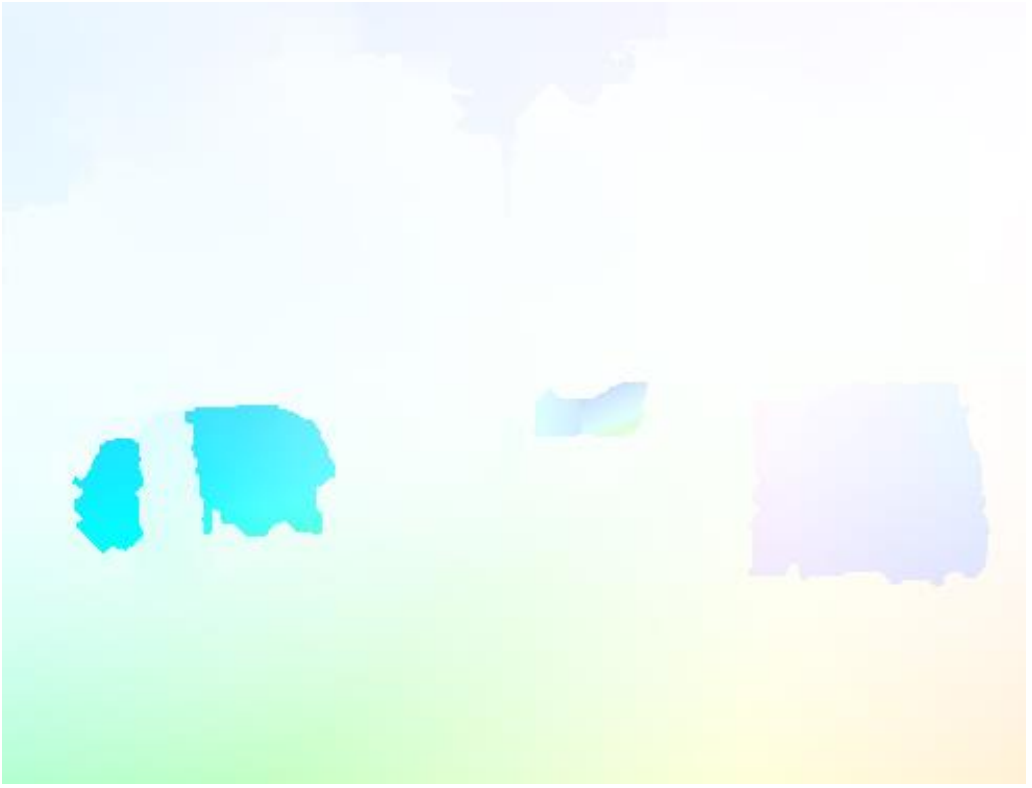}\\%[0.1in]
\hspace{-0.4cm}
\includegraphics[width=0.164\textwidth]{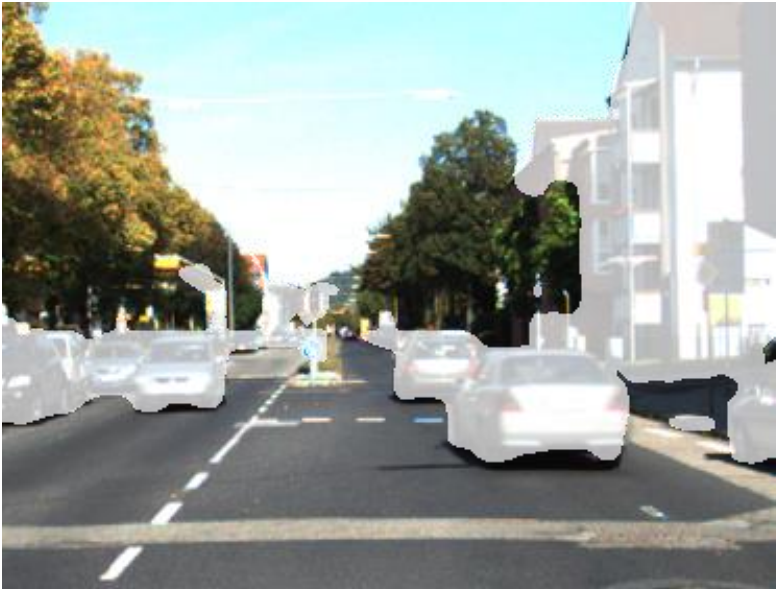}
&\hspace{-0.45cm}
\includegraphics[width=0.164\textwidth]{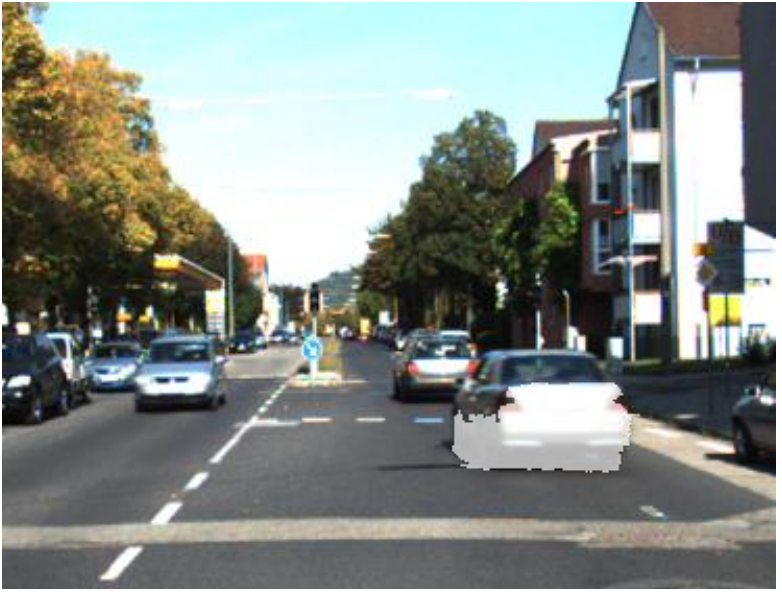}
&\hspace{-0.4cm}\includegraphics[width=0.164\textwidth]{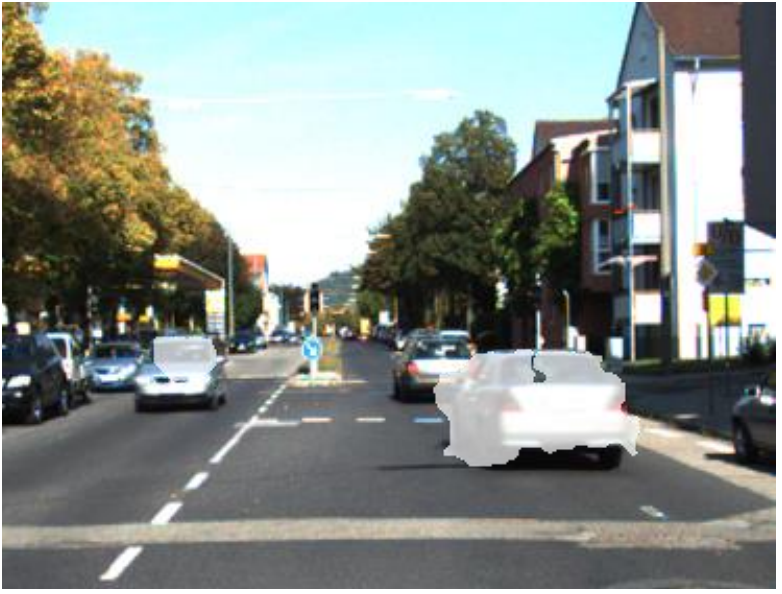}
&\hspace{-0.4cm}\includegraphics[width=0.164\textwidth]{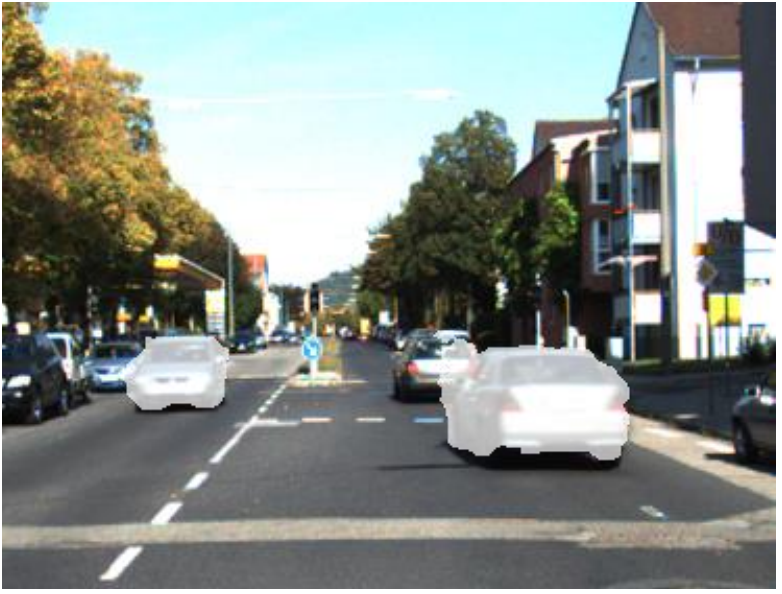}
&\hspace{-0.4cm}\includegraphics[width=0.164\textwidth]{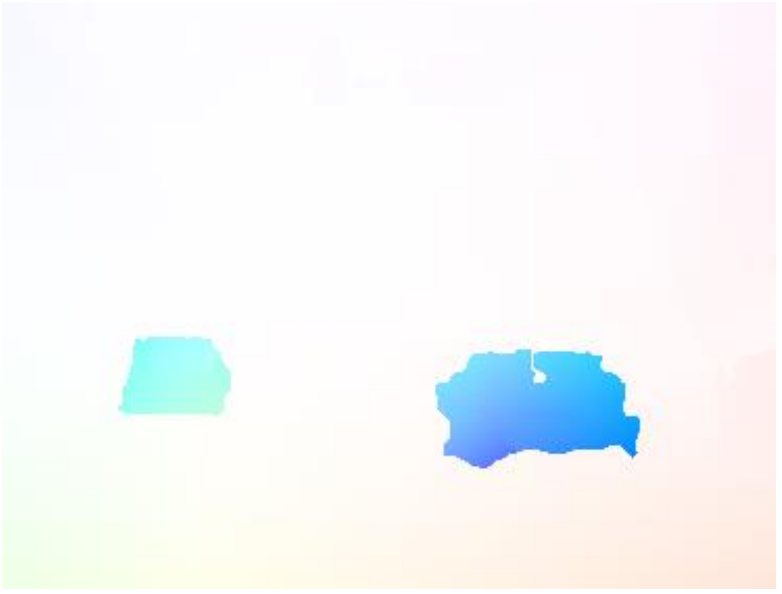}
&\hspace{-0.4cm}\includegraphics[width=0.164\textwidth]{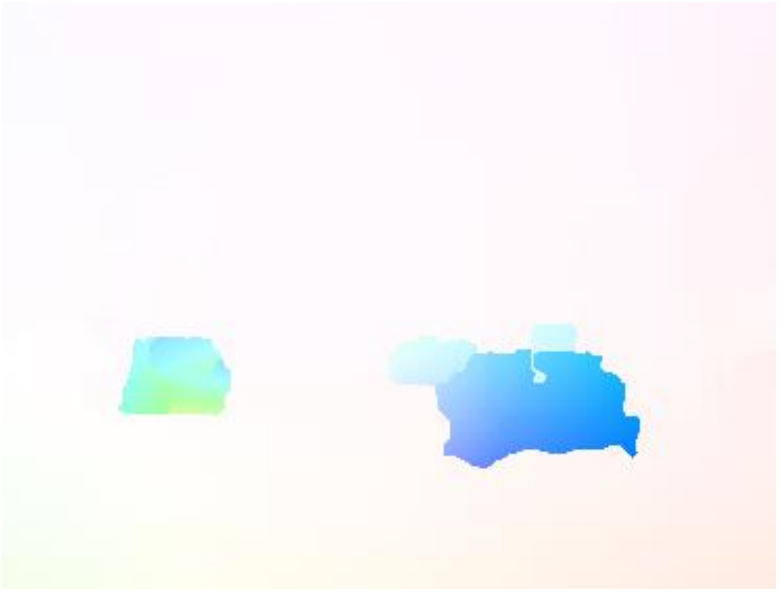}\\
\hspace{-0.4cm}
(g) Initial mask
&\hspace{-0.45cm} (h) FST~\cite{papazoglou2013fast}
&\hspace{-0.4cm}(i) NLC  ~\cite{faktor2014video}
&\hspace{-0.4cm}(j) \rc{Ours}
&\hspace{-0.4cm}(k) Pan~\etal~\cite{Pan_2017_CVPR}
&\hspace{-0.4cm}(l) Ours
\end{tabular}
% }
% \vspace{-1 mm}
\caption{\rc{Qualitative comparison of our approach with baselines for deblurring, Moving object segmentation, and flow estimations. Our method use (a) blurred image and (g) Initial semantic prior from {\bf BlurData-1} as input. (b) Ground-truth latent image. (c) Deblurring results by Kim and Lee~\cite{hyun2015generalized}. (d) Stereo deblurring results by Sellent \etal~\cite{sellent2016stereo}. (e) and (f) show our deblurring results w/o imposing semantic priors, respectively; (h) Segmentation result by~\cite{papazoglou2013fast}.  (i) Segmentation result by~\cite{faktor2014video}. (j) Our segmentation result. (k) and (l) show the optical flow estimation results w/o imposing semantic priors. Best viewed in colour on the screen.}
}
\label{fig:adddeblurresult}
\end{center}
\end{figure*}

%================= fig: real data from eccv16============
\begin{figure*}
\begin{center}
\resizebox{\textwidth}{!}{
\begin{tabular}{cccc}
\includegraphics[width=0.244\textwidth]{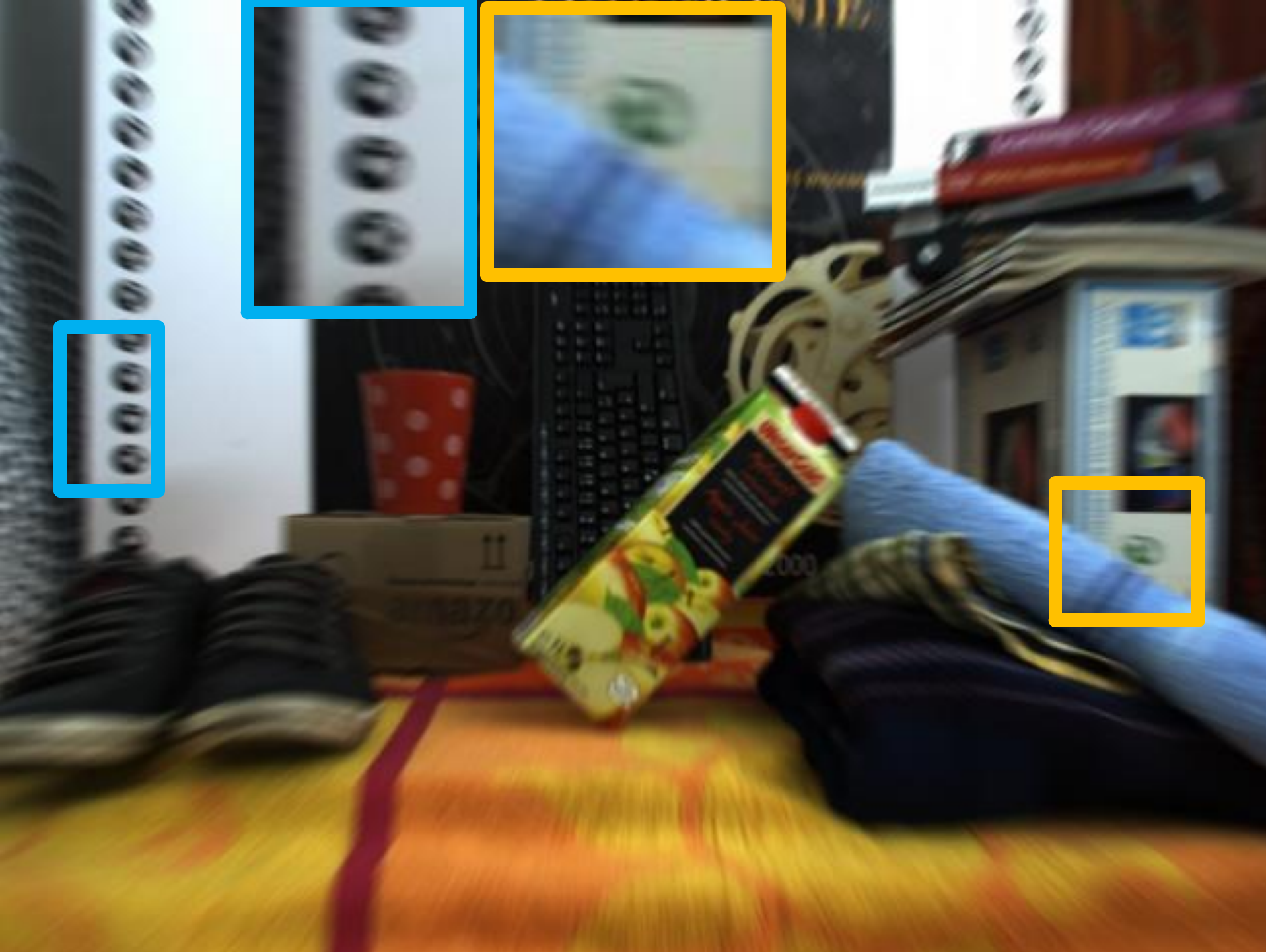}
&\includegraphics[width=0.244\textwidth]{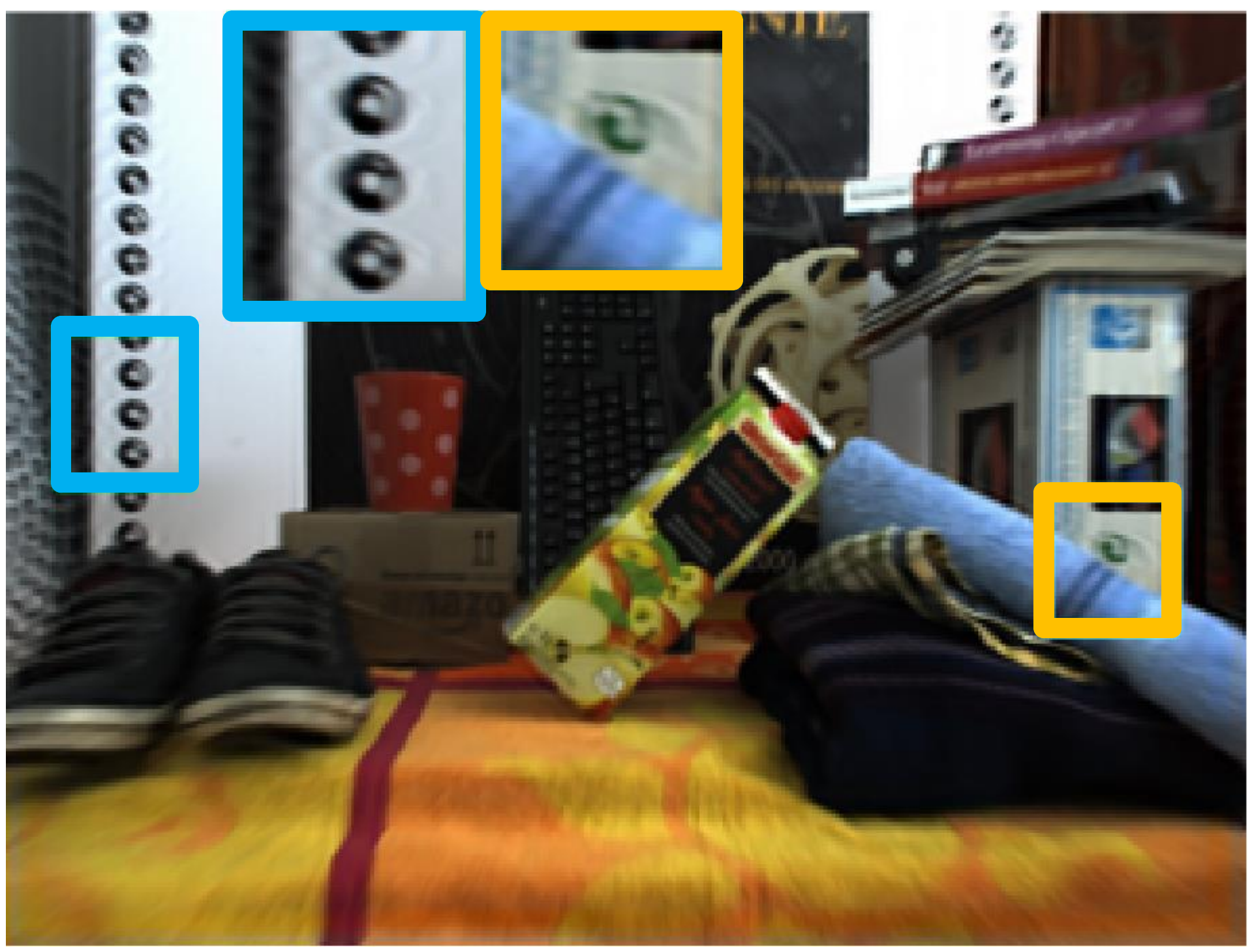}
&\includegraphics[width=0.244\textwidth]{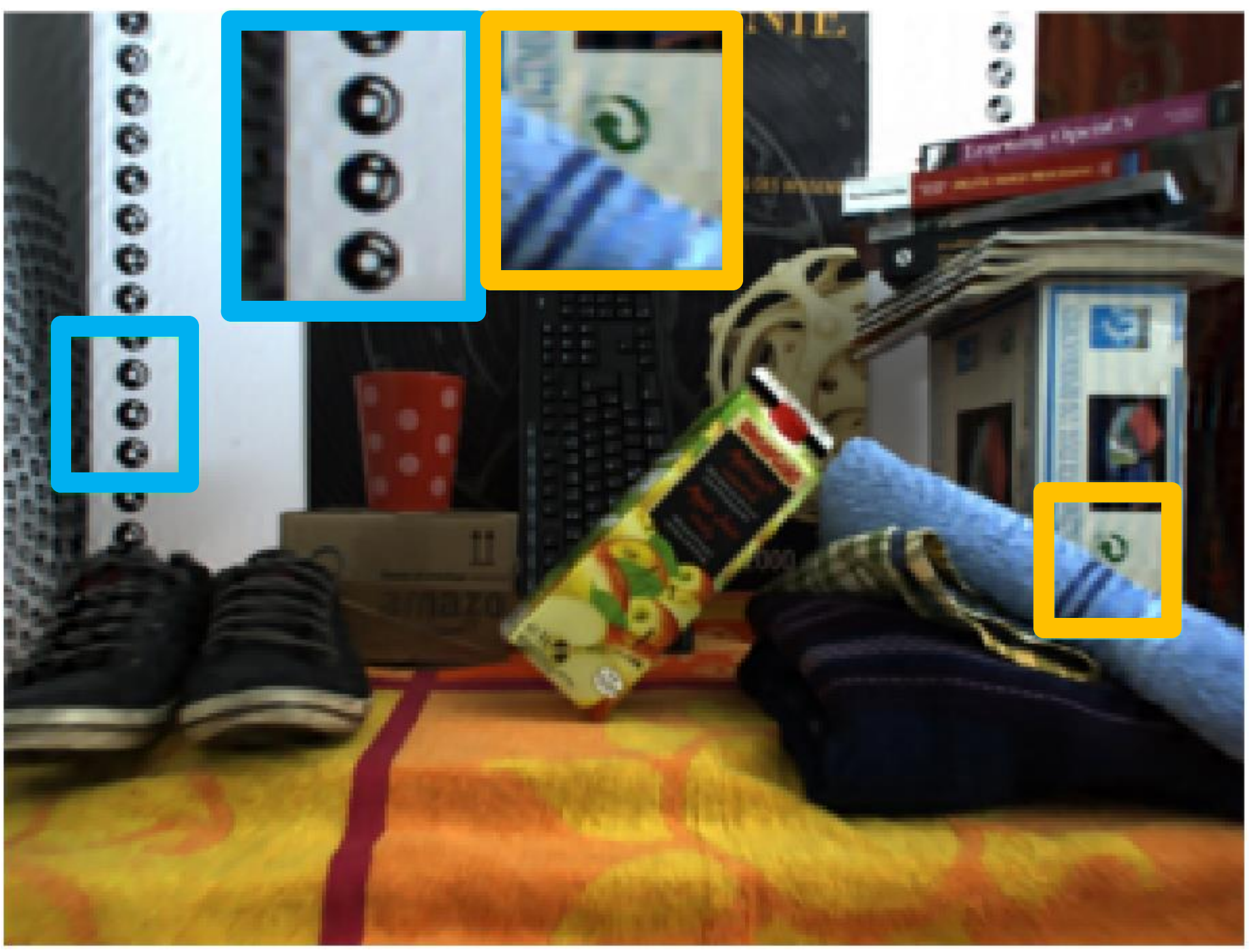}
&\includegraphics[width=0.244\textwidth]{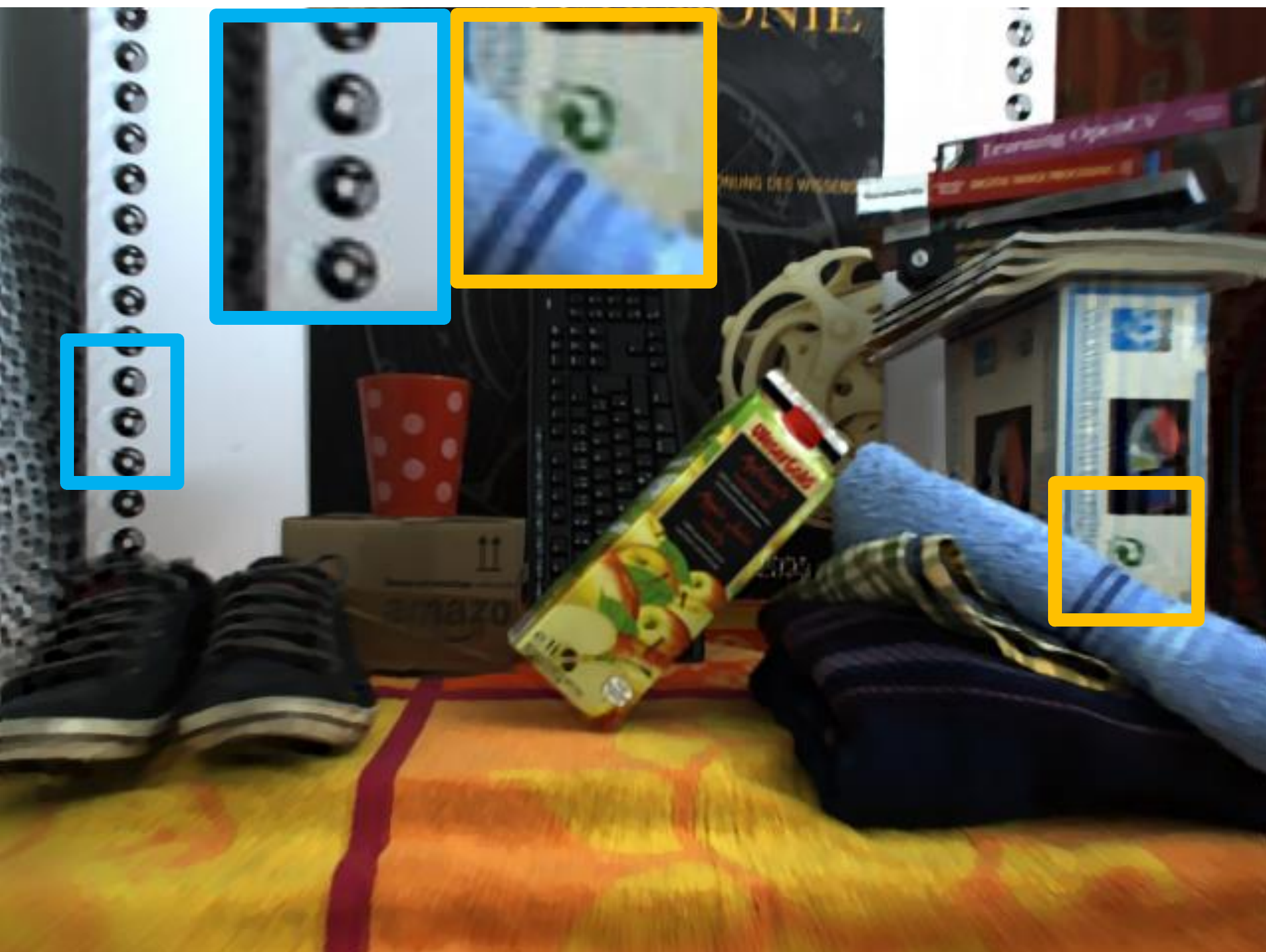}\\%[0.1in]
\includegraphics[width=0.244\textwidth]{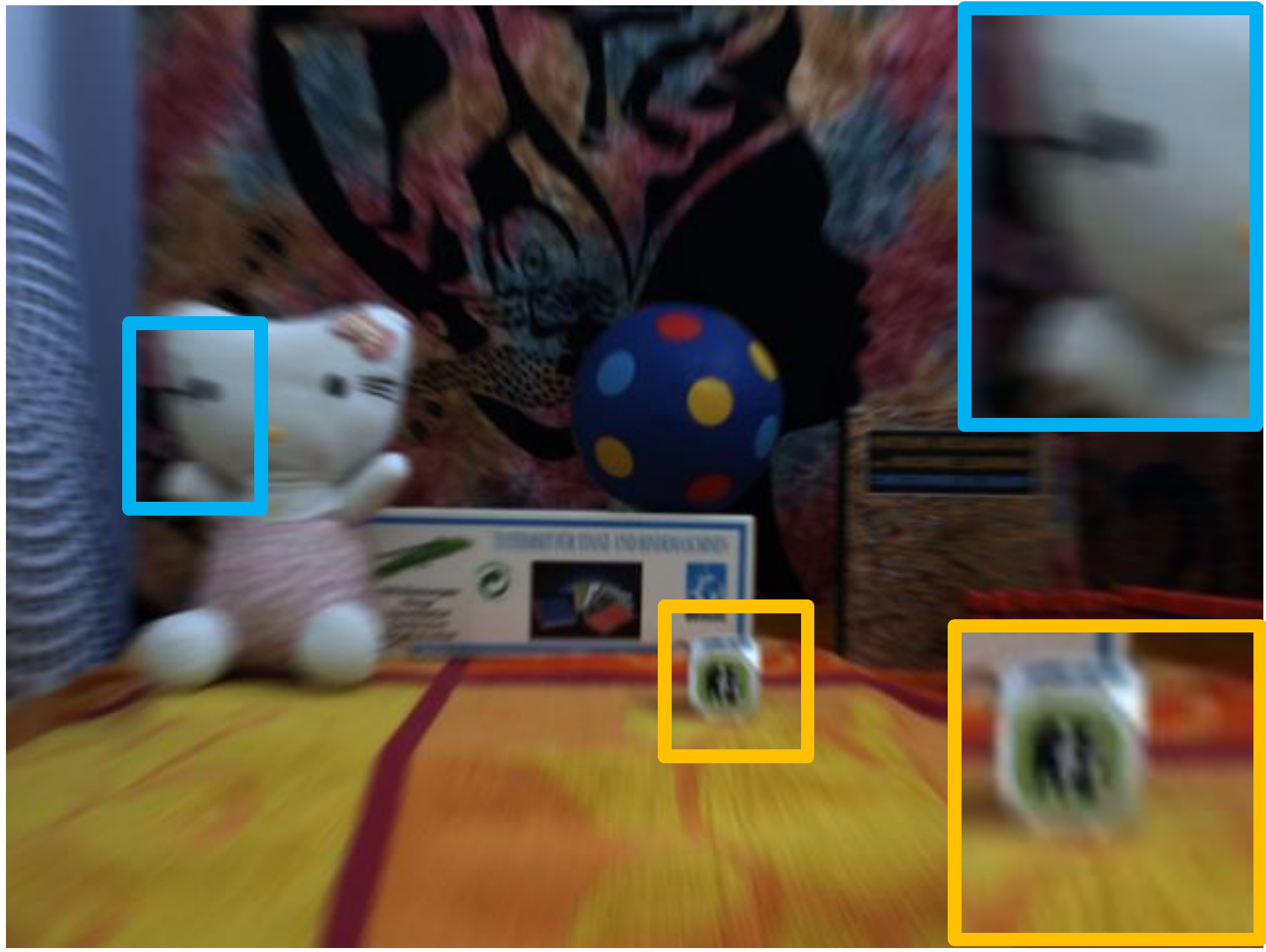}
&\includegraphics[width=0.244\textwidth]{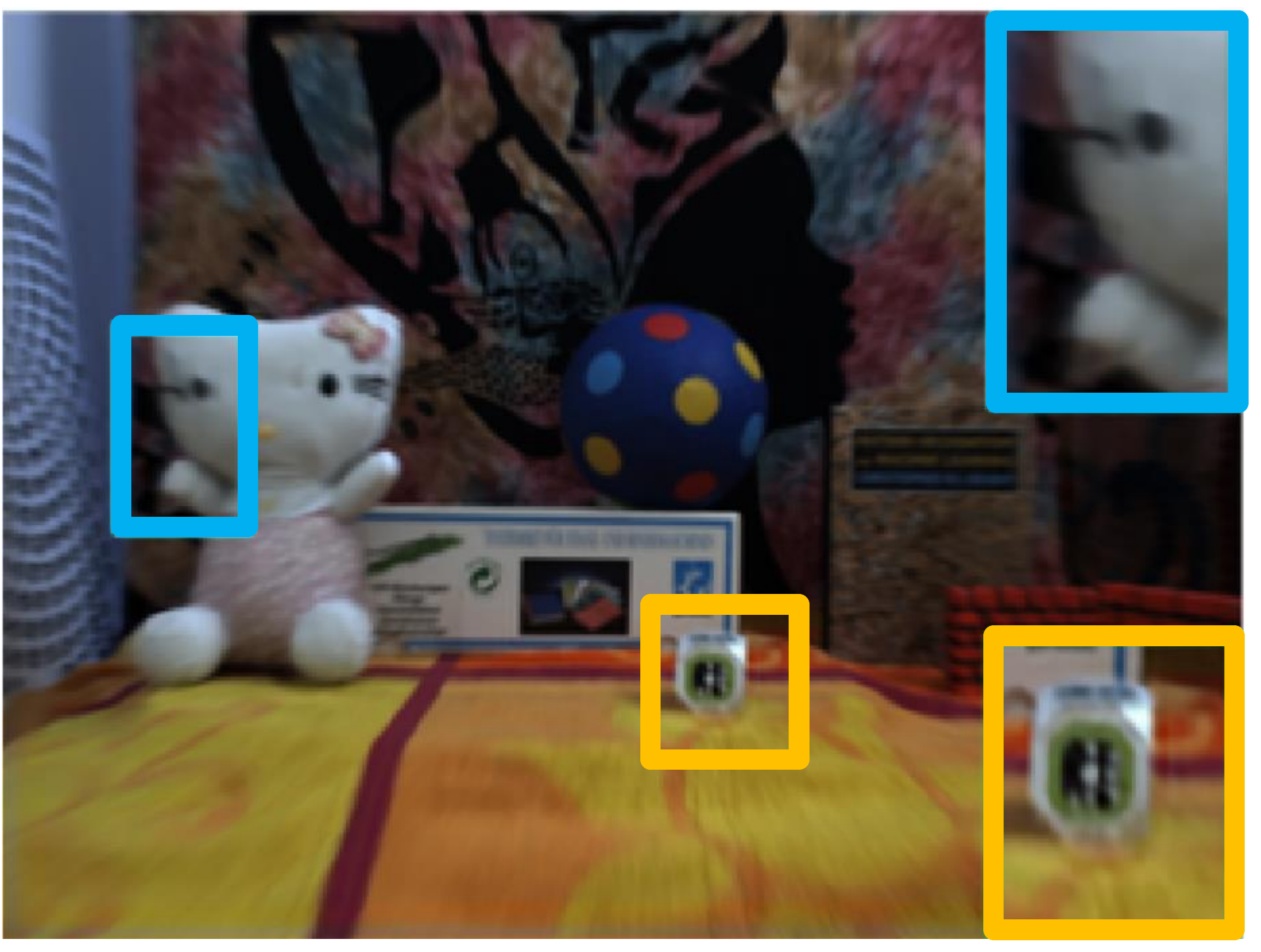}
&\includegraphics[width=0.244\textwidth]{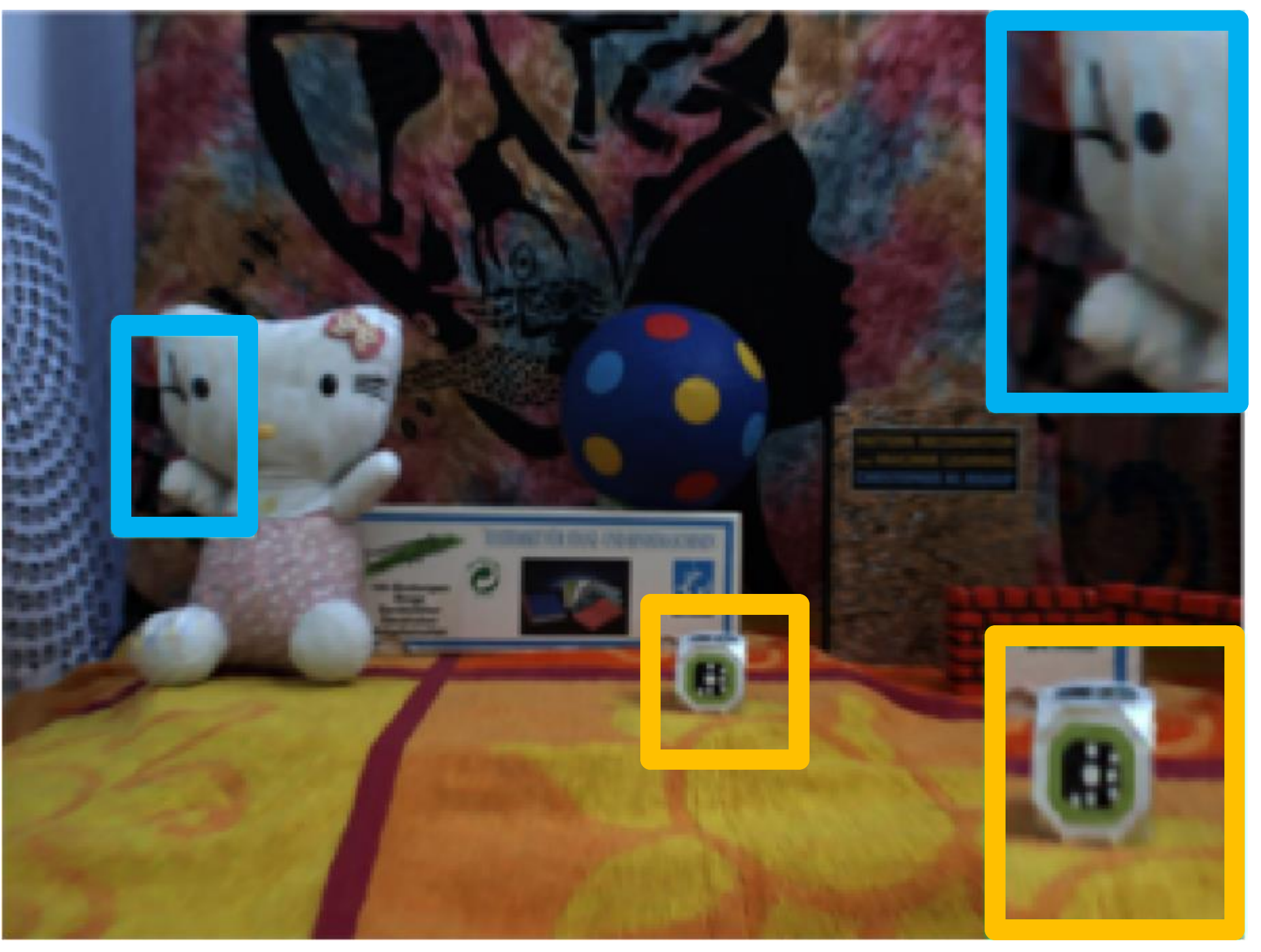}
&\includegraphics[width=0.244\textwidth]{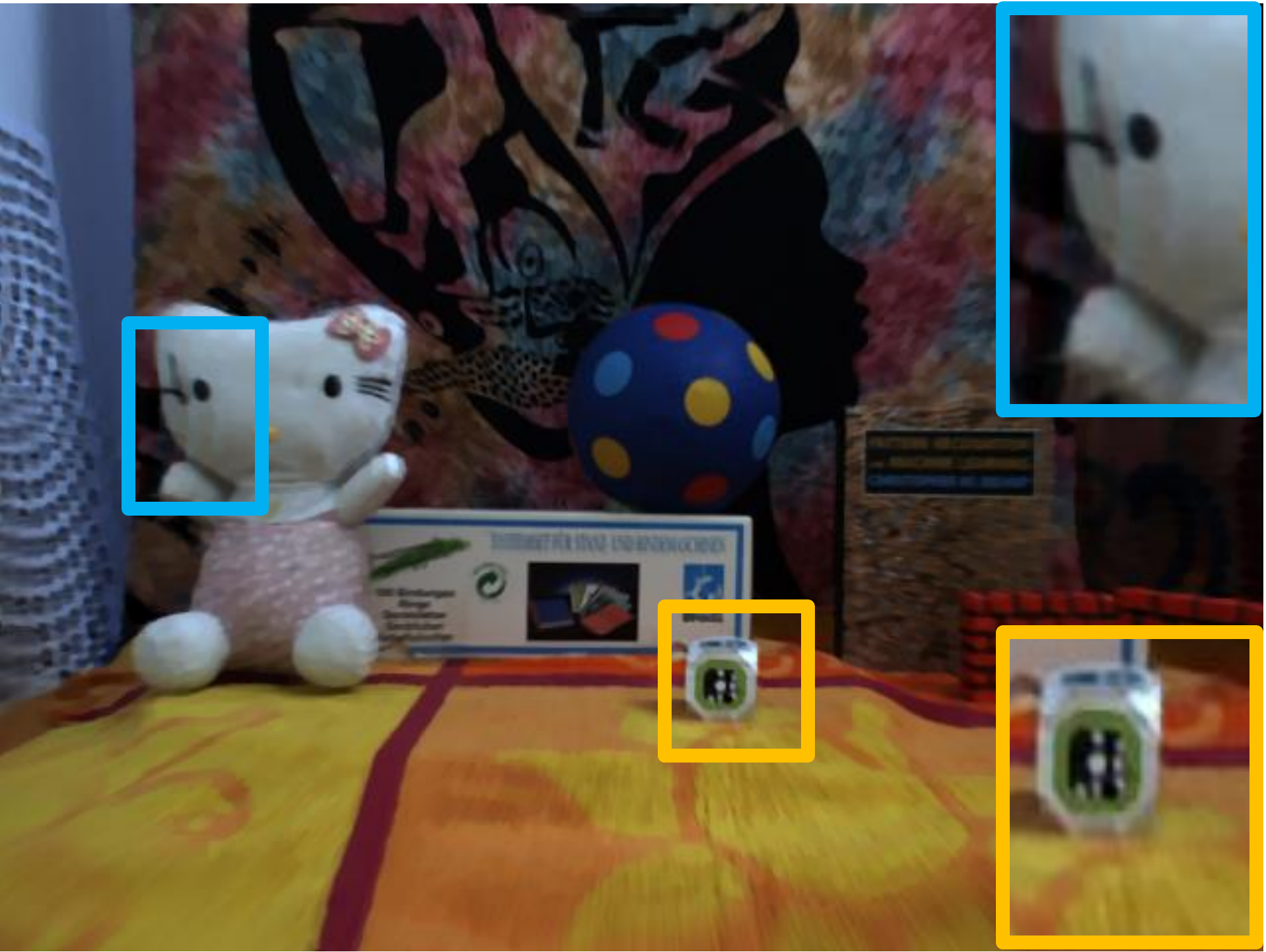}\\%[0.1in]
% \includegraphics[width=0.215\textwidth]{./fig_5/in.pdf}
% &\includegraphics[width=0.215\textwidth]{./fig_5/kim.pdf}
% &\includegraphics[width=0.215\textwidth]{./fig_5/as.pdf}
% &\includegraphics[width=0.215\textwidth]{./fig_5/our.pdf}\\%[0.1in]
\includegraphics[width=0.244\textwidth]{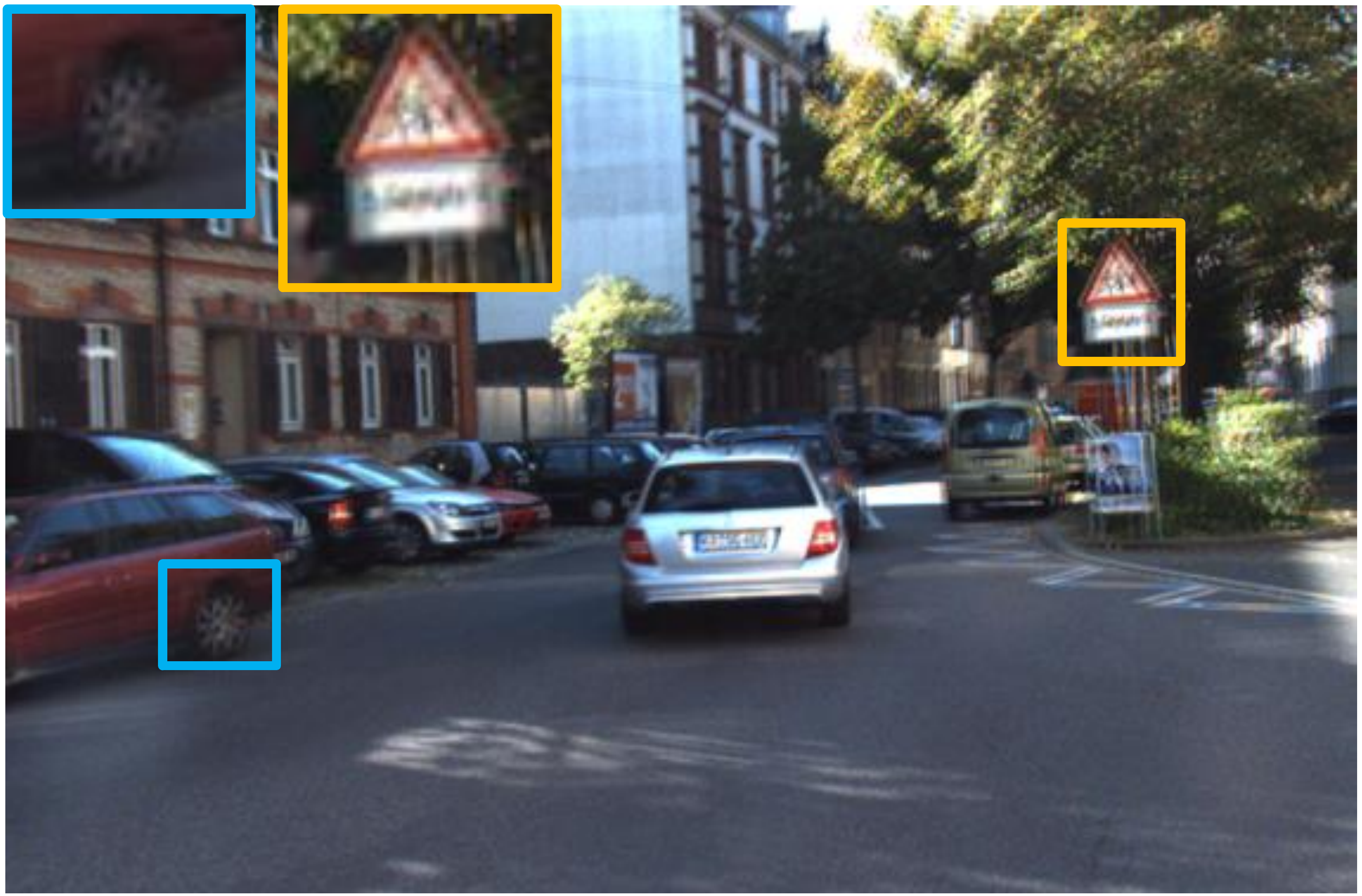}
&\includegraphics[width=0.244\textwidth]{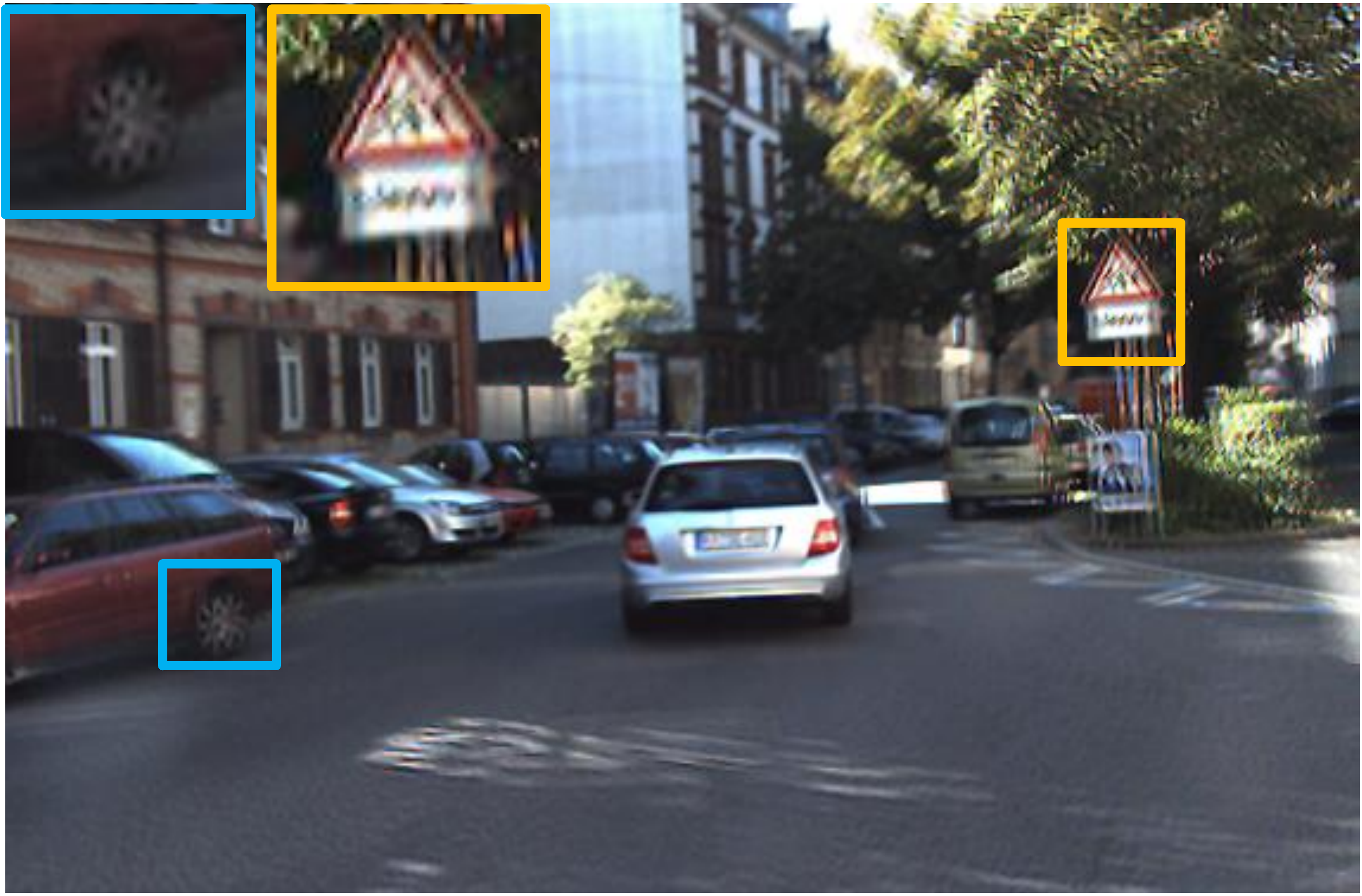}
&\includegraphics[width=0.244\textwidth]{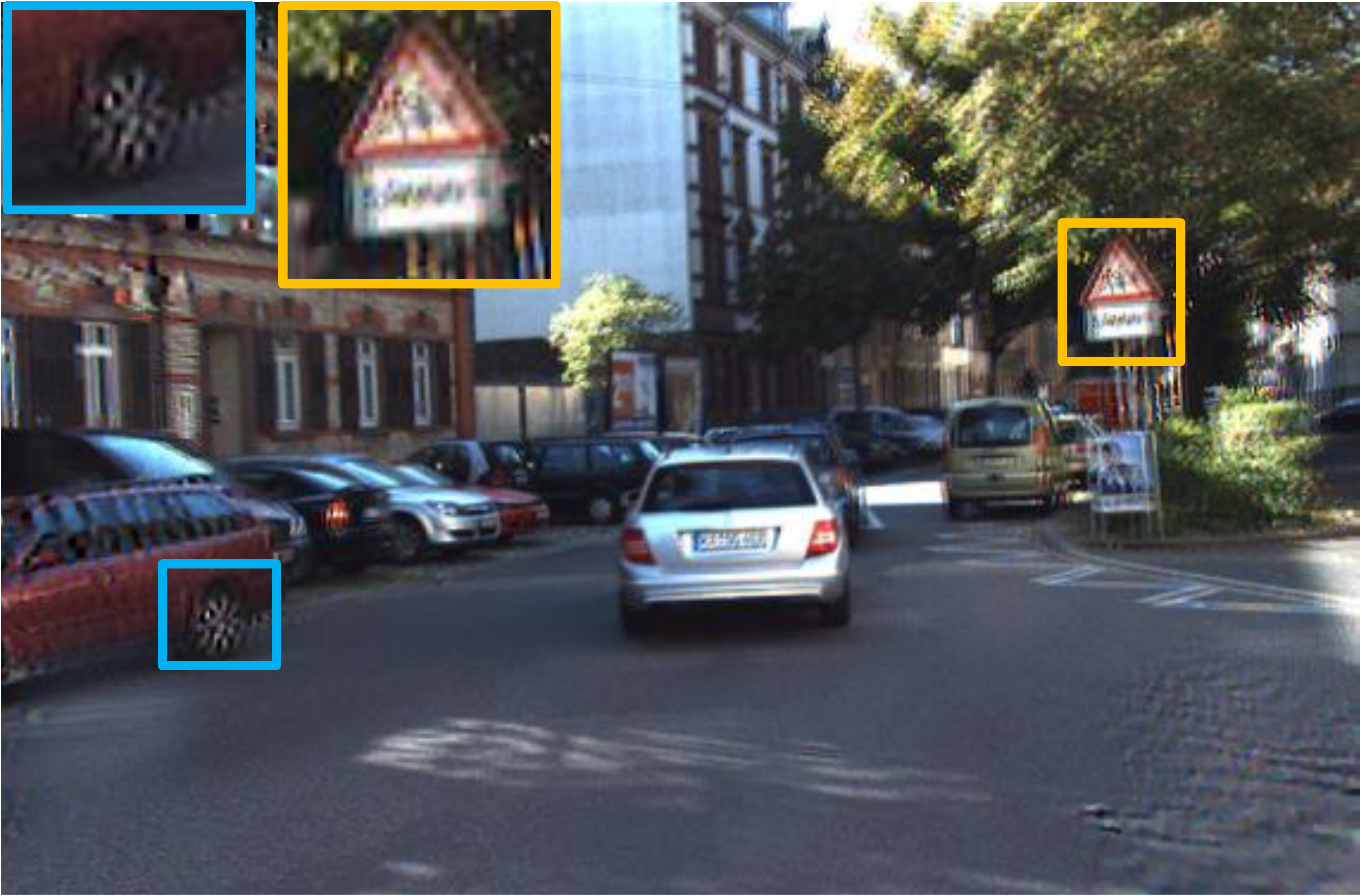}
&\includegraphics[width=0.244\textwidth]{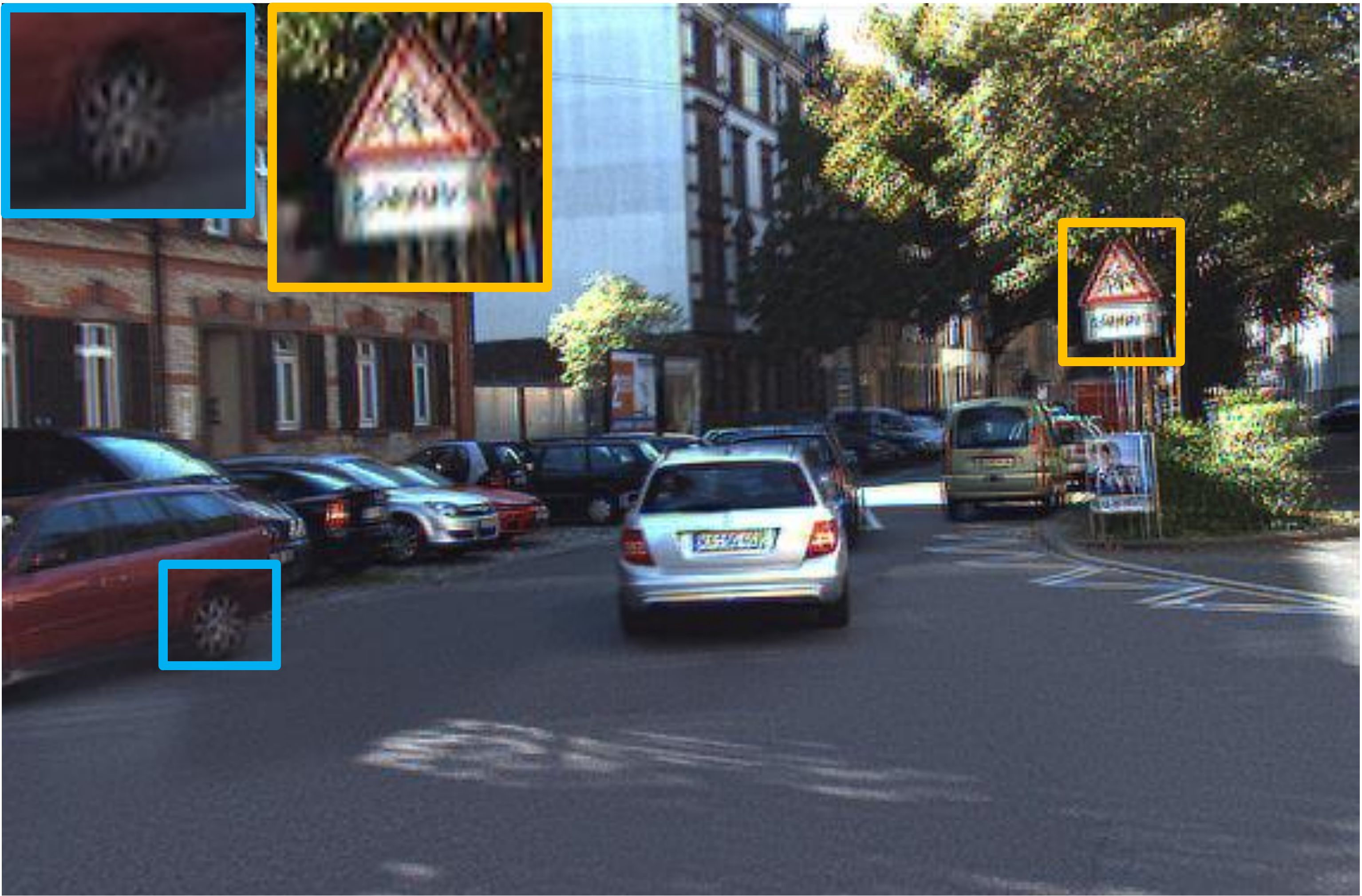}\\%[0.1in]
(a) Blurred Images 
& (b) Kim and Lee~\cite{hyun2015generalized}
& (c)  Sellent \etal ~\cite{sellent2016stereo}  
& (d) Our results\\
\end{tabular}
}
% \vspace{-2 mm}
\caption{\rc{Sample deblur results on the real image dataset from Sellent \etal~\cite{sellent2016stereo} in $1^{st}$ and $2^{nd}$ row, and average model dataset in $3^rd$ row. It shows that our 'generalized stereo deblur' model can tackle different kinds of motion blur model and get better results. Best viewed in colour on the screen.}}
\label{fig:realeccv16}
\end{center}
\end{figure*}
%=====================================================

%================= fig: real data from cai ============
\begin{figure}
\begin{center}
% \resizebox{\textwidth}{!}{
\begin{tabular}{cc}
\includegraphics[width=0.223\textwidth]{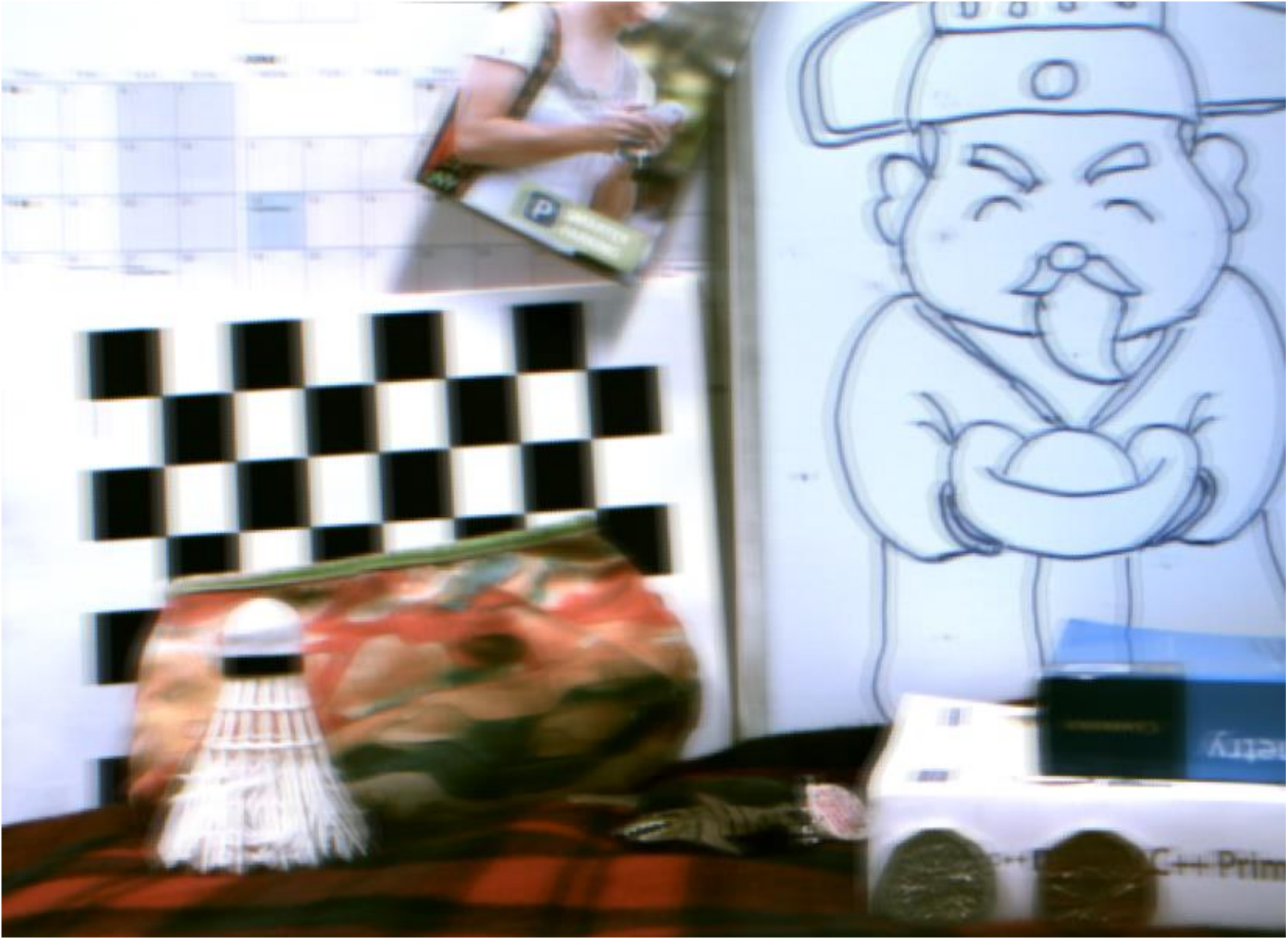}
&\includegraphics[width=0.223\textwidth]{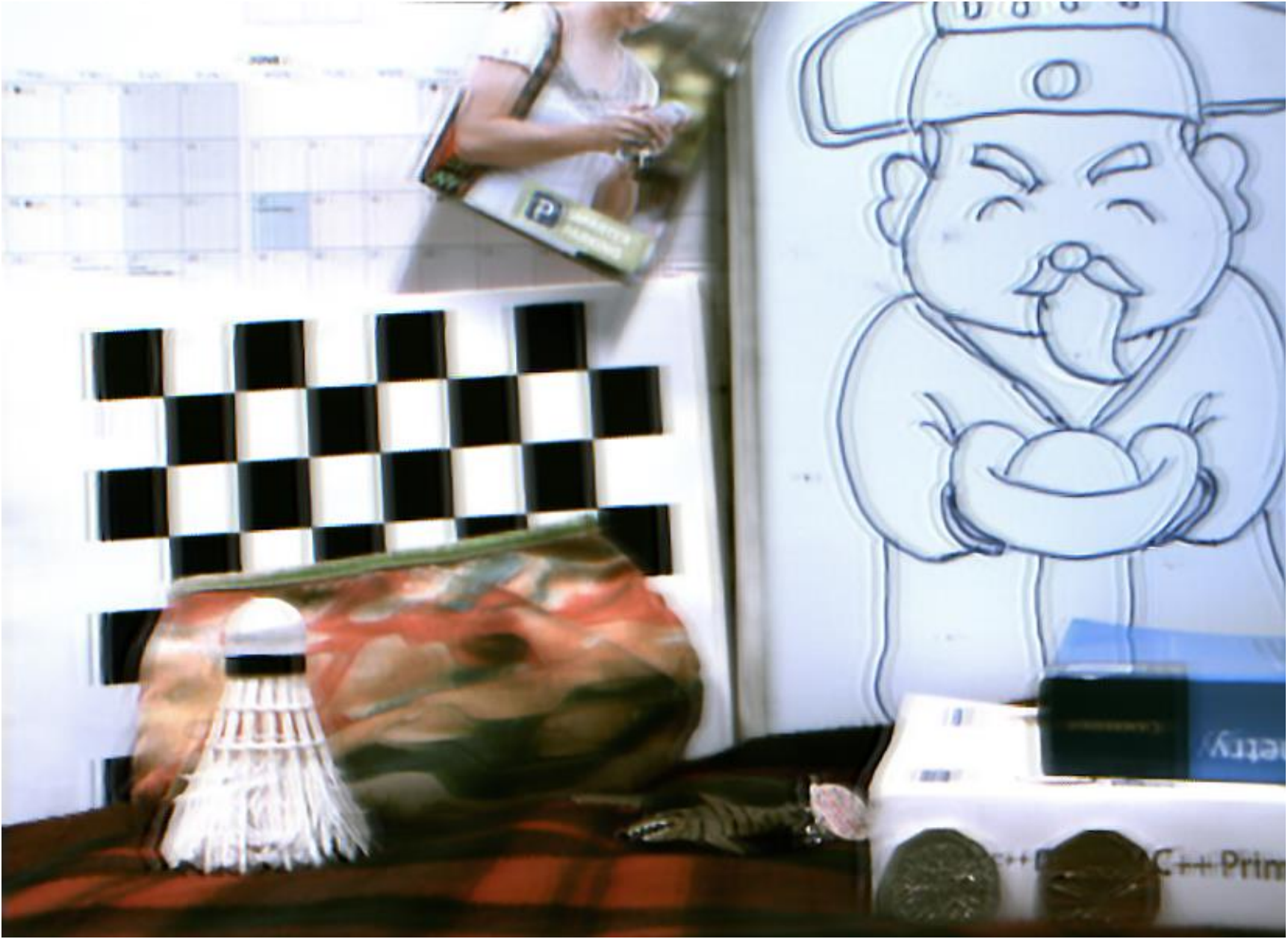}\\
(a) Blurred Images 
& (b) Kim and Lee~\cite{hyun2015generalized}\\
\includegraphics[width=0.223\textwidth]{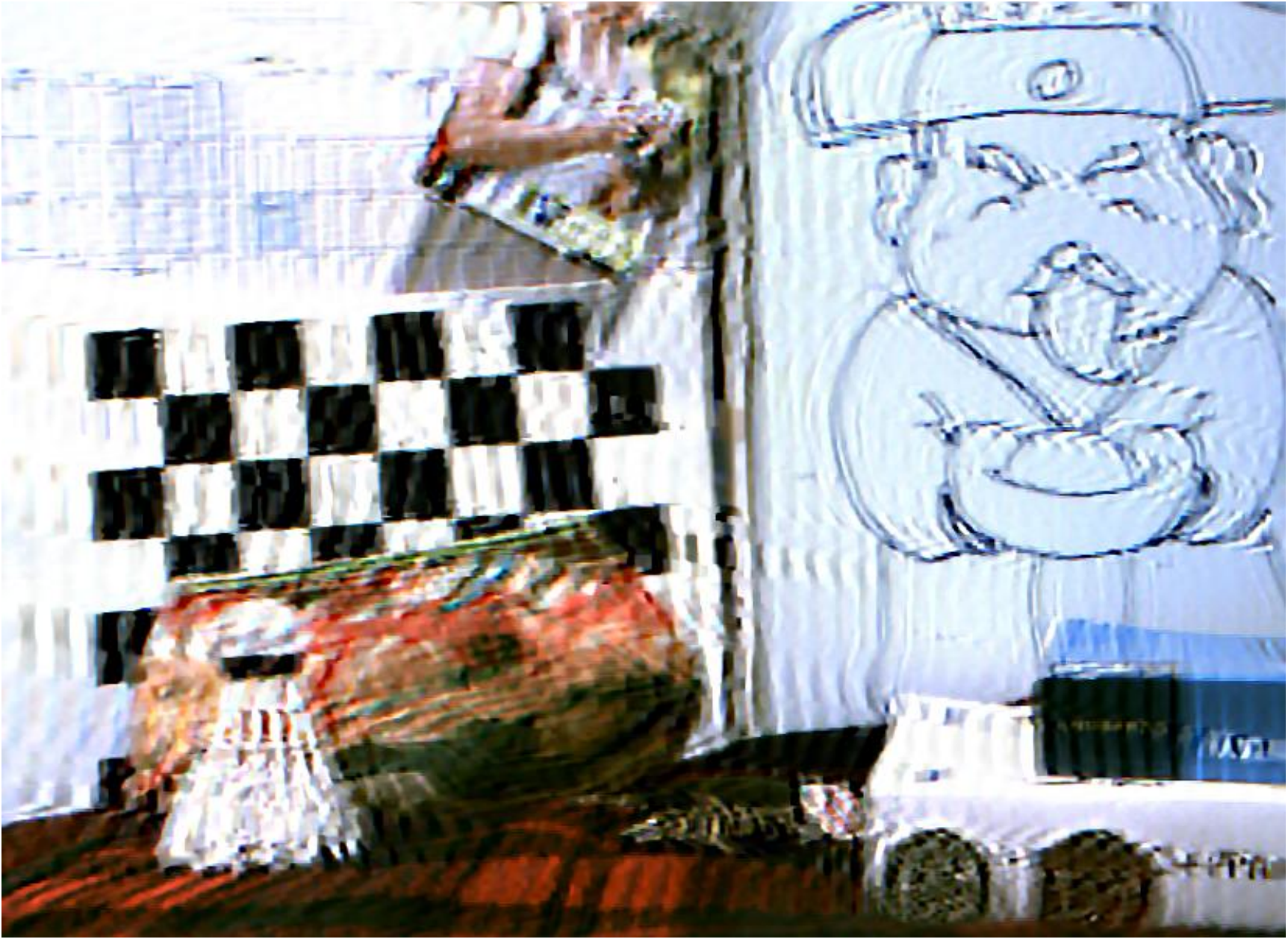}
&\includegraphics[width=0.223\textwidth]{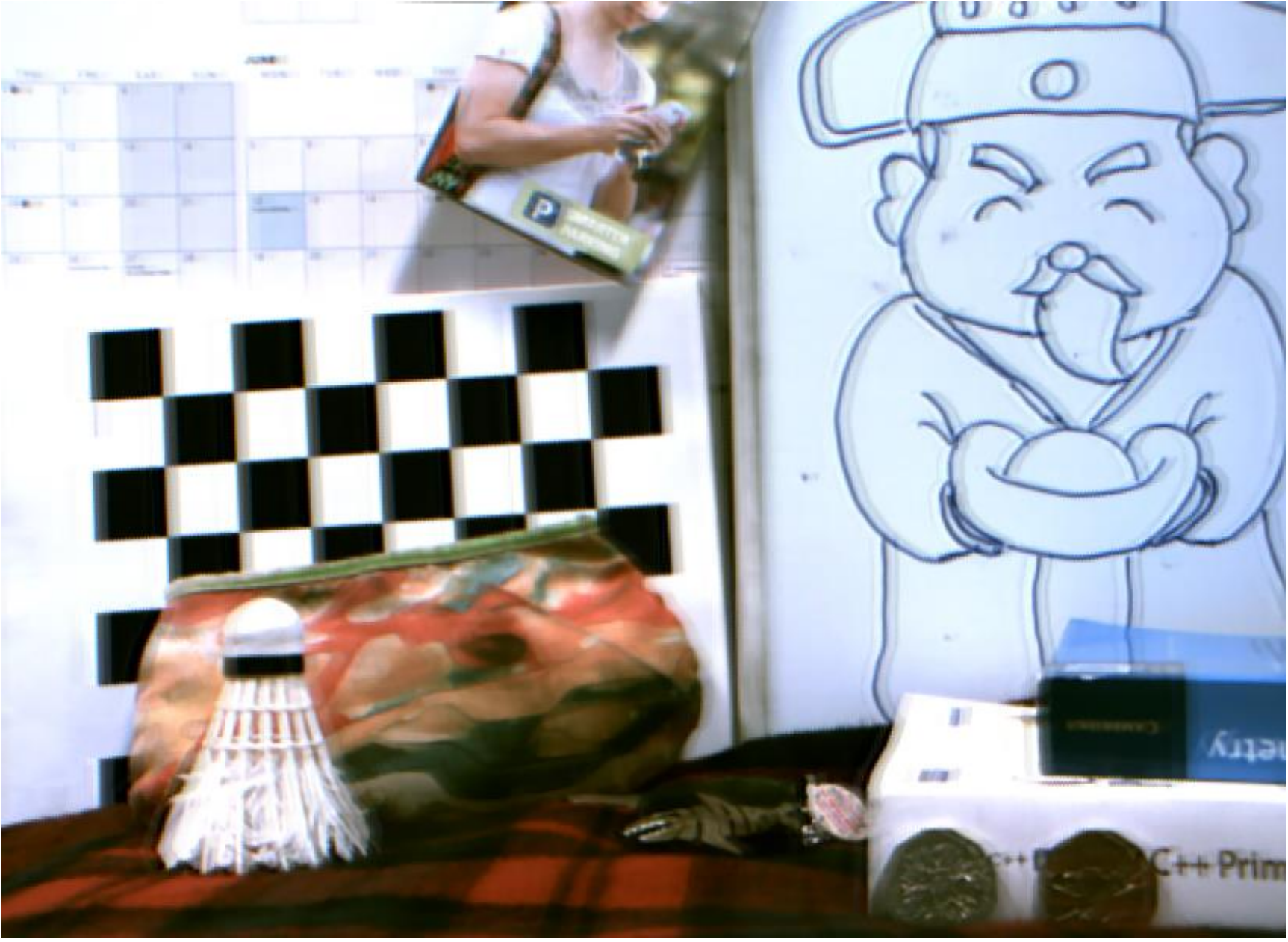}\\%[0.1in]
 (c)  Sellent \etal ~\cite{sellent2016stereo}  
& (d) Pan \etal ~\cite{Pan_2017_CVPR}\\
\includegraphics[width=0.223\textwidth]{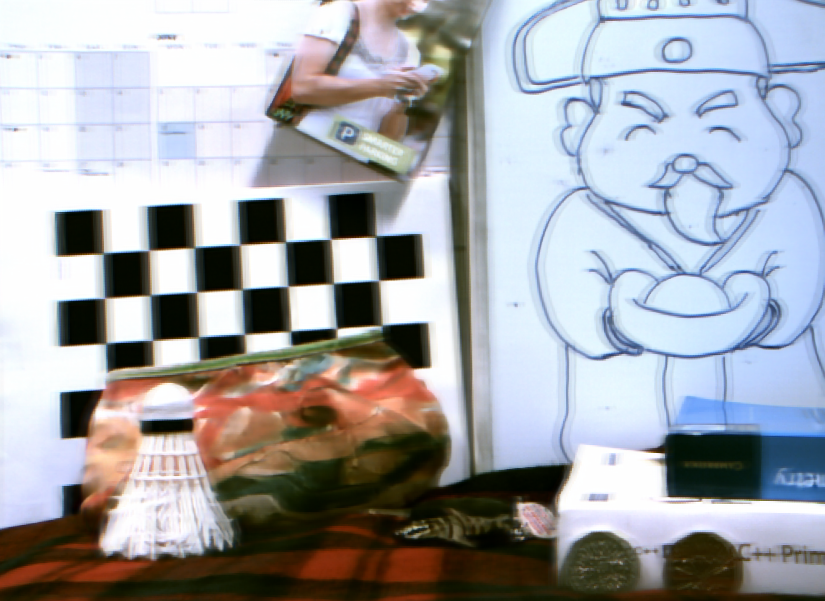}
&\includegraphics[width=0.223\textwidth]{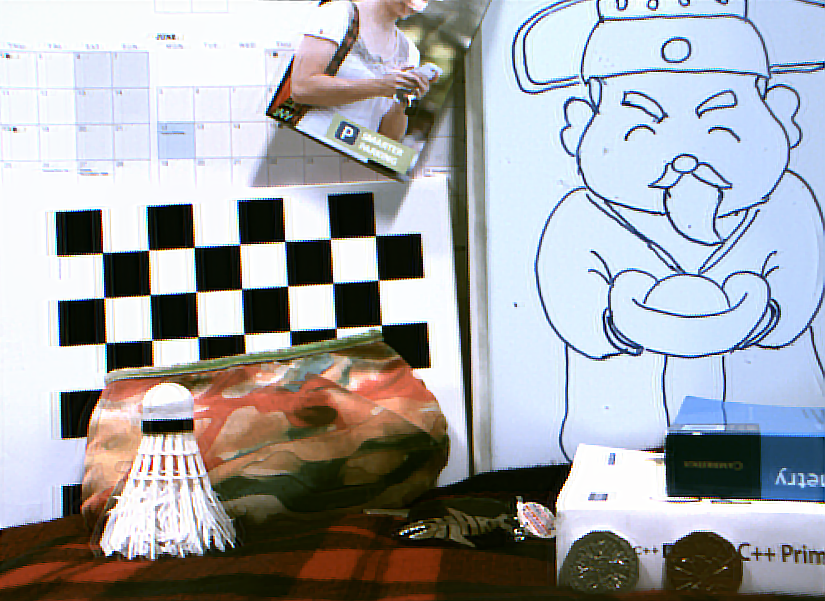}\\
(c)  Tao \etal ~\cite{Tao_2018_CVPR}  
& (d) Our results\\
\end{tabular}
% }
% \vspace{-2 mm}
\caption{\rc{Deblurring results on our Blur dataset. (a) The blurred image. (b) Deblurring results by Kim and Lee~\cite{hyun2015generalized}. (c) Stereo deblurring results by Sellent \etal ~\cite{sellent2016stereo}. (d) Deblurring results by Pan \etal ~\cite{Pan_2017_CVPR}. (e) Deblurring results by Tao \etal~\cite{Tao_2018_CVPR}. (f) Our result. It shows that our 'generalized stereo deblur' model can get competitive result compared with the state-of-the-art deblurring methods results. Best viewed in colour on the screen.}}
\label{fig:realcai}
\end{center}
\end{figure}
%=====================================================

\vspace{-2 mm}
\subsection{Results on KITTI}
To the best of our knowledge, currently, there are no realistic benchmark datasets that provide blurred images and corresponding ground-truth deblurring and scene flow. We take advantage of the KITTI dataset~\cite{geiger2013vision} to create a synthetic {\bf blurry image dataset} on realistic scenery, which contains 199 challenging outdoor sequences. Each sequence includes 6 images ($375 \times 1242$). Our blur image dataset is generated in two different ways.
First, we follow the general practice in image deblurring and generate the blur image dataset, referred to as~{\bf BlurData-1}, using the piecewise linear 2D kernel in Eq.~\ref{eq:kimblurKernel} which is defined on the dense scene flow.  
We use method~\cite{menze2015object} to generate dense ground-truth flows. \rc{In addition, $\tau = 0.23$ and the number of frame is $N = 20$~(see Fig.~\ref{fig:pipeline} for details).}

Second, we follow the way of generating blurry image in~\cite{kim2016dynamic}, by averaging the reference image together with its neighbouring frames. In particular, we average 7 frames in total (3 on either side of the reference frame). Note that
the image sequence in KITTI, in general, has large relative motion. We therefore only choose 10 sequences to generate blurry images based on averaging, which is denoted as {\bf BlurData-2}. In the following, we report results on our generated two synthetic datasets, respectively.

%Since the KITTI benchmarking dataset does not provide dense ground-truth flow, we use a state-of-the-art scene flow method~\cite{menze2015object} to generate dense ground-truth flows. Given the dense scene flow, the blurred images are generated using the piecewise linear 2D kernel~\ref{eq:piecewiseLinearKernal}, where the frame rate is set as $\tau = 0.23$ and the number of frame is $N = 20$~(see Figure~\ref{fig:pipeline} for details). Note that Our blur generation is quite realistic and follows the general practice in image deblurring.

\vspace{1mm}
\noindent{\bf Deblurring and Scene Flow Results.} We evaluated our approach by averaging errors and PSNR scores over $m$ and $m+1$ stereo image pairs.
%\MM{Why do we only evaluate on two pairs instead of three?}
Table~\ref{all_all} shows the PSNR values, disparity errors, and flow errors averaged over 199 test sequences on {\bf BlurData-1}. Note that our method consistently outperforms all baselines. We achieve the minimum error scores of 9.83\% for optical flow and 6.18\% for the disparity in the reference view. Figure~\ref{fig:err_compare} and Figure~\ref{fig:psnr_compare} show the estimated flows and deblurring results of the KITTI stereo flow benchmark, which includes 199 scenes. Figure \ref{fig:iter_psnr} (left) shows the performance of our deblurring stage with respect to the number of iterations. While we use 5 iterations for all our experiments, our experiments indicate that only 3 iterations are sufficient in most cases to reach an optimal performance under our model. In Figure \ref{fig:adddeblurresult}, we show qualitative results.% of our method and other methods on sample sequences from our dataset.

\vspace{1mm}
\noindent{\bf Moving Object Segmentation Results.} We report the quantitative comparison of our results with the baselines in Table~\ref{tab:movingComparison}. It shows that our approach significantly outperforms the baselines by a large margin. Fig.~\ref{fig:adddeblurresult}(g-k) show the qualitative comparison of our approach with baselines. The results show that our final segmentation follows the boundary of the moving objects very well. It further demonstrates that our approach can segment the moving objects more accurately than other approaches. Therefore, we can achieve a conclusion that joint scene flow estimation, deblurring, and moving object segmentation benefit each task.

% \vspace{1mm}
% \noindent{\bf Ablation Study.}
% We make the quantitative comparison of our model w/o explicitly imposing semantics priors for our flow and deblurring results in Fig~\ref{fig:err_compare}. Our previous method (Pan CVPR17) is the one that has no semantics priors. The comparison clearly show that the performance is improved significantly with the semantics as priors. In Fig.~\ref{fig:adddeblurresult}(d-e) and (j-k) qualitatively show that imposing the semantics prior can lead to better flow estimation (see the first row for the flow estimation).  

\vspace{-2 mm}
\subsection{Results on Other Dataset}
In order to evaluate the generalization ability of our approach on different images, we use the datasets based on the 3D kernel model and average kernel model which is different from our {\bf Blurred image dataset}.
In order to compare our performance on images blurred by the 3D kernel model, we also use the data courtesy of Sellent~\cite{sellent2016stereo}.
%\footnote{Courtesy of Sellent}. 
Those sequences contain four real and four synthetic scenes and each of them have six blurred images with its sharp images. %As only the ground-truth scene flow of synthetic chair sequence is available, we can only evaluate the performance of our method quantitatively for this scene(see Table \ref{chair} for details, where the evaluation results are averaged over 4 images). (see Figure \ref{fig:realeccv16} $1^{st}$ and $2^{nd}$ rows show the performance of the result of real scene)
The synthetic sequences are blurred by the 3D kernel model and have ground-truth for those sequences. Figure \ref{fig:iter_psnr} (right) shows the performance of several baselines on synthetic dataset. This plot affirms our assumption that jointly and simultaneously solving scene flow and video deblur benefit each other. It also shows that a simple combination of two stages cannot achieve the targeted results.
For real scenes, they use real images captured with a stereo camera which moves forward very slowly and attached to a motorized rail. By averaging the frames, they obtain motion blurred images where all objects in the scene are static and the camera moves toward the scene. For these reasons, we give the semantic segmentation map as all background (see Figure \ref{fig:realeccv16} $1^{st}$ and $2^{nd}$ rows show the performance of the result of the real scene).

In Fig. \ref{fig:realeccv16}( the $3^{rd}$ and $4^{th}$ rows.), we show qualitative results of our method and other methods on sample sequences from this two datasets, where our method again achieves the best performance.

\vspace{-3 mm}
\subsection{\rc{Limitations}}
\rc{Our method is based on calibrated stereo cameras which seem sometimes not convenient for routine application. The framework may fail in the texture-less case, the scene with strong reflection or under low lighting conditions. The occlusion will also reduce the accuracy of the segmentation boundaries. Our model cannot tackle defocus blur and scenery with transparency or translucency. }
\rcs{Following the recent deblurring works such as~\cite{kim2014segmentation,gupta2010single,dai2008motion,whyte2012non}, we make the similar assumption that the intensity integral happens in colour space during the exposure time, while we are aware of several methods model the integration in the raw sensor value and consider the effects of CRFs on motion deblurring \cite{Nah_2017_CVPR,tai2013nonlinear}.
We leave these limitations as future works.}